\documentclass[12pt, ams,  onecolumn, amssymb, longbibliography, floatfix]{revtex4-2}

\usepackage[utf8]{inputenc}
\usepackage{amsmath}
\usepackage{amsfonts}
\usepackage{amssymb}
\usepackage{graphicx}
\usepackage{bm}% bold math
\usepackage{float}
\usepackage{ulem}
\usepackage{cancel} 
\usepackage{slashed}
\usepackage{comment}
\usepackage{mathtools}
\usepackage{booktabs}

\usepackage{adjustbox}

\usepackage{float}
\usepackage{multirow}

\usepackage{ulem}

\usepackage{enumitem}

\usepackage{caption}
\usepackage{subcaption}

\usepackage{fixltx2e,amsmath}
\usepackage[titletoc,toc]{appendix} 
\usepackage{extarrows}

\usepackage{natbib}
\newcommand{\mycomment}[1]{%
}% 

\usepackage[colorlinks=true,
            linkcolor=OrangeRed,
            urlcolor=blue,
            citecolor=gray]{hyperref}
 \hypersetup{citecolor=RoyalBlue} 
\usepackage[usenames,dvipsnames]{color}
\definecolor{dkgray}{RGB}{145,145,145}
\definecolor{violet}{RGB}{50,0,200}

\numberwithin{equation}{section}

\begin{document}

\title{Spontaneous Kolmogorov-Arnold Geometry in Shallow MLPs}
\author{Michael H. Freedman$^{1,2}$}
\email{michael.freedman@logicalintelligence.com, mfreedman@cmsa.fas.harvard.edu}
\author{Michael Mulligan$^{1,3}$}
\email{michael.mulligan@ucr.edu}
\affiliation{$^{1}$Logical Intelligence}
\affiliation{$^{2}$Center of Mathematical Sciences and Applications, Harvard University, Cambridge, MA 02138, USA}
\affiliation{$^{3}$Department of Physics and Astronomy, University of California, Riverside, CA 92521, USA}

\date{\today}

 \bigskip
 \bigskip
 \bigskip

\begin{abstract}
The Kolmogorov-Arnold (KA) representation theorem constructs universal, but highly non-smooth inner functions (the first layer map) in a single (non-linear) hidden layer neural network. 
Such universal functions have a distinctive local geometry, a ``texture," which can be characterized by the inner function's Jacobian $J({\mathbf{x}})$, as $\mathbf{x}$ varies over the data. 
It is natural to ask if this distinctive KA geometry emerges through conventional neural network optimization. 
We find that indeed KA geometry often is produced when training vanilla single hidden layer neural networks. 
We quantify KA geometry through the statistical properties of the exterior powers of $J(\mathbf{x})$: number of zero rows and various observables for the minor statistics of $J(\mathbf{x})$, which measure the scale and axis alignment of $J(\mathbf{x})$.
This leads to a rough understanding for where KA geometry occurs in the space of function complexity and model hyperparameters. 
The motivation is first to understand how neural networks organically learn to prepare input data for later downstream processing and, second, to learn enough about the emergence of KA geometry to accelerate learning through a timely intervention in network hyperparameters. 
This research is the ``flip side" of KA-Networks (KANs). 
We do not engineer KA into the neural network, but rather watch KA emerge in shallow MLPs.
\end{abstract}

\maketitle

\bigskip

\newpage

\setcounter{page}{1}

%%%%%%%%%%%%%%%%%%%%%%%%%%%%%%%%%%%%%%%%%%%%%%%%%%%%%%%%%%%%%%%%%%%%%%%%%%%%%%%%%%%%%%%%%
\definecolor{napiergreen}{rgb}{0.16, 0.5, 0.0}
\definecolor{officegreen}{rgb}{0.0, 0.5, 0.0}

\definecolor{seagreen}{rgb}{0.18, 0.55, 0.34}
\definecolor{sacramentostategreen}{rgb}{0.0, 0.34, 0.25}
\definecolor{upforestgreen}{rgb}{0.0, 0.27, 0.13}

\definecolor{tropicalrainforest}{rgb}{0.0, 0.46, 0.37}
%\definecolor{tropicalrainforest}{rgb}{0.0, 0.5, 0.3}

\definecolor{viridian}{rgb}{0.25, 0.51, 0.43}

\definecolor{pakistangreen}{rgb}{0.0, 0.35, 0.0}

\definecolor{GreenForTableofCont}{rgb}{0.0, 0.5, 0.32}

\definecolor{navyblue}{rgb}{0.0, 0.0, 0.5}
\definecolor{persianblue}{rgb}{0.11, 0.22, 0.73}

	\definecolor{brickred}{rgb}{0.8, 0.25, 0.33}
 	\definecolor{brown(web)}{rgb}{0.65, 0.16, 0.16}

{  \hypersetup{linkcolor=persianblue}
   \tableofcontents
}

%%%%%%%%%%%%%%%%%%%%%%%%%%%%%%%%%%%%%%%%%%%%%%%%%%%%%%
\newpage

%\vskip 1cm

\section{Introduction}

The quadratic formula expresses the roots of a 2nd degree polynomial in terms of its coefficients. 
That formula, and similar formulas for the roots of third and fourth degree polynomials, combine interesting single-variable functions (radicals) using only addition (and multiplication, which can be reduced to addition using $\log$ and $\exp$).
Abel famously showed that such concise formulas for roots cannot exist beyond degree 4.
In his thirteenth problem, Hilbert asked mathematicians to prove a vast generalization of Abel's theorem that would eliminate the possibility of expressing arbitrary continuous multivariate functions in terms of general continuous single-variable functions using only addition.
In the mid-1950s Kolmogorov and Arnold (KA) surprised the mathematical world by showing, contrary to Hilbert's expectation, that neural networks with general continuous but nonlinear functions connecting the neurons---KA networks---suffice to represent any continuous multivariate function. 

When computer scientists first considered KA's work \cite{HechtNielsen1987KolmogorovsMN}, its relevancy was discounted on several grounds \cite{10.1162/neco.1989.1.4.465}.
The theorem dealt with representation rather than training.
Although the KA networks are remarkably small (a single hidden layer of fixed width), the inter-layer functions are quite far from smooth, even if the function $f(x_1, \ldots,x_n)$ to be represented is analytic. 
As such, it was felt that it would be difficult to explicitly describe a KA network, much less discover one through training. 

This paper continues a modern reconsideration \cite{10.1162/neco.1991.3.4.617, liu2025kan, ji2025comprehensivesurveykolmogorovarnold, 2024arXiv241008451F, 2025arXiv250405255D} of that judgement. 
Our approach here is not to use the KA construction as motivation for building novel neural network architectures, as it was in \cite{liu2025kan}, but instead to view the KA network as one of many possible effective solutions available to conventional neural network optimization and to determine the extent to which such networks are realized through training.
(By ``conventional neural network," we mean one in which a layer map consists of an affine transformation followed by a nonlinear activation function.)
Our motivation for doing this is that KA-like solutions are empirically favorable for certain tasks \cite{liu2025kan, ji2025comprehensivesurveykolmogorovarnold}  and, when this is the case, we should encourage their development. 

This paper focuses on nearly the simplest possible setting: curve-fitting 3-variable functions.
We abstract certain essential features of the KA construction---collectively called KA geometry or, simply, KA---and show that (approximate) KA geometry can develop spontaneously through training in 
shallow (1-hidden layer) neural networks of the most ordinary variety (linear layer maps with GeLU activation).
This occurs within an identifiable window in the space of target functions, given a fixed model capacity---a sort of ``Goldilocks regime" where learning is both not too easy and not impossible.
For us, this ``Goldilocks regime" is represented by the \texttt{xor} function---defined in \eqref{xorfunction}---and its perturbations.
It is exciting that ``nature," i.e., gradient descent, hits upon the same (well, similar) construction to achieve expressivity as teleologically devised by two great mathematicians of the previous century. 
The original KA construction provides an ingenious universal map from data to a latent manifold, which prepares the data propitiously. 
We find this construction to be attractive not only for representation, but also for learning and generalization.
In fact, it is so canonical that gradient descent (more specifically, Adam) can land on the same strategy, absent any coaching.
 
\section{KA Geometry}

\subsection{What It Is}
\label{whatissection}

Let us discuss what KA geometry is in the ideal case of the KA theorem and then abstract from this ideal case while continuing to call certain features (approximate) KA geometry when we find them in conventional neural networks.

The KA theorem says that any continuous function $f: I^n \rightarrow \mathbb{R}$ can be represented by a 1-hidden layer neural network with $m \geq 2n + 1$ hidden neurons:
\begin{align}
\label{kadef}
f(x_1, \ldots, x_n) = \sum_{j = 1}^m g_j \Big( \sum_i^n \phi_{ij}(x_i)  \Big).
\end{align}
Here, $\phi_{ij} : I \rightarrow \mathbb{R}$ are general nonlinear functions that may be taken Lipschitz \cite{fridman1967improvement, 2017arXiv171208286A}, while the $g_j : \mathbb{R} \rightarrow I$ are continuous.
We use KA geometry to refer to specific properties of the inner function $\Phi: I^n \rightarrow \mathbb{R}^m$, the map that embeds the input data into the hidden dimension and facilitates the construction of the outer functions $g_j$.
Denote $\Phi_j =  \sum_i \phi_{ij}(x_i)$.
(The inner and outer functions can be further refined:
the inner function can be taken to be $\Phi_j = \lambda_1 \phi_j\big(x_1\big) + \ldots + \lambda_n \phi_j\big(x_n\big)$, for certain real numbers $\lambda_i$ and Lipschitz functions $\phi_j: I \rightarrow \mathbb{R}$;
the outer functions $g_j$ may all be chosen to be the same $g_j = g$.)

For concreteness, let's briefly consider $n = 2$ input dimensions and $m = 5$ hidden neurons, the smallest nontrivial case. 
The inner function $\Phi: I^2 \rightarrow \mathbb{R}^5$ can be chosen universally (i.e., independently of the target function $f$) with a remarkable property: At every input point $\mathbf{x} = (x_1, x_2) \in I^2$, the Jacobian of $\Phi$---the $5 \times 2$ matrix $J_{ji}({\bf x}) = \partial \Phi_j({\bf x}) / \partial x_i$---has at least three rows identically zero. 
This means the embedding is locally constant along a majority of coordinate directions in $\mathbb{R}^5$. 
Which three rows vanish varies as $(x_1, x_2)$ moves through $I^2$.
This patterning of the Jacobian is one aspect of KA geometry.
The locally constant values of the inner map's Jacobian should be rationally independent, though we rely on genericity for this condition and do not study it. 
By Rademacher's theorem, Lipschitz functions have almost everywhere defined Jacobians, making this analysis well defined.
In fact, the KA-construction results in an almost everywhere defined Jacobian with at most a single nonzero entry per column, so $J(x_1, x_2)$ generally looks something like $\begin{pmatrix} 0 & a & 0 & 0 & 0 \cr 0 & 0 & 0 & b & 0  \end{pmatrix}^T$, for constants $a, b$.
(Note the transpose, since ``rows" and ``columns" are discussed here.)
Under the KA construction as $(x_1,x_2)$ varies the Jacobian rank will briefly drop and then climb back to $2$ (or $n$, in general).  
The most common situation, losing one variational direction followed by gaining a new and different one, amounts, when integrated, to a local $90$ degree rotation of the embedded data manifold.
This produces a topological skeleton within the input (data) manifold where the Jacobian rank becomes less than full, decreasing further with the skeletal-codimension.
This feature of KA's construction is harder to identify in numerical experiments, since sampling from a codimension-$j$ skeleton requires search over a $j$-dimensional parameter space, and has been omitted from this preliminary numerical study.

This majority of inactive coordinates---three out of five in our example---enables the construction of the outer function $g$ (or outer functions $g_j$) through an iterative approximation scheme. 
The key insight \cite{shapiro2006topics, lorentz1996constructive} is that even a crude guess for $g$ on the stationary coordinates suffices to make progress. Specifically, at any point $\mathbf{x}$, where three coordinates of $\Phi$ are locally constant, we can assign $g$ the value $f(\mathbf{x}')/3$ on each of these stationary coordinates, where $\mathbf{x}'$ is a point selected uniformly close to $\mathbf{x}$. 
(In the general case, the guess $f(\mathbf{x}')/3$ is replaced  by $f(\mathbf{x}')/(m-n)$.)
This simple equal distribution of the target function's value yields an approximation $f_\text{approx}$ (the RHS of \eqref{kadef} with this guess for $g$) satisfying $\|f - f_\text{approx}\| < c \|f\|$ for some constant $c < 1$.
Treating the error $f - f_\text{approx}$ as a new target function, we repeat the process, achieving exponential convergence to an exact representation of $f$ in the limit. 
Crucially, this convergence requires that inactive coordinates outnumber active ones---here, three inactive versus two active. 
This majority condition explains the theorem's requirement that $m \geq 2n + 1$: ensuring that at least $(m-n)$ of the $m$ coordinates are stationary at every point.
One observation is that, while $m \geq 2 n + 1$ guarantees convergence of the above iterative scheme, the rate of convergence increases with increasing hidden dimension $m$.

We will look exclusively at conventional 1-hidden layer fully-connected neural networks (MLPs),
\begin{align}
\label{modeldef}
f(x_1, \ldots, x_n) \approx f_{\rm MLP}(x_1, \ldots, x_n) = \sigma \Big( {\bf x} \cdot A + a \Big) \cdot B + b,
\end{align}
Here, $A$ and $B$ are learnable $n \times m$ and $m \times 1$ matrices; $a$ is an $m$-dimensional vector and $b$ is a scalar; the nonlinearity $\sigma({\bf z}) = {\rm GeLU}({\bf z}) = \big(z_1 F(z_1), \ldots, z_m F(z_m) \big)$, with $F(z)$ the standard normal cumulative distribution function and ${\bf z} = {\bf x} \cdot A + a$.
Notice that the first layer map in \eqref{modeldef} gently folds the input manifold at most $m$ times.
These gentle folds each achieve an integrated rotation of (only) $45$ degrees, following the graph of the activation function.
The ideal KA embedding has a plethora of sharper $90$ degree bends, the actual number approaching infinity as the approximation error goes to zero.
Comparing networks \eqref{kadef} and \eqref{modeldef}, we see that $\sigma({\bf z})$ plays the role of $\Phi({\bf x})$ and the sum of outer functions $\sum_j g_j$ corresponds to the affine map defined by $B$ and $b$.
The Jacobian of $\sigma$ is (no sum over $j$)
\begin{align}
\label{jacobianformula}
J_{j i}({\bf x}) = \sigma_j' A^T_{j i} = \big(F(z_j) + z_j F'(z_j) \big) A^T_{j i}. 
\end{align}
Thus a row of $J_{j i}({\bf x})$ is zero when either $\sigma_j'$ vanishes or a row of $A^T$ is zero.

\begin{figure}[htbp]
    \centering
    
    % Left subplot with weight matrices
    \begin{subfigure}[t]{0.52\textwidth}
        \centering
        \adjustbox{valign=t}{\includegraphics[height=6.0cm]{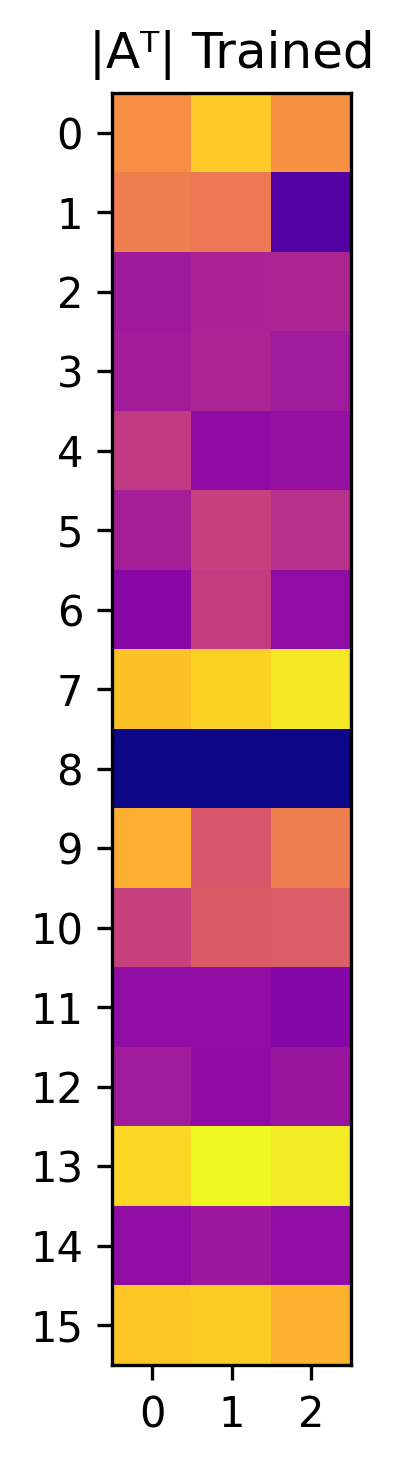}}%
        \hspace{0.05cm}%
        \adjustbox{valign=t}{\includegraphics[height=6.0cm]{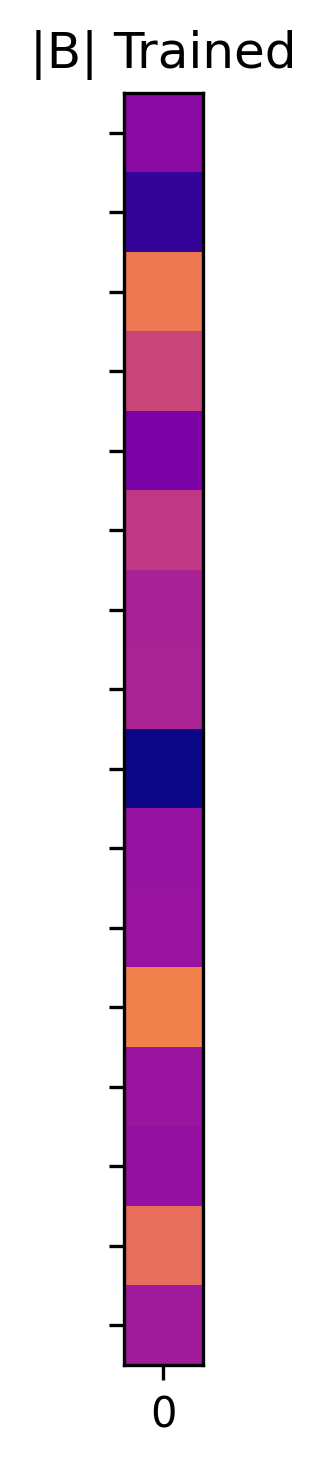}}%
        \hspace{0.2cm}%
        \adjustbox{valign=t}{\includegraphics[height=6.0cm]{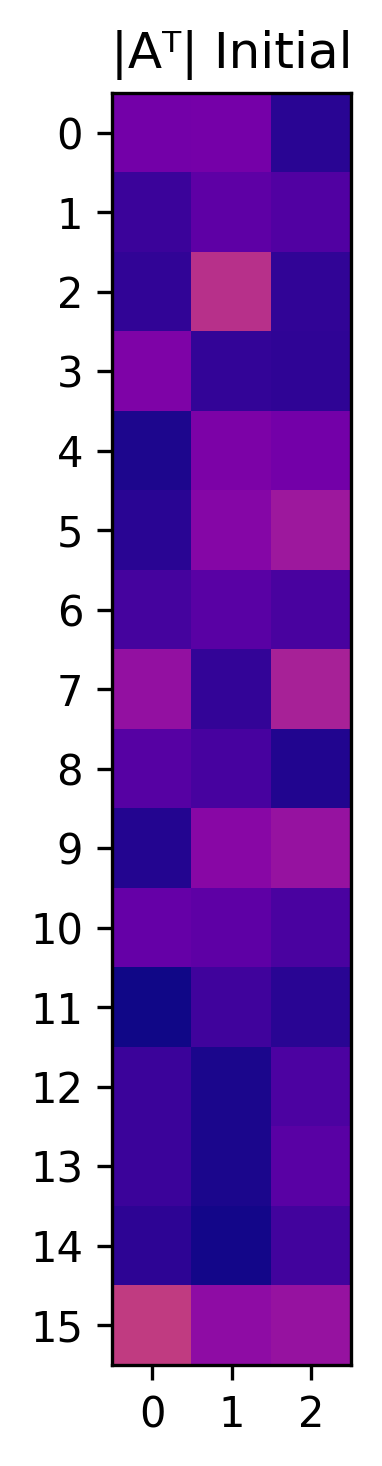}}%
        \hspace{0.05cm}%
        \adjustbox{valign=t}{\includegraphics[height=6.0cm]{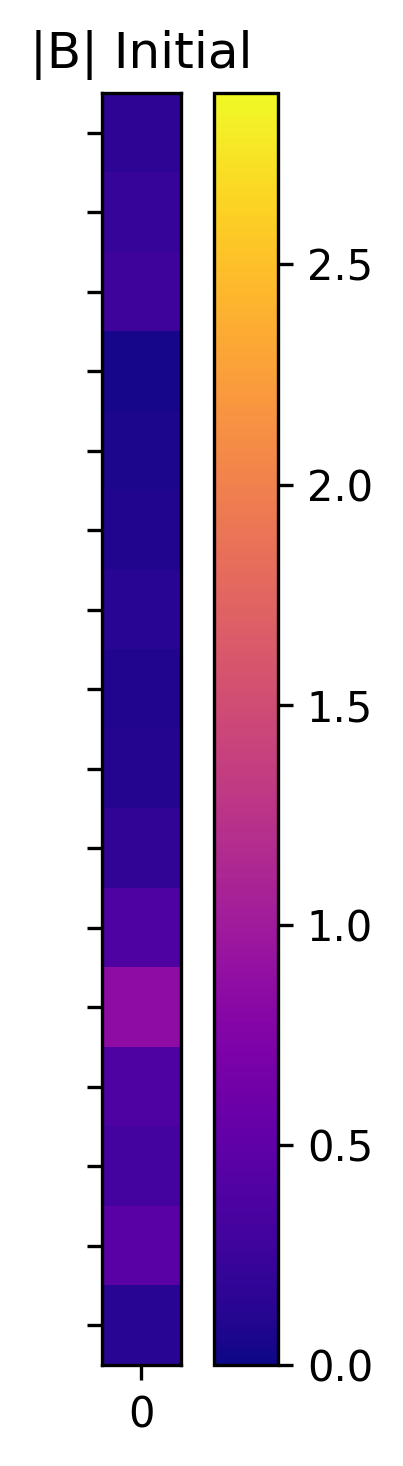}}
        \caption{Weights}
        \label{fig:weight_matrices}
    \end{subfigure}
    \hfill
    % Right subplot with Jacobian matrices
    \begin{subfigure}[t]{0.45\textwidth}
        \centering
        \adjustbox{valign=t}{\includegraphics[height=6.0cm]{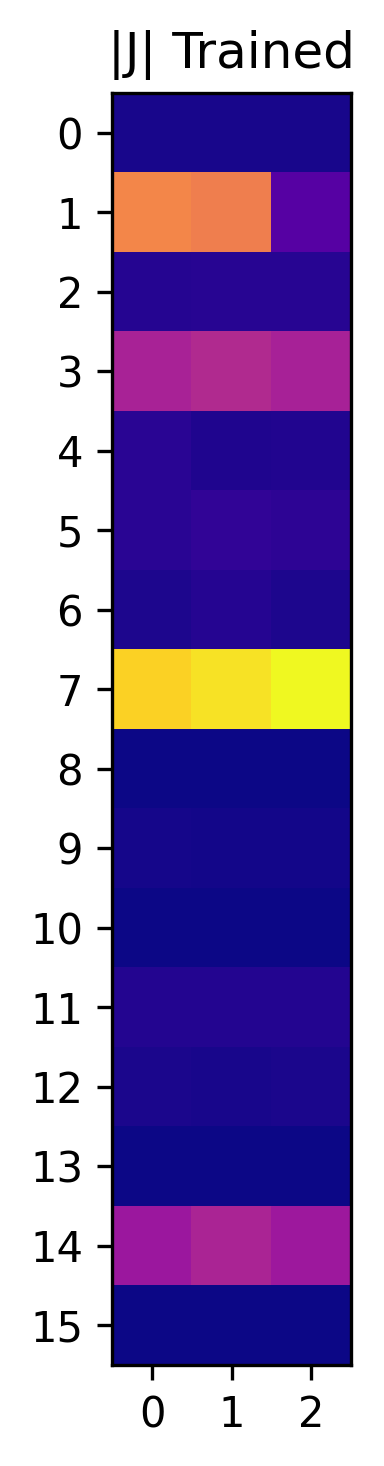}}%
        \hspace{0.1cm}%
        \adjustbox{valign=t}{\includegraphics[height=6.0cm]{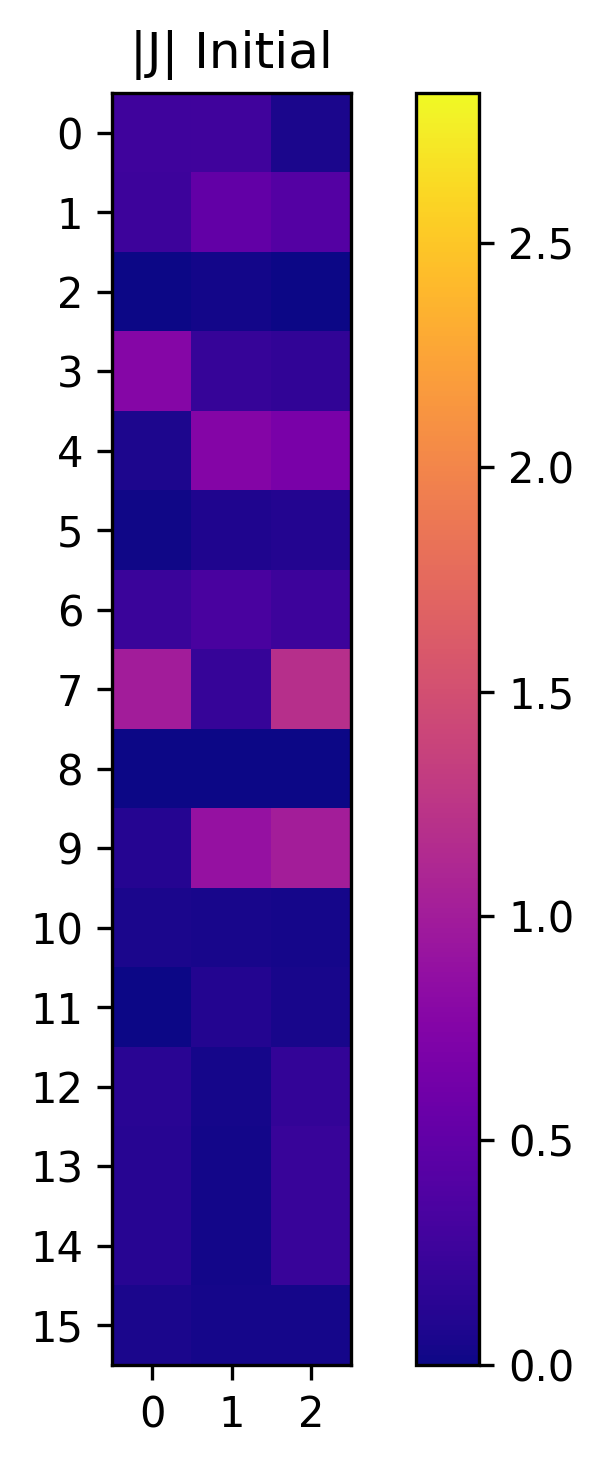}}
        \caption{Jacobians}
        \label{fig:jacobian_matrices}
    \end{subfigure}
    
    \vspace{0.5cm}
    \caption{Weight and Jacobian heatmaps of a model before and after training, plotted on a linear scale.
The model has $16$ hidden neurons and was trained on the \texttt{xor} function \eqref{xorfunction}.
The right panel is the Jacobian of a single, randomly chosen example.
There is one row of relatively small values in the trained weight matrix $A^T$: row 8.
Consistent with \eqref{jacobianformula} this row produces a correspondingly small row in the trained Jacobian.
There are many relatively small/large rows of the trained Jacobian that are not directly correlated with the weight matrix $A^T$, but must instead result from the derivative factor in \eqref{jacobianformula}, suggestive of KA geometry.}
    \label{MinorWeightHeatmaps}
\end{figure}
The zero row aspect of KA geometry is considered first.
It is too naive to expect a row of $J_{ji}$ to be exactly zero in our numerical experiments (see Fig.~\ref{MinorWeightHeatmaps}).
Because we observe data, it is natural to select some numerical cutoff, below which an entry or even an entire row of the Jacobian is declared to be ``zero."
It is harder than one might expect to do this in a principled way.
During training we observe large shifts in the quantiles of row means (among many metrics).
This leads us to formulate a definition of ``zero" that we believe to be a more honest measure than any fixed numerical cutoff (see \S \ref{zerorowsection}).
Once this is done we can ask about the number of ``zero rows" of the Jacobian (evaluated on both training and test data) as training progresses. 
Throughout we distinguish between rows that are consistently zero across (nearly) all data points---representing inactive hidden neurons---versus example-dependent zero rows---reflective of KA geometry.

We then proceed to look at $k \times k$ minors of the Jacobian, for $k = 2, 3$, which, in the language of matrices, are the determinants of the submatrices obtained by restricting to those entries lying in some fixed collection of $k$ rows and some fixed collection of $k$ columns.
Mathematically, $k \times k$ minors of the Jacobian are the entries in the $k$-th exterior power of the Jacobian $J^{(k)}$, in the natural neuron bases in domain and range.
We notice no sign asymmetries, so, in practice, it is the absolute value of a minor that we consider.
We make an effort at each stage to filter out observations which are a direct consequence of lower-order effects.
For example if (either of) a pair of rows of the Jacobian are (nearly) zero, it is no surprise $2 \times 2$ minors selected from these are also zero. 
But ``dependent" $2 \times 2$ minors (i.e., those resulting from nearly proportional rows) represent a new feature. 
This filtering allows us to highlight genuinely ``2-body," ``3-body," $\ldots$ behavior.
In the ideal KA case, the presence of $(m-n)$ zero rows has a particularly acute effect for the top $n \times n$ minors of the Jacobian: There is (at most) a single nonzero $n \times n$ minor of $J_{j i}({\bf x})$ for each ${\bf x}$.

Let us pause here to say more about our leading actors,
the $k \times k$ minors of the Jacobian $J_{ji}$ of the inner map $\sigma(\mathbf{z})$ (the analog of $\Phi$ in the model we study).
Minors are rarely discussed in numerical linear algebra because, unlike measures such as the SVD, they are not rotationally invariant---they depend on a choice of basis.
But the linear algebra of neural networks allows them to shine, since the vector spaces of each layer come with neurons as their preferred bases. 
(Our surprise is not that they are important, but, in view of their naturality in the context of neural networks, that they have previously been so little investigated.)
By definition, $J_{ji}$ is the linearization of the inner map $\sigma(\mathbf{z})$.
For any linear map $L$, its size-$k$ minors tell you how $L$ maps (oriented) $k$-planes.
Thus, the size-$k$ minors of $J_{ji}$ say how the inner map transforms input $k$-planes into the hidden dimension, with respect to the neuron bases. 

Another description of KA geometry is minor concentration \cite{2024arXiv241008451F}.
This has two aspects: lots of (nearly) zero minors after training and a few, very large (in absolute value) ones.
Minor concentration has two sources: ``alignment" and ``scale."
Alignment refers to whether the $k$-th exterior power of the Jacobian (in standard neuron coordinates) $\mathbb{R}^n \to \mathbb{R}^m$ ``spreads'' the mass of the image coordinate $k$-plane out into a superposition of $\binom{m}{k}$ possibilities or instead concentrates the image in only a few of these.
Intuitively, alignment says which groups of $k$-neurons in one layer are attempting to influence the collective state of $k$ neurons in the next layer.
One can invent tales \cite{2024arXiv241008451F} where this behavior could signal the synthesis of a new abstraction in one layer from earlier abstractions deduced within the previous layer.

Scale refers to the $k$-volume of an image $k$-plane.
If a linear map $L$ is replaced by $cL$, with $c$ a constant, the $k$-th exterior power $L^{(k)}$ is replaced by $c^kL^{(k)}$, and the entries (minors) are all scaled by $c^k$.
(Scaling, of course, will usually not be uniform, as in this simplest example.)
The role of scale in our data was initially a surprise to us, but, in retrospect, is part and parcel of the KA theorem.
In \eqref{kadef}, the $\phi_{ij}$ together produce an embedding $\Phi$ of the data into the hidden layer.
In Kolmogorov's original proof (and in most subsequent works), $\Phi$ is only H\"older continuous.
It was shown in \cite{Vitushkin1964, vitushkin1967} that no universal $\Phi$ can be $C^1$.
Like Brownian motion, typical H\"older functions are nowhere differentiable, suggesting that, to the extent that the maps we train into our hidden layer reflect the geometry of the KA theorem, we should expect their derivatives to blow up as training proceeds \cite{hayou2024a}.
This can be seen in Fig.~\ref{fig:jacobian_matrices}: After training the map is brighter---that part is scale.
The horizontal bars of the heat map---that they are sharp and not vertically diffused---is the result of coordinate alignment.
Important work, both classical and recent \cite{fridman1967improvement, 2017arXiv171208286A}, has led to improved universal inner functions $\phi_{ij}$, which are Lipschitz continuous (and by scaling one may take $\text{Lip}=1$).
Roughly speaking they follow a load balancing strategy: careful tuning achieves tamer inner functions $\phi_{ij}$ at the expense of wilder outer functions $g_j$.
In view of these results one might question if our observed divergence of certain Jacobian entries is truly dictated by the KA analogy.
We think it is for two reasons.
The first is that the additional subtleties in making the universal $\phi_{ij}$ Lipschitz would appear to make it harder for a training algorithm to stumble upon them.
A second reason, specific to our single hidden layer model, is that our outer map is forced to be linear: It cannot share any ``wild'' behavior.

\subsection{What It Is Not}
\label{whatnotsection}

KA geometry is different from the familiar weight matrix sparsity (e.g., \cite{10.5555/2969830.2969903, frankle2018lottery}), since it's not directly correlated with the raw weight values.
Instead it reflects a fine-scale texture of the first-layer map, a sort of patterning on a weight ``substrate."
During training there can and should be macroscopic changes in weights.

For example, token embeddings in language models must evolve inside a modest (hundreds, usually) dimensional ``token embedding space'' so that their inner products grow to reflect myriad pair-wise and higher correlations between words and word fragments.
It is widely observed that during training both coarse and fine structure have jobs to do and must evolve for successful learning.
Models often work well when ``quantized'' (i.e., have their weights truncated to fewer bits) after training but not during.
From this we learn that details---texture---matter for learning.
Fine detail seems to make gradient solvers (and their more sophisticated descendants such as Adam) hum.
(There is, of course, some scale, sufficiently fine, beyond which detail is truly irrelevant; gradients and Hessians are only abstractions, albeit quite useful ones.
We do not know the theory for determining this ``too-fine'' scale in advance.
So here ``fine'' refers to the smaller scales which are still large enough to be relevant for training.)

The KA inner functions, $\phi_{ij}$, are far from unique: It is the textures of the first-layer map (and, more generally, we can consider patterning in other maps between layers $i$ and $j$, $i<j$, and $j$ not the output layer, in deeper models), rather than its macroscopic values that are important.
In fact, a central point of the KA proof is that, mathematically, the ``good'' $\phi_{ij}$ are ``second category'' in the space of continuous functions with sup-norm topology, i.e., they are generic.
This immediately tells us that KA can only be part of any learning story: Large scale structure
is not the business of KA.

To reiterate, learning through training involves both gross rearrangements and fine-scale patterning of data.
Our discussion is at the level of Jacobians and so is blind to all large scale information.
The surprise is the coupling of the two: the fine-scale KA patterning appears (in many cases) to be essential to Adam finding useful large scale moves.
In fact, we've seen hints of this: If a term is added to the loss function to discourage the emergence of KA-patterning during training, it can prevent learning.
What about the exciting direction: reverse the sign of that term in the loss function to see if learning accelerates?
Not surprisingly it is easier to break things than to improve them, and, indeed, experiments suggest that considerable care will be required to push models in a positive direction---this avenue is in our queue for future work.

\subsection{Measuring It}
\label{measuringsection}

We study KA geometry through statistical analysis of ensembles of Jacobian minors across input samples.
For each input point $\mathbf{x}$, the Jacobian $J_{ji}(\mathbf{x})$ of the inner map $\sigma(\mathbf{z})$ provides a sample of the local geometry.
By aggregating these samples across training data and comparing trained models against their initialization, we can identify statistically significant signatures of KA geometry.

What distinguishes KA geometry statistically?
The key signatures are zero rows and minor concentration.
Zero rows---where entire rows of the Jacobian or its $k$-th exterior power vanish (or nearly so)---reflect the locally constant coordinates central to the KA construction.
Minor concentration manifests as simultaneous peakedness near zero and heavy tails (large outliers) in the distributions of $k \times k$ minors.
Together, these behaviors reflect the interplay of alignment and scale discussed earlier.
Moreover, these patterns intensify with increasing minor size $k$ and correlate with model performance, suggesting they are an integral part of learning.
We have very large datasets, so even small shifts in our observables cannot be due to chance.

The training effect on minors of the model trained on the \texttt{xor} function \eqref{xorfunction}, shown in Figs.~\ref{rankvalueplots} and \ref{minordistributioncomparison}, is visually a true embarrassment  of riches and pauperizes our effort to quantify it numerically.
In attempting to do so, we selected four observables to quantify KA geometry:
\begin{enumerate}
\item[A] \textbf{Zero Rows} \eqref{zerorowdef}: The fraction of rows in minor matrices (absolute value of the $k$-th exterior power of the Jacobian) falling below thresholds based on initial row mean quantiles.
This captures the emergence of $\mathbf{x}$-dependent inactive coordinates, a hallmark of ideal KA geometry.

\item[B (i)] \textbf{Participation Ratio} \eqref{participationratiodef}: The ratio of $L_1$ to $L_2$ norms of minor matrices, $\text{PR}^{(k)} = \|J^{(k)}\|_1 / \|J^{(k)}\|_2$.
Lower values indicate greater concentration, with a few large minors dominating the distribution while most remain near zero.

\item[B (ii)] \textbf{Random Rotation Ratio} \eqref{randomrotationratio}: The ratio of the actual maximum minor to those obtained after rotating Jacobian columns by a random orthogonal matrix.
Values exceeding unity indicate that large minors arise from structured alignment rather than chance.

\item[B (iii)] \textbf{Column Divergence} \eqref{kldef}: KL divergence (cross-entropy) between trained and initial minor matrix columns, viewed as probability distributions.
This measures both the degree to which training alters the model's minors from their initial values and, given the Shannon entropy of the model's minors at initialization, how training affects the alignment of the trained inner map.
\end{enumerate}

Beyond static comparisons between initial and trained models, we track these metrics dynamically through two experimental designs.
Evolution experiments monitor KA geometry as training progresses, while interpolation experiments examine models trained on functions of varying difficulty (parameterized from learnable to unlearnable).
In both cases, we use $R^2$ \eqref{r2def} as a common coordinate---essentially tracking how KA geometry covaries with model performance.
This shows whether KA geometry emerges gradually or suddenly, and whether it degrades smoothly or catastrophically as tasks become impossible.

These metrics reveal a consistent pattern: for simple (\texttt{linear}) or unlearnable (\texttt{random}) functions, trained models show statistics similar to their initial values, when training near the critical batch size; see the discussion in \S \ref{setupsection}.
However, for learnable nonlinear functions (\texttt{xor}) in an appropriate complexity regime---what we call the ``Goldilocks regime"---training induces statistically significant KA geometry.
The following sections present detailed evidence across function types and model architectures.

\section{Setup}
\label{setupsection}

We considered several different functions $f: [-1, 1]^n \rightarrow \mathbb{R}$ to learn.
We will present results for the \texttt{xor}, \texttt{linear}, and \texttt{random} functions:
\begin{align}
\label{xorfunction}
\texttt{xor}({\bf x}) & = \prod_{i=1}^n \sin \big(\pi x_i \big), \\
\label{linearfunction}
\texttt{linear}({\bf x}) & = {\bf x} \cdot {\bf c} + c_0, \\
\label{randomfunction}
\texttt{random}({\bf x}) & = Y.
\end{align}
Here, ${\bf c} = (c_1, \ldots, c_n)$ and $c_0$ are fixed random coefficients chosen uniformly from $[-1,1]$, and $Y \sim {\rm Uniform}(-1,1)$.
\texttt{linear} and \texttt{xor} represent ``easy" and ``hard" learnable functions; \texttt{random} is unlearnable.
\texttt{xor} is a continuous version of the discrete (Boolean) XOR, a classic function for which a 1-hidden layer neural network (with nontrivial nonlinearity) is required for accurate approximation \cite{minsky1988perceptrons}.

All experiments in this paper were done on 1-hidden layer fully-connected neural networks (MLPs), with model function $f_{\rm MLP}$ given in \eqref{modeldef}, using a CPU.
The input dimension $n=3$ was fixed and we considered hidden dimensions $m \in \{4, 8, 16, 32 \}$.
We trained 4 models for each function type and hidden dimension $m$.
Model weights were Kaiming normal initialized (i.i.d.~$A_{i j} \sim {\cal N}(0, 1/n)$ and $B_j \sim {\cal N}(0, 1/m)$) with the same model seeds across all function types and hidden dimensions; the biases $a$ and $b$ were initially zero.
This means the initializations of all models of a fixed hidden dimension are identical across target function types.
We used the mean-squared error (MSE) loss and the Adam optimizer, with standard beta values $(\beta_1, \beta_2) = (0.9, 0.999)$ and a learning rate equal to $0.01$.
Models were trained on a set of $1000$ examples $\{ \big( \mathbf{x}, f(\mathbf{x}) \big) \}$ for at most $10,000$ epochs.
We allowed early-stopping with a patience parameter equal to $50$ epochs.

In training, we used a fixed batch size given by the (approximate) critical batch size \cite{2018arXiv181206162M} of the \texttt{xor}$(m)$ model.
By ``\texttt{xor}$(m)$ model," we mean a model with $m$ hidden neurons, trained on the \texttt{xor} function.
We will discuss the effect of batch size as a variable in Appendix \ref{batchsizeappendix}; until then, all our assertions pertain to experiments near critical batch size.
The critical batch size $B_c$ can be defined as 
\begin{align}
\label{criticalbatchdef}
B_c = E_{\rm min}/S_{\rm min},
\end{align} 
where $E_{\rm min}$ is the minimal number of examples the network needs to process to achieve some criterion and $S_{\rm min}$ is the minimal number of optimization steps needed to achieve that same criterion.
In practice, we determine $E_{\rm min}$ using training runs with a batch size equal to $64$ and $S_{\rm min}$ with training runs with a batch size equal to $1000$, finding $B_c = 250$.
(This is the largest $B_c$ of the \texttt{xor}$(m)$ model across $m \in \{4, 8, 16, 32 \}$.)
While our small experiments are easily carried out using a full batch, we believe the critical batch size is motivated by large experiments where the compromise between compute and time cannot be ignored.

We didn't try to optimize this training setup.
Our use of Adam, rather than, say, vanilla gradient descent, was only motivated by faster training.
We did not use dropout or any explicit regularization.

Model performance was measured by (the coefficient of determination)
\begin{align}
\label{r2def}
R^2 = 1 - {\sum_{I=1}^B \big( y_I - \hat y_I \big)^2 \over \sum_{J=1}^B \big( y_J - \bar y \big)^2},
\end{align} 
where $y_I = f({\bf x}_I)$, $\hat y_I = f_{\rm MLP}({\bf x}_I)$, $\bar y = {1 \over B} \sum_I y_I$, and $B=1000$.
Given our use of MSE, this is a rescaled loss that allows the comparison of different models.
\begin{table}[htbp]
\centering
\renewcommand{\arraystretch}{1.3}
\begin{tabular}{|l@{\hspace{0.5cm}}|c@{\hspace{0.5cm}}|c@{\hspace{0.5cm}}|c@{\hspace{0.5cm}}|c|}
\hline
Function & $m=4$ & $m=8$ & $m=16$ & $m=32$ \\
\hline
\texttt{xor} & $0.3819 \pm 0.1286$ & $0.6864 \pm 0.0163$ & $0.9412 \pm 0.0269$ & $0.9850 \pm 0.0081$ \\
\hline
\texttt{linear} & $1.0000 \pm 0.0000$ & $1.0000 \pm 0.0000$ & $1.0000 \pm 0.0000$ & $1.0000 \pm 0.0000$ \\
\hline
\texttt{random} & $0.0042 \pm 0.0009$ & $0.0067 \pm 0.0031$ & $0.0102 \pm 0.0054$ & $0.0210 \pm 0.0029$ \\
\hline
\end{tabular}
\caption{Training $R^2$ (mean $\pm$ standard deviation) for \texttt{xor}$(m)$, \texttt{linear}$(m)$, and \texttt{random}$(m)$ models.
}
\label{train_r2_table}
\end{table}
Table \ref{train_r2_table} shows how the training $R^2$ varies with the number of hidden neurons $m$ in \texttt{xor}$(m)$, \texttt{linear}$(m)$, and \texttt{random}$(m)$; the model's performance on test data is indistinguishable (i.e., no ``overfitting" or grokking behavior \cite{2022arXiv220102177P}) and not displayed.
Performance steadily increases with $m$ for the \texttt{xor}$(m)$ model; 
unsurprisingly, the models perfectly learn \texttt{linear} and fail to memorize \texttt{random}.

\section{Searching for KA Geometry}

We study KA geometry with the size-$k$ minor matrices, $J^{(k)}$, defined as the absolute value of the $k$-th exterior power of the Jacobian.
$J^{(k)}$ is a ${{m}\choose{k}} \times {{n}\choose{k}}$ matrix, whose rows/columns are labeled by $k$-tuples $\mathbf{i} = (i, j, \ldots)$ of row/column indices $i < j < \ldots$ of the Jacobian, with elements equal to the absolute value of the determinant of the indicated $k \times k$ submatrix.
$J^{(1)}$ is the elementwise absolute value of the Jacobian \eqref{jacobianformula};
the size-$2$ minor matrix $J^{(2)}$, for example, is 
\begin{align}
J^{(2)}_{\mathbf{i} \mathbf{i'}}({\bf x}) = J^{(2)}_{(i,j), (i', j')}({\bf x}) = \Big| \det\begin{pmatrix} J_{i i'}({\bf x}) & J_{i j'}({\bf x}) \cr J_{j i'}({\bf x}) & J_{j j'}({\bf x}) \end{pmatrix} \Big|,
\end{align}
for $i, j \in \{1, \ldots, m\}$, $i', j' \in \{1, \ldots, n \}$, with $i < j$ and $i' < j'$.
We will typically refer to minors and the absolute value of minors interchangeably.
Why absolute value? 
We have not observed meaningful asymmetry when looking at signed minors.

The notation $J^{(1)}$ may seem excessive since these are merely the absolute values of the entries of the Jacobian.
We use this notation to help knit together a continuous story connecting the familiar phenomenon of sparsification \cite{10.5555/2969830.2969903, frankle2018lottery} with the new phenomenon of minor concentration.
$J^{(k)}$ form a family of observable (parameter $k$) which, in the domain of interest---our ``Goldilocks regime"---seems to have an increasing signal as $k$ grows toward $n$.
Classically, sparsification is generally discussed at the level of weight matrices and here it is discussed on the level of the Jacobian; as seen in \eqref{jacobianformula}, the two are directly correlated when Jacobian sparsity is independent of $\mathbf{x}$.

\subsection{Zero Rows}
\label{zerorowsection}

\begin{figure}[h]
\centering
\begin{subfigure}[b]{0.32\textwidth}
    \centering
    \includegraphics[width=\textwidth]{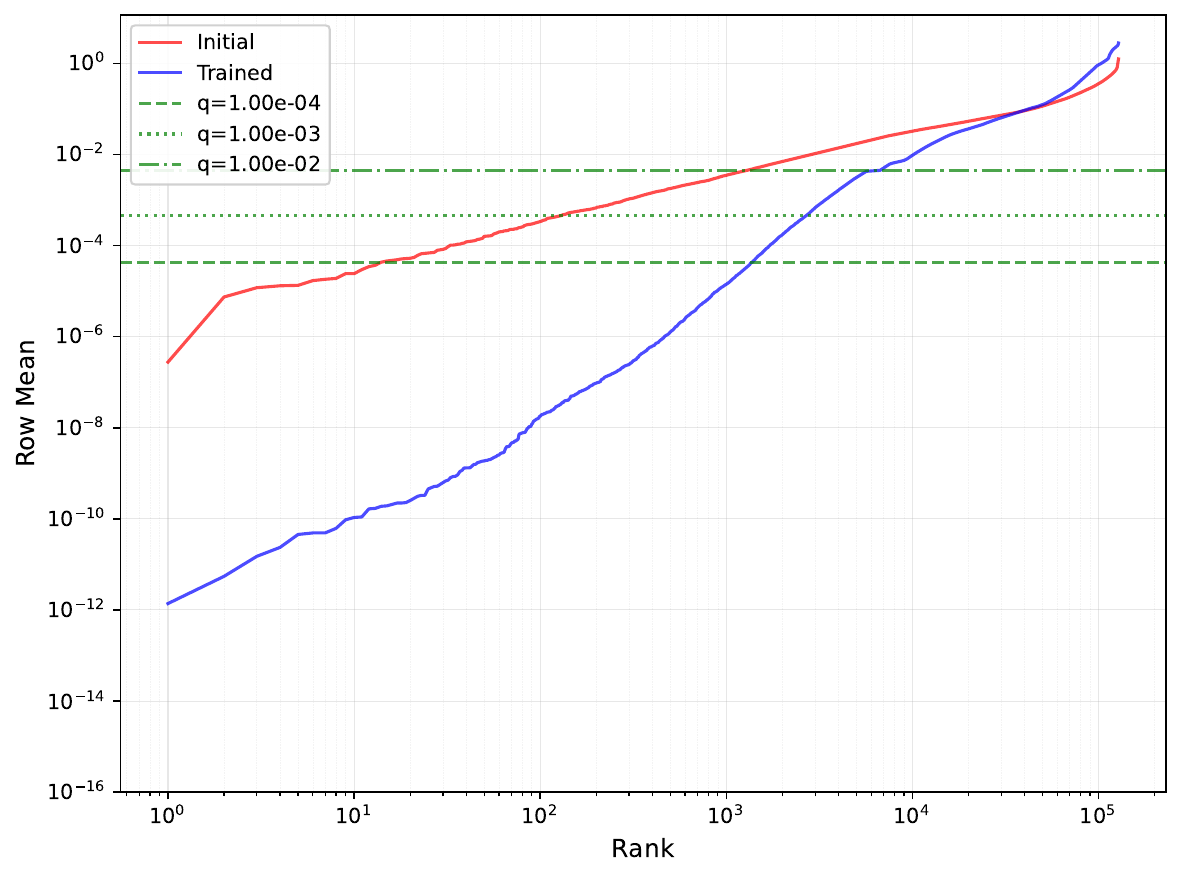}
    \caption{$k=1$}
    \label{k1rowmeans}
\end{subfigure}
\hfill
\begin{subfigure}[b]{0.32\textwidth}
    \centering
    \includegraphics[width=\textwidth]{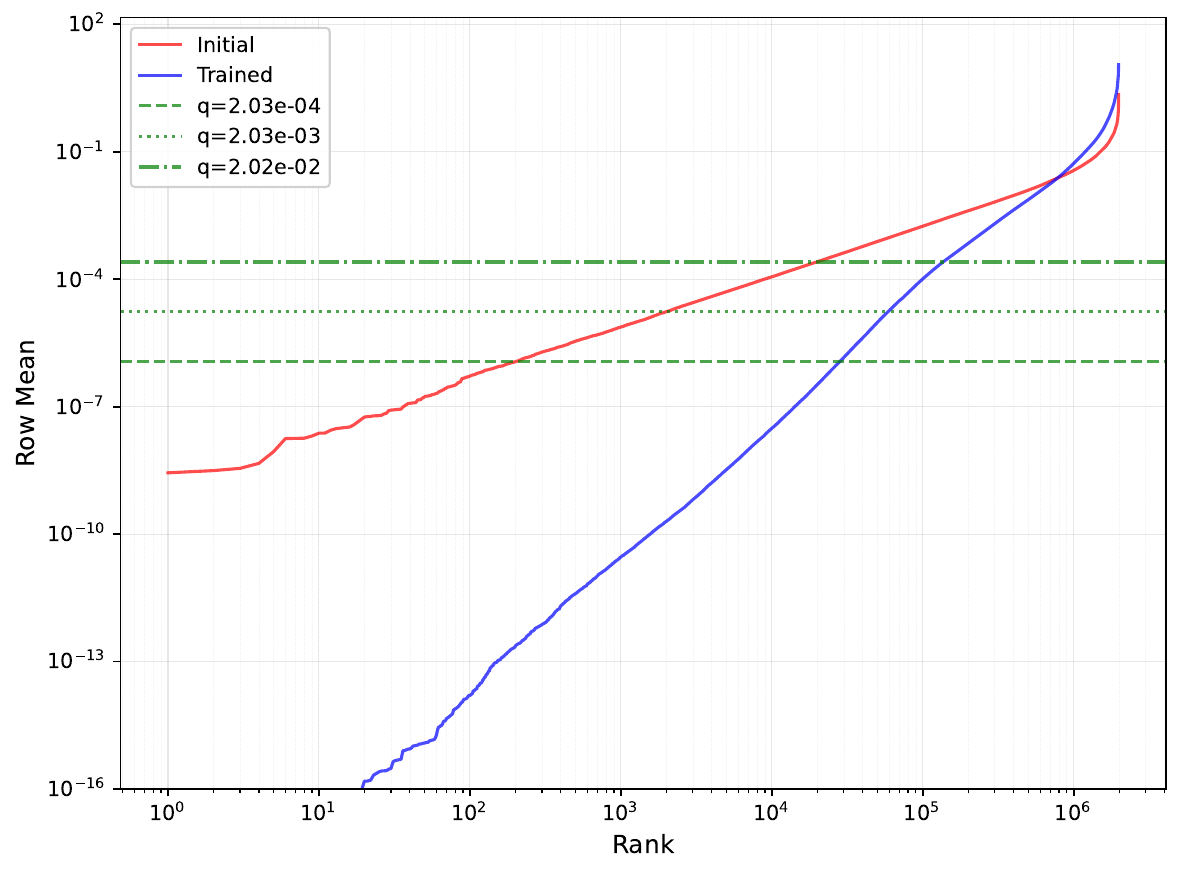}
    \caption{$k=2$}
    \label{k2rowmeans}
\end{subfigure}
\hfill
\begin{subfigure}[b]{0.32\textwidth}
    \centering
    \includegraphics[width=\textwidth]{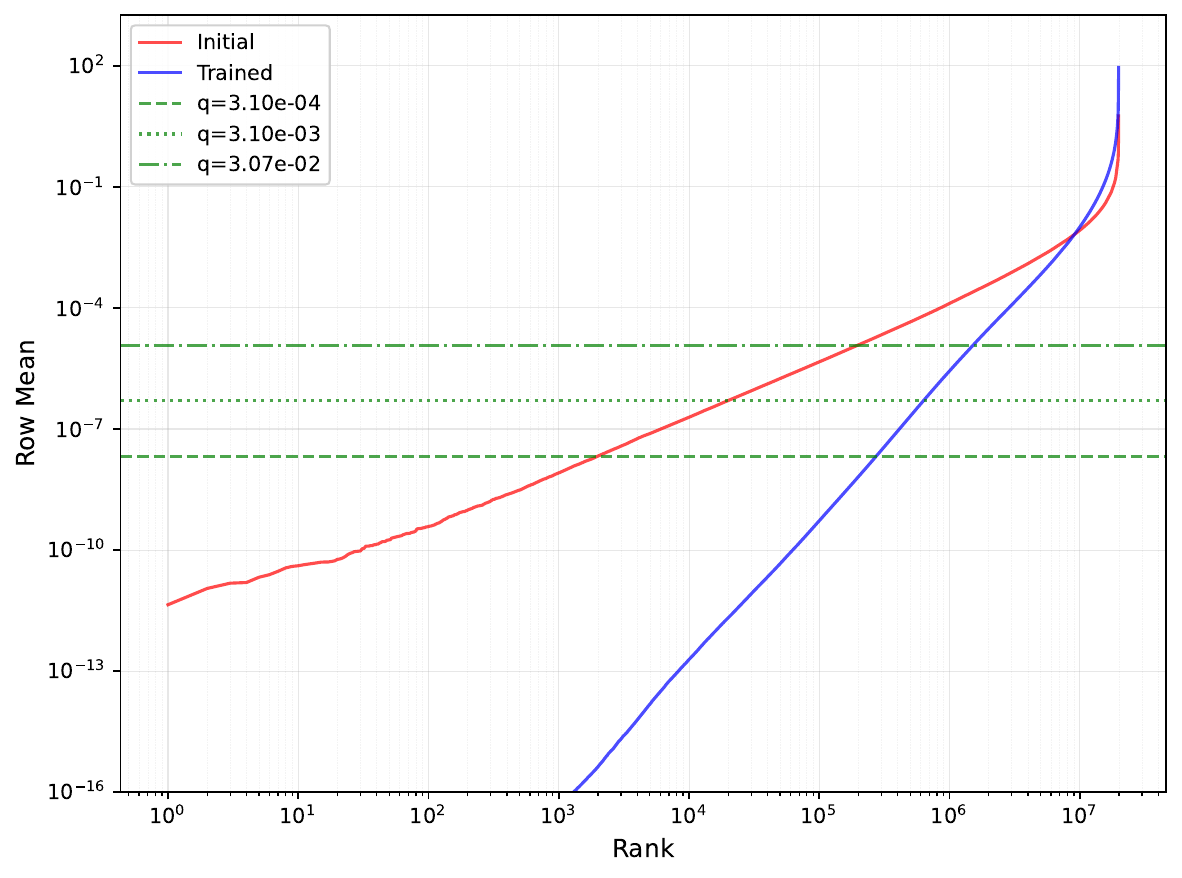}
    \caption{$k=3$}
    \label{k3rowmeans}
\end{subfigure}
\caption{
Row means of size-$k$ minor matrices of the trained and initial \texttt{xor}$(32)$ model, ranked in ascending order.
Horizontal lines mark row mean quantiles of the model at initalization.
Size-$3$ row means can approach machine precision (not shown with the $10^{-16}$ lower cutoff). % of the $y$-axes).
}
\label{rankvalueplots}
\end{figure}

Let us begin our search for KA geometry by looking for zero rows in the size-$k$ minor matrices of trained models.
For this we will use the size-$k$ minor matrix row means $\bar \rho^{(k)}_{\mathbf{j}}(\mathbf{x})$:
\begin{align}
\label{rowmeandef}
\bar \rho^{(k)}_{\mathbf{j}}(\mathbf{x}) = \binom{n}{k}^{-1} \sum_{\mathbf{i}} J^{(k)}_{\mathbf{j} \mathbf{i}} (\mathbf{x}).
\end{align}
Fig.~\ref{rankvalueplots} shows Zipf plots (i.e., rank-value plots) of the \texttt{xor}$(32)$ model's row means after training and at initialization.
The row means at initialization have a fairly consistent rise over roughly the first $50\%$ of ranks, before crossing over to a distinct high-rank behavior.
In contrast, the trained model exhibits a novel low-rank regime that reflects the emergence zero rows in $J^{(k)}$.

We say that a row in the size-$k$ minor matrix is zero when its mean \eqref{rowmeandef} is below the $q^{(k)}$-th quantile $Q^{(k)}$ of minor matrix row means at initialization:
\begin{align}
\label{zerorowdef}
\bar \rho^{(k)}_{\mathbf{j}}(\mathbf{x}) < Q^{(k)}.
\end{align}
(For $q \in (0, 1)$, the $q$-th quantile of a distribution is $Q(q) = F^{-1}(q)$, where $F$ is the cumulative distribution function.)
Since the novel low-rank regime is absent before training (Fig.~\ref{rankvalueplots}), we can think of $q^{(k)}$ as a ``false-positive rate" for zero row identification in the model at initialization.
This means that we should only consider there to be a statistically significant number of zero rows in the trained model when the fraction of zero rows sufficiently exceeds $q^{(k)}$.

In the size-$1$ plot in Fig.~\ref{k1rowmeans}, we consider $q^{(1)} \in \{0.0001, 0.001, .01\}$.
Having made a choice for $q^{(1)}$, we can then use \eqref{zerorowdef} to identify zero rows in the size-$1$ minor matrix.
What should we take for the ``false-positive rates" $q^{(k)}$ for minor sizes $k > 1$?
One option is to take $q^{(k)}$ to be the same for all $k$. 
This would miss the fact (in the ideal situation) that an identically zero row $j$ in a size-$1$ minor matrix causes all rows $\mathbf{j} = (\ldots, j, \ldots)$, labeled by tuples containing index $j$, in the size-$k$ minor matrix to be zero.
If the fraction $q^{(1)}$ of rows in the initial size-$1$ minor matrix are treated as being identically zero, then a generally different fraction of zero rows will be induced in the size-$k$ minor matrix.
We take $q^{(k)}$ to be this fraction:
\begin{align}
q^{(k)} = 1 - { \binom{m(1 - q^{(1)})}{k} \over \binom{m}{k}} \approx k \cdot q^{(1)}, \quad k > 1.
\end{align}
Above, $\binom{m}{k}$ is the total number of rows in the size-$k$ minor matrix;
$\binom{m(1 - q^{(1)})}{k}$ is the number of nonzero rows in the size-$k$ minor matrix, assuming the fraction $q^{(1)}$ of rows ``mis-classified" as zero in the initial Jacobian are strictly zero.
(For noninteger argument, the factorial is defined in terms of the Gamma function: $x! = \Gamma(x + 1)$.)
These higher-$k$ ``false-positive rates" $q^{(k)}$ are indicated in the $k=2$ and $k=3$ panels in Figs.~\ref{k2rowmeans} and \ref{k3rowmeans}.
This definition of $Q^{(k)}\big(q^{(k)}\big)$ fulfills our introductory promise to account for zero rows in the higher exterior powers of the Jacobian in a hierarchical manner that separates true $2$- and $3$-body effects from the formal consequences of $1$-body statistics.

\begin{figure}[h]
\centering
\begin{subfigure}[b]{0.32\textwidth}
    \centering
    \includegraphics[width=\textwidth]{rankvalue_k3_xor_h32_train}
    \caption{\texttt{xor}$(32)$}
    \label{k3xorrowmeans}
\end{subfigure}
\hfill
\begin{subfigure}[b]{0.32\textwidth}
    \centering
    \includegraphics[width=\textwidth]{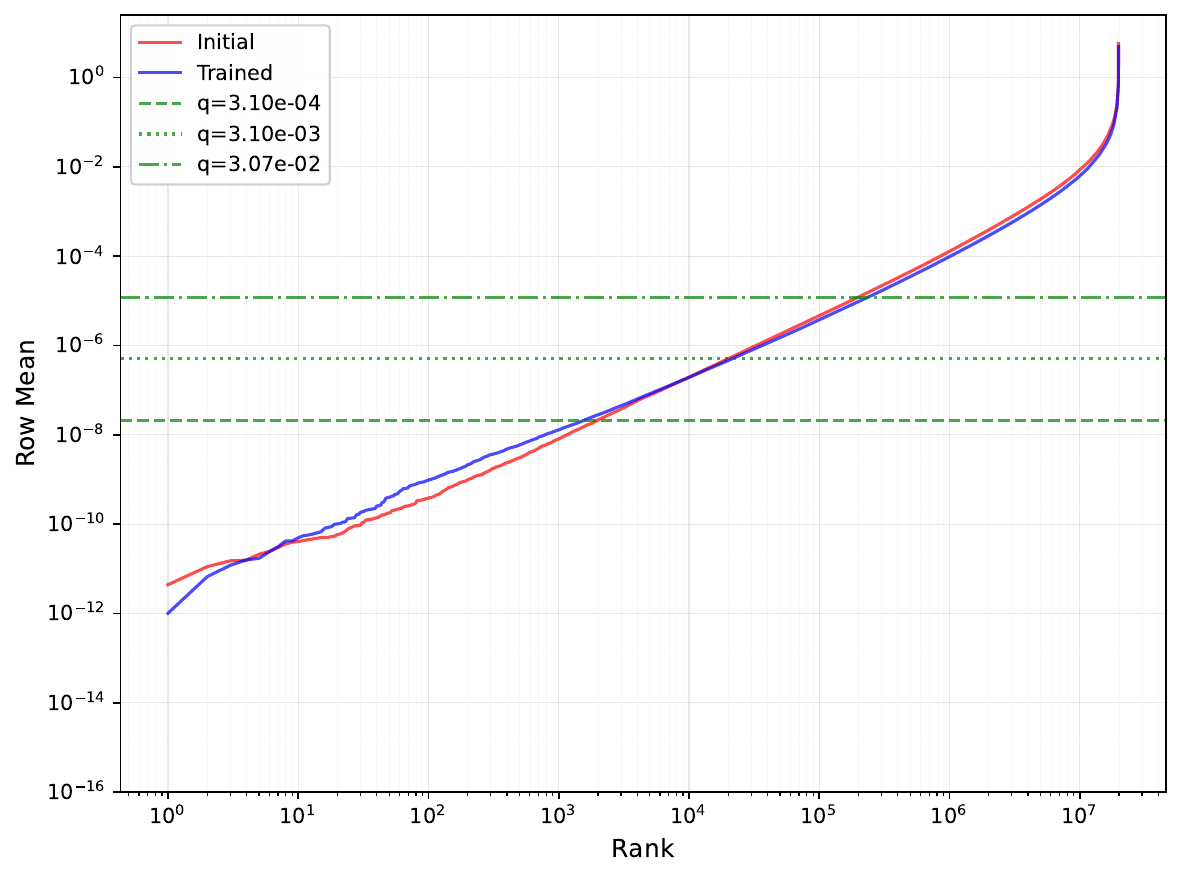}
    \caption{\texttt{linear}$(32)$}
    \label{k3linearrowmeans}
\end{subfigure}
\hfill
\begin{subfigure}[b]{0.32\textwidth}
    \centering
    \includegraphics[width=\textwidth]{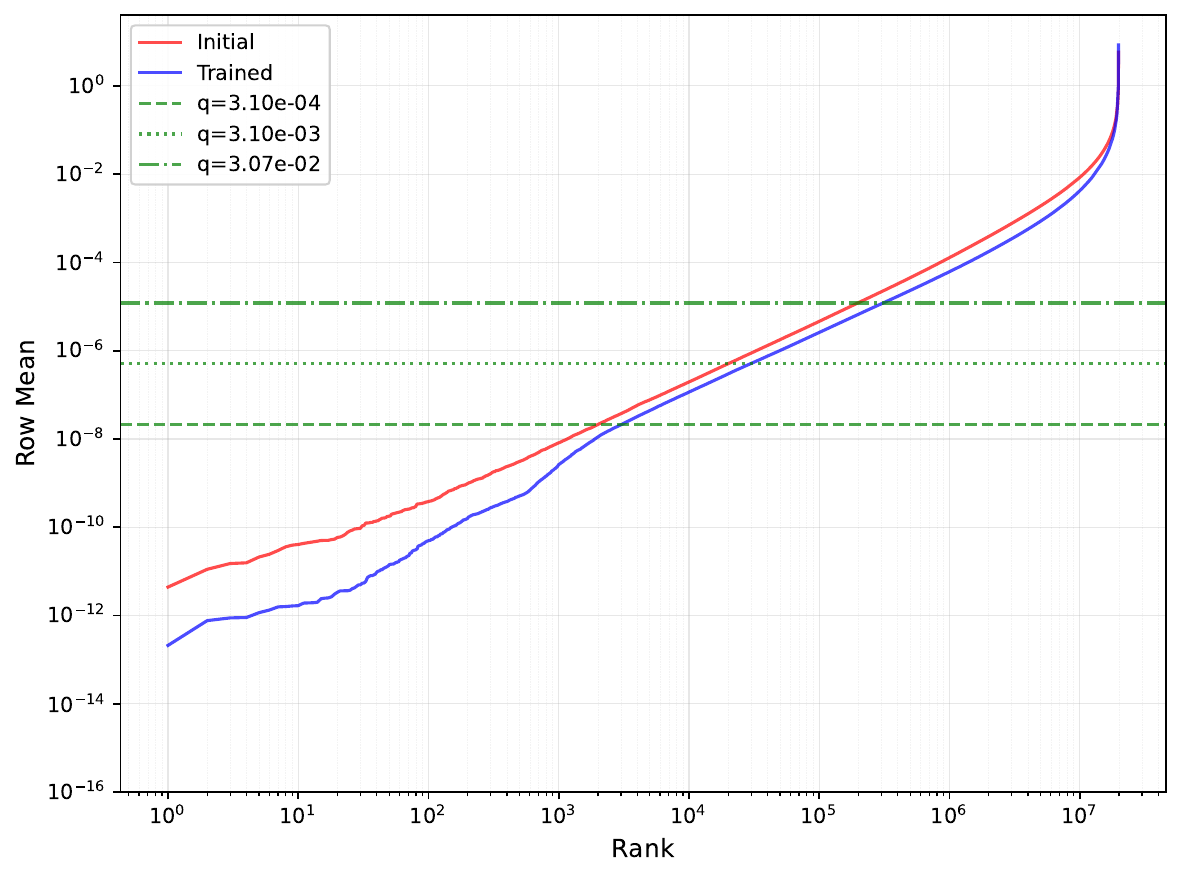}
    \caption{\texttt{random}$(32)$}
    \label{k3randomrowmeans}
\end{subfigure}
\caption{
Size-$3$ row mean comparison across target functions.
}
\label{rankvaluefunctioncomparison}
\end{figure}
Fig.~\ref{rankvaluefunctioncomparison} compares the size-$3$ row means for models trained on the \texttt{xor} (\ref{k3xorrowmeans}), \texttt{linear} (\ref{k3linearrowmeans}), and \texttt{random} (\ref{k3randomrowmeans}) functions. 
Plots for minor sizes $k=1$ and $k=2$ have broadly similar features, though less well developed.
We observe a progression from the \texttt{linear}$(32)$ model,  which closely tracks the initial row means, through the \texttt{random}$(32)$ model, which exhibits some deviation from the initial row means, to the \texttt{xor}$(32)$ model, for which a noticeable low-rank regime appears.
Table \ref{zerorowtable} gives the percentages of zero rows \eqref{zerorowdef} in the size-$3$ minor matrix in these models for $q^{(1)} \in \{0.0001, 0.001, .01\}$.
As the Zipf plots suggest, the \texttt{xor}$(32)$ model exhibits a significant number of zero rows. 
On the other hand, the zero row counts of the \texttt{linear}$(32)$ and \texttt{random}$(32)$ models closely track the ``false-positive rates" $q^{(k)}$.
The percentages of zero rows observed in the \texttt{xor}$(32)$ model are well below that of the KA ideal, in which we'd have, for example, a single nonzero row on average in the size-$3$ minor matrix.
The point (for this study) is not the specific values observed but rather that {\it any} statistically significant percentages of zero rows are found.
\begin{table}[htbp]
\centering
\renewcommand{\arraystretch}{1.3}
\begin{tabular}{|l@{\hspace{0.5cm}}|c@{\hspace{0.5cm}}|c@{\hspace{0.5cm}}|c|}
\hline
 & $q^{(1)} = 0.01\%$ & $q^{(1)} = 0.1\%$ & $q^{(1)} = 1.0\%$ \\
\hline
\multicolumn{4}{|c|}{\texttt{xor}$(32)$} \\
\hline
Zero (\%) & $2.068 \pm 0.651$ & $4.822 \pm 1.390$ & $12.339 \pm 3.012$ \\
Dependent$(2)$ (\%) & $0.311 \pm 0.079$ & $0.771 \pm 0.172$ & $2.360 \pm 0.425$ \\
Dependent$(3)$ (\%) & $0.143 \pm 0.046$ & $0.445 \pm 0.120$ & $1.116 \pm 0.192$ \\
\hline
\multicolumn{4}{|c|}{\texttt{linear}$(32)$} \\
\hline
Zero (\%) & $0.029 \pm 0.003$ & $0.361 \pm 0.035$ & $3.671 \pm 0.256$ \\
Dependent$(2)$ (\%) & $0.013 \pm 0.002$ & $0.117 \pm 0.012$ & $1.046 \pm 0.089$ \\
Dependent$(3)$ (\%) & $0.014 \pm 0.003$ & $0.149 \pm 0.022$ & $1.127 \pm 0.073$ \\
\hline
\multicolumn{4}{|c|}{\texttt{random}$(32)$} \\
\hline
Zero (\%) & $0.045 \pm 0.004$ & $0.473 \pm 0.019$ & $4.947 \pm 0.152$ \\
Dependent$(2)$ (\%) & $0.013 \pm 0.001$ & $0.147 \pm 0.015$ & $1.526 \pm 0.108$ \\
Dependent$(3)$ (\%) & $0.024 \pm 0.004$ & $0.208 \pm 0.013$ & $1.581 \pm 0.077$ \\
\hline
\end{tabular}
\caption{Size-$3$ zero row and size-$2$/size-$3$ dependent zero row percentages in \texttt{xor}$(32)$, \texttt{linear}$(32)$, and \texttt{random}$(32)$ models across 4 seeds (mean $\pm$ standard deviation).
}
\label{zerorowtable}
\end{table}

We say that a row $\mathbf{j}$ is consistently zero if \eqref{zerorowdef} is satisfied across $99\%$ of examples $\mathbf{x}$.
(When aggregating over multiple model seeds, this needs to be computed for each model seed independently since there is no reason a particular row $\mathbf{j}$ should be zero across arbitrary initializations.)
Rows that are consistently zero across multiple examples must be a direct result of the (structured) sparsification of the weight matrix (recall formula \eqref{jacobianformula} for the Jacobian).
(For ${\rm GeLU}$ activation, it's possible for consistently zero rows to arise from large in magnitude negative biases---see \eqref{jacobianformula}. 
This has not been the cause of the consistently zero rows in the models that we've studied.) 
For training that uses the critical batch size \eqref{criticalbatchdef}, presented here, we find no appreciable consistently zero rows; consistently zero rows can appear as the batch size is varied (see Appendix \ref{batchsizeappendix}).
This means that essentially all zero rows are due to KA-like patterning of the Jacobian (i.e., vanishing $\sigma'$ in \eqref{jacobianformula}).

The second zero row characterization is that of a size-$k$ dependent zero row for $k > 1$.
As the name suggests, any such row in a size-$k$ minor matrix is zero due to linear dependence of rows in the size-$(k-1)$ minor matrix.
A zero row $(i, j)$ is a size-$2$ dependent zero row, if neither $i$ nor $j$ index a zero row in the size-$1$ minor matrix.
Likewise, a zero row $(i, j, l)$ is a size-$3$ dependent zero row, if no subset $\{ (i, k), (i, l), (k, l) \}$ index zero rows in the size-$2$ minor matrix.
Table \ref{zerorowtable} gives the percentages of dependent zeros. 
These ratios are quite small, indicating that most, but importantly, not all zero rows in higher-$k$ minor matrices result from zeros in the Jacobian.

\subsection{Minor Concentration}
\label{participationratiosection}

\begin{figure}[h]
\centering
\begin{subfigure}[b]{0.32\textwidth}
    \centering
    \includegraphics[width=\textwidth]{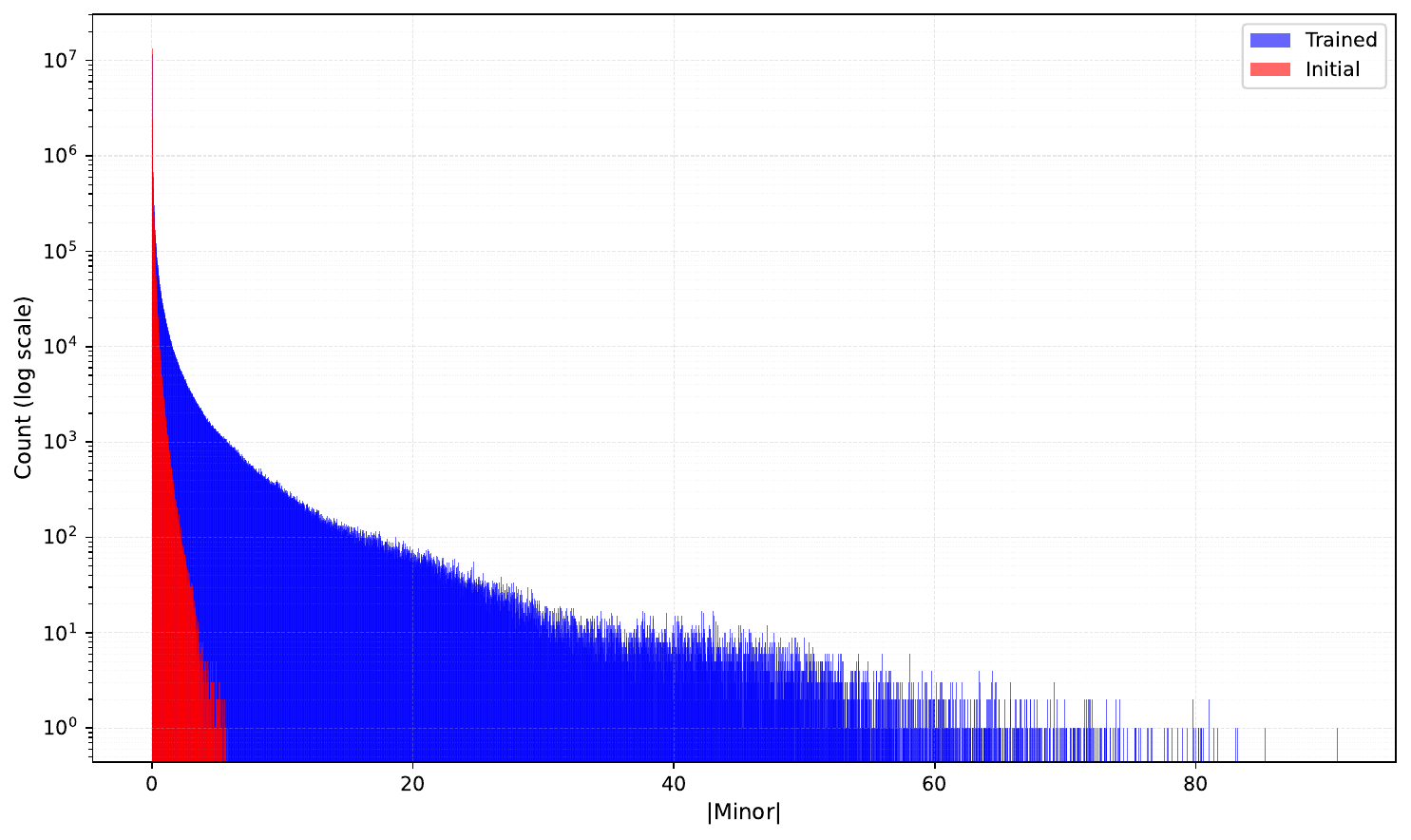}
    \caption{\texttt{xor}$(32)$}
    \label{fig:panel_a}
\end{subfigure}
\hfill
\begin{subfigure}[b]{0.32\textwidth}
    \centering
    \includegraphics[width=\textwidth]{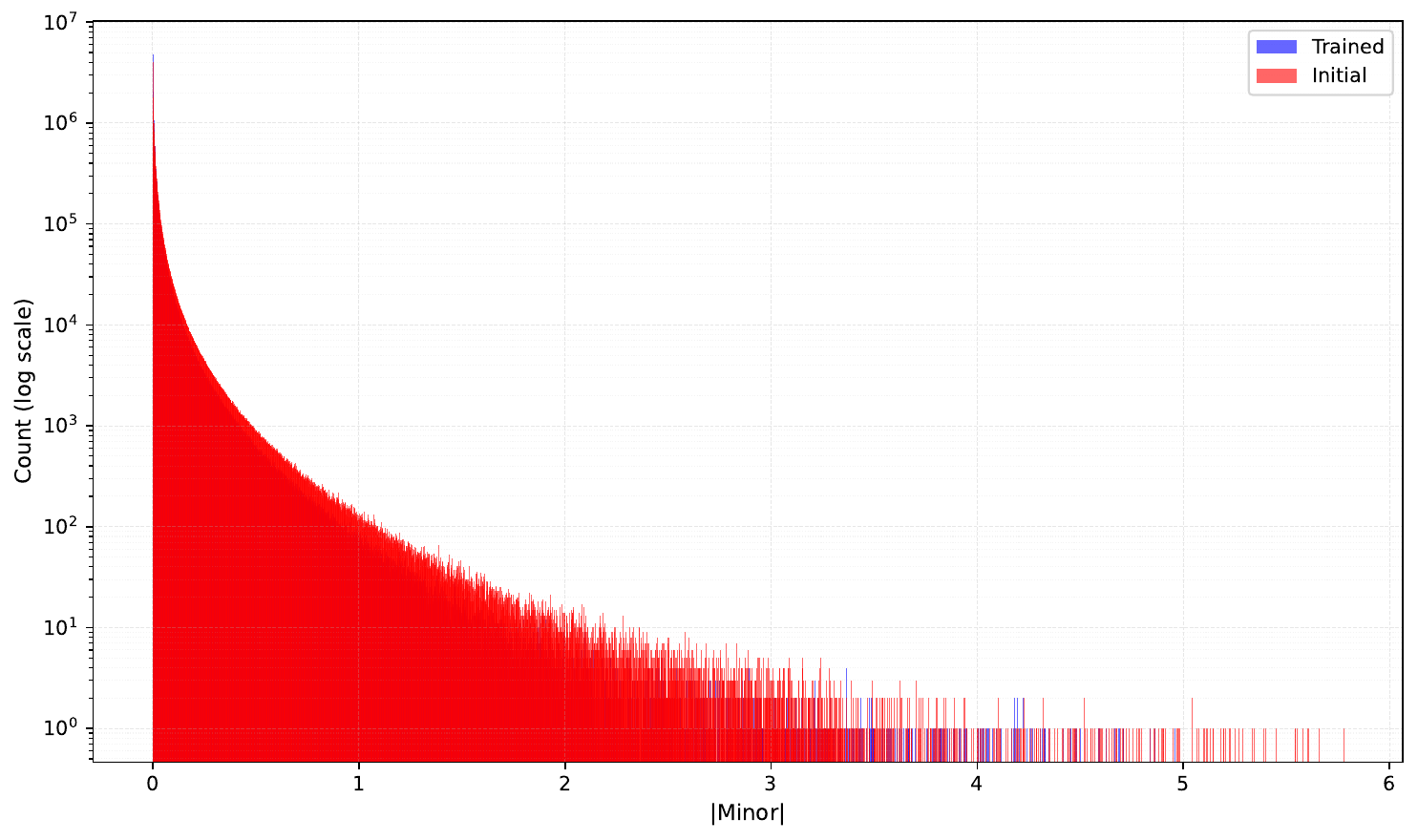}
    \caption{\texttt{linear}$(32)$}
    \label{fig:panel_b}
\end{subfigure}
\hfill
\begin{subfigure}[b]{0.32\textwidth}
    \centering
    \includegraphics[width=\textwidth]{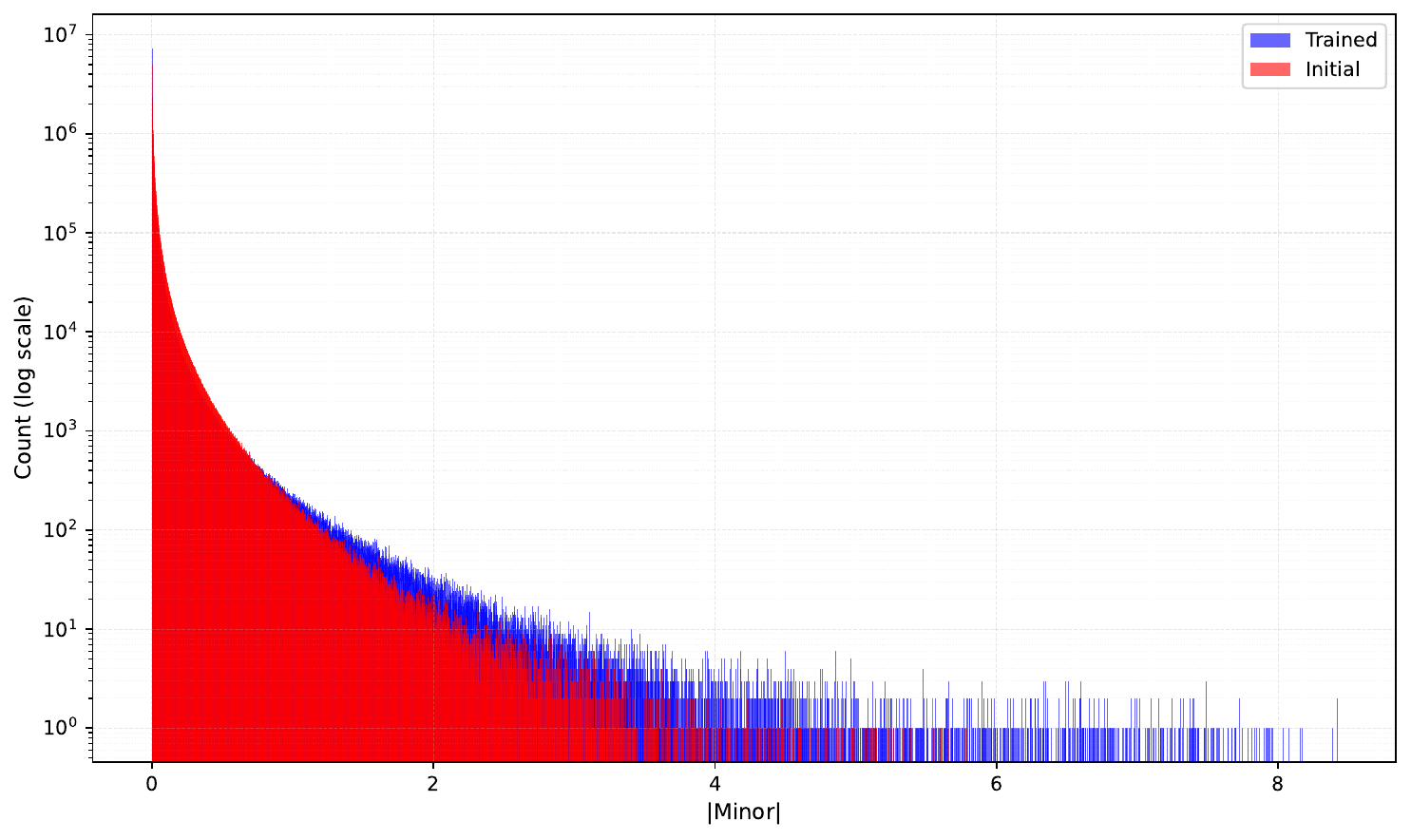}
    \caption{\texttt{random}$(32)$}
    \label{fig:panel_c}
\end{subfigure}
\caption{
Size-$3$ minor distribution comparison across target functions.
}
\label{minordistributioncomparison}
\end{figure}
We next look for KA geometry through minor concentration.
Fig.~\ref{minordistributioncomparison} shows the size-$3$ minor distributions of the \texttt{xor}$(32)$, \texttt{linear}$(32)$, and \texttt{random}$(32)$ models.
These plots illustrate the relatively heavy tails of minor distributions produced by training \texttt{xor}$(32)$, in addition to the peakedness near zero, already implied by the row mean Zipf plots in Fig.~\ref{rankvalueplots}. 
We will interpret the distributional shifts induced by training in terms of minor concentration.

\subsubsection{Participation Ratios}
\label{participationratiosection}

Recall from \S \ref{whatissection} that minor concentration is sourced by alignment and scale.
Both of these are measured by the size-$k$ participation ratios:
\begin{align}
\label{participationratiodef}
{\rm PR}^{(k)} = {1 \over B} \sum_{\mathbf{x}} {\|J^{(k)}\big(\mathbf{x} \big)\|_1 \over  \|J^{(k)}\big(\mathbf{x} \big)\|_2},
\end{align}
where $B$ is the number of examples $\mathbf{x}$ and $\| J^{(k)}(\mathbf{x}) \|_p$ is the $L_p$ norm of the elements of $J^{(k)}(\mathbf{x})$, flattened into a vector.
This ratio is bounded from below by $1.0$ and from above by ${\rm PR}^{(k)}_{\rm max} \equiv \sqrt{{{m}\choose{k}} {{n}\choose{k}}}$, the square-root of the number of elements in $J^{(k)}(\mathbf{x})$.
KA geometry favors small values of this ratio.

\begin{table}[htbp]
\centering
\renewcommand{\arraystretch}{1.3}
\begin{tabular}{|l@{\hspace{0.5cm}}|c@{\hspace{0.5cm}}|c@{\hspace{0.5cm}}|c|}
\hline
Model & $k=1$ & $k=2$ & $k=3$ \\
\hline
\ \texttt{xor}$(32)$ & $6.08 \pm 0.21$ & $13.26 \pm 0.81$ & $16.24 \pm 1.47$ \\
\ \texttt{linear}$(32)$ & $6.31 \pm 0.04$  & $18.64 \pm 0.21$ & $26.62 \pm 0.64$ \\
\ \texttt{random}$(32)$ & $5.67 \pm 0.18$ & $15.42 \pm 0.63$  & $20.97 \pm 1.00$ \\
\hline
\ Initial$(32)$ & $6.29 \pm 0.09$ & $18.64 \pm 0.53$ & $27.25 \pm 0.89$ \\
\ ${\rm PR}^{(k)}_{\rm max}$ & $9.80$ & $38.57$ & $70.43$ \\
\hline
\end{tabular}
\caption{Size-$k$ participation ratios (mean $\pm$ standard deviation).
The ``Initial" row contains the participation ratios of the models at initialization.
The ${\rm PR}^{(k)}_{\rm max}$ row contains the theoretical maximum participation ratios for each $k$.
}
\label{participationratiotable}
\end{table}
Table \ref{participationratiotable} lists the participation ratios of the \texttt{xor}$(32)$, \texttt{linear}$(32)$, and \texttt{random}$(32)$ models.
It's noteworthy that all models, even the initial, have participation ratios that are significantly less than the upper bound given by ${\rm PR}^{(k)}_{\rm max}$.
Similar to what we found when looking at zero rows, we see that the participation ratios of \texttt{linear}$(32)$ are similar to their initial values. 
Minor concentration in the \texttt{xor}$(32)$ model, relative to the \texttt{linear}$(32)$ and \texttt{random}$(32)$ models,  is not readily apparent in the size-$1$ participation ratios; instead, it appears in the ${\rm PR}^{(k)}$ of higher-$k$ minors.
\begin{figure}[h]
\centering
\begin{subfigure}[b]{0.32\textwidth}
    \centering
    \includegraphics[width=\textwidth]{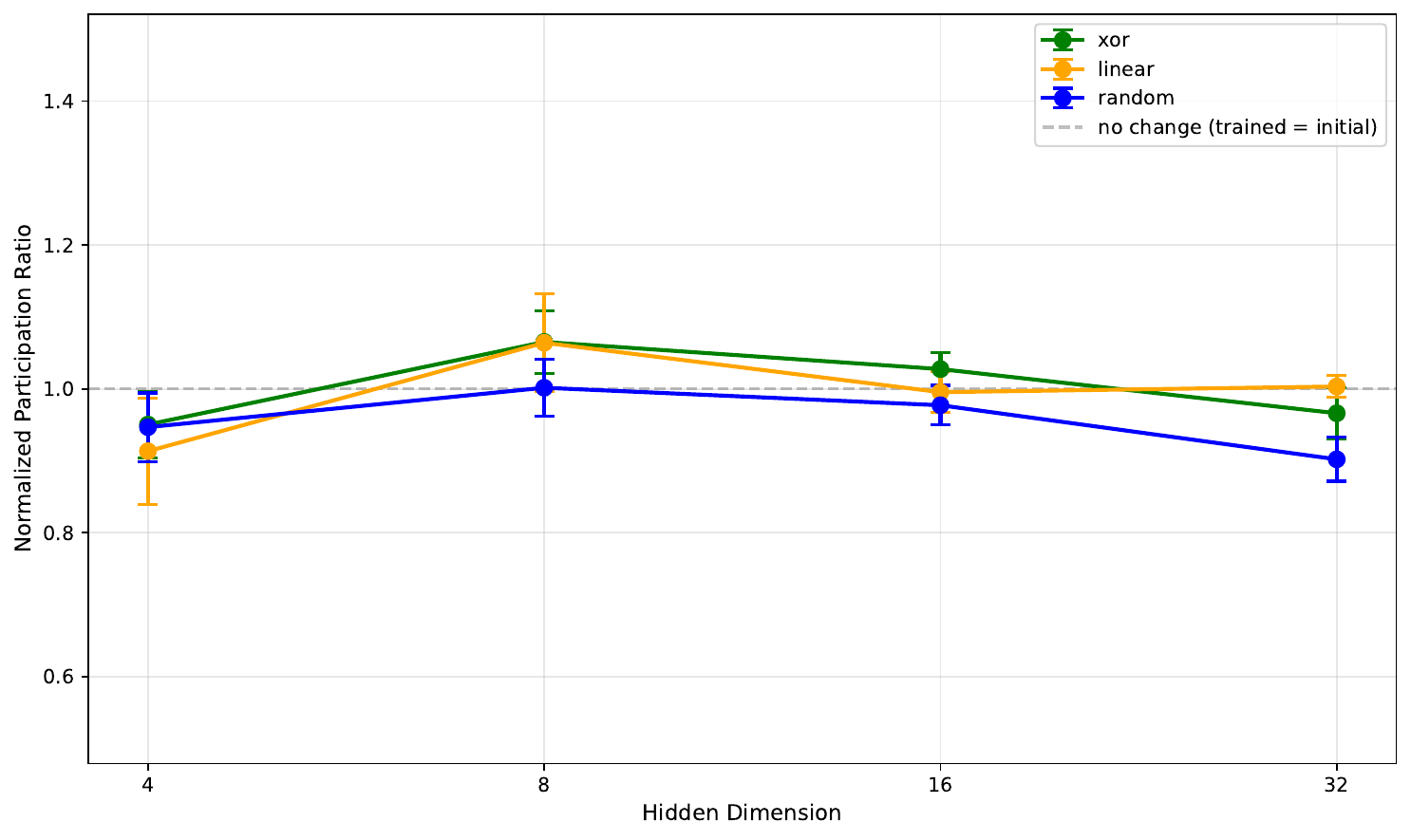}
    \caption{$k=1$}
    \label{fig:panel_a}
\end{subfigure}
\hfill
\begin{subfigure}[b]{0.32\textwidth}
    \centering
    \includegraphics[width=\textwidth]{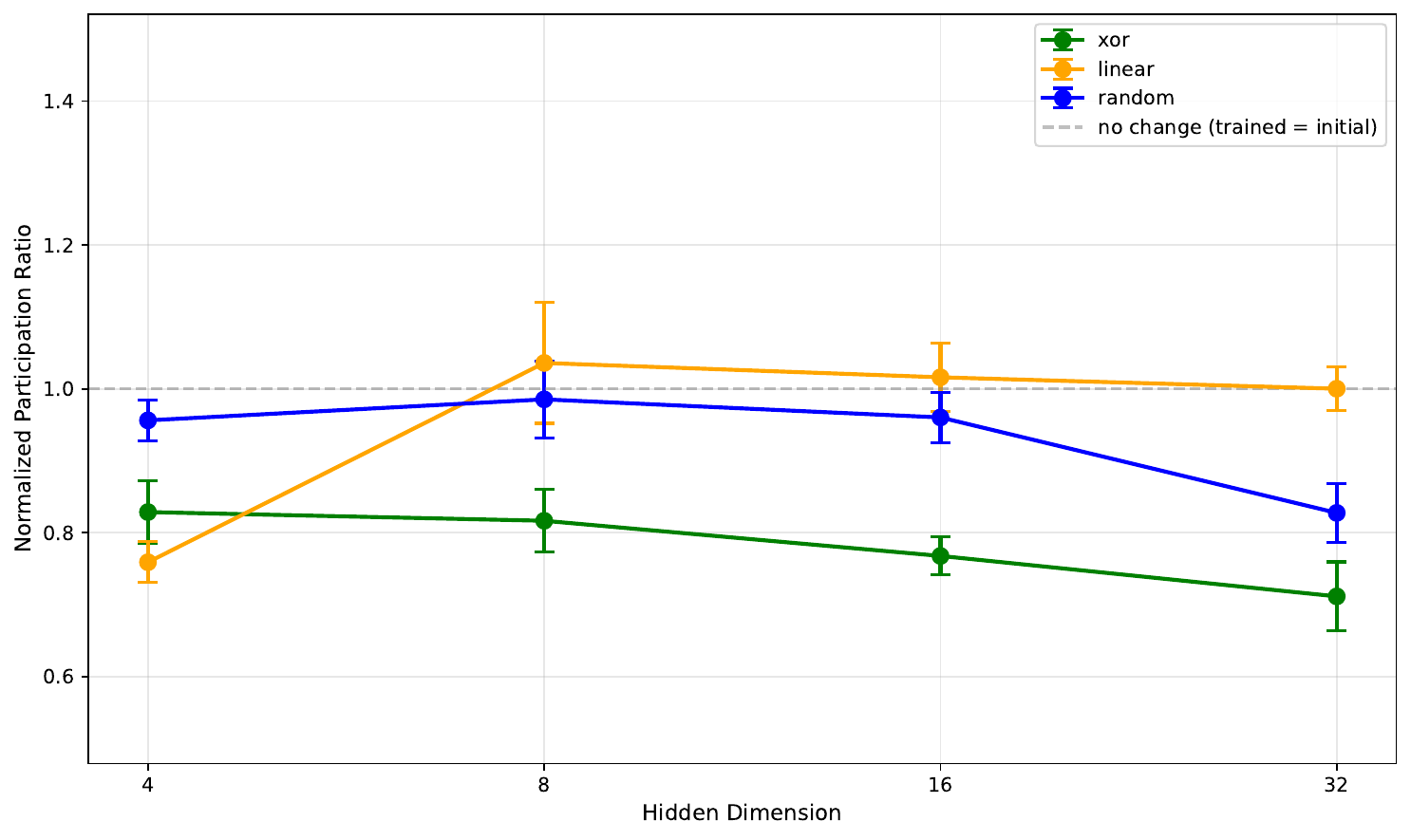}
    \caption{$k=2$}
    \label{fig:panel_b}
\end{subfigure}
\hfill
\begin{subfigure}[b]{0.32\textwidth}
    \centering
    \includegraphics[width=\textwidth]{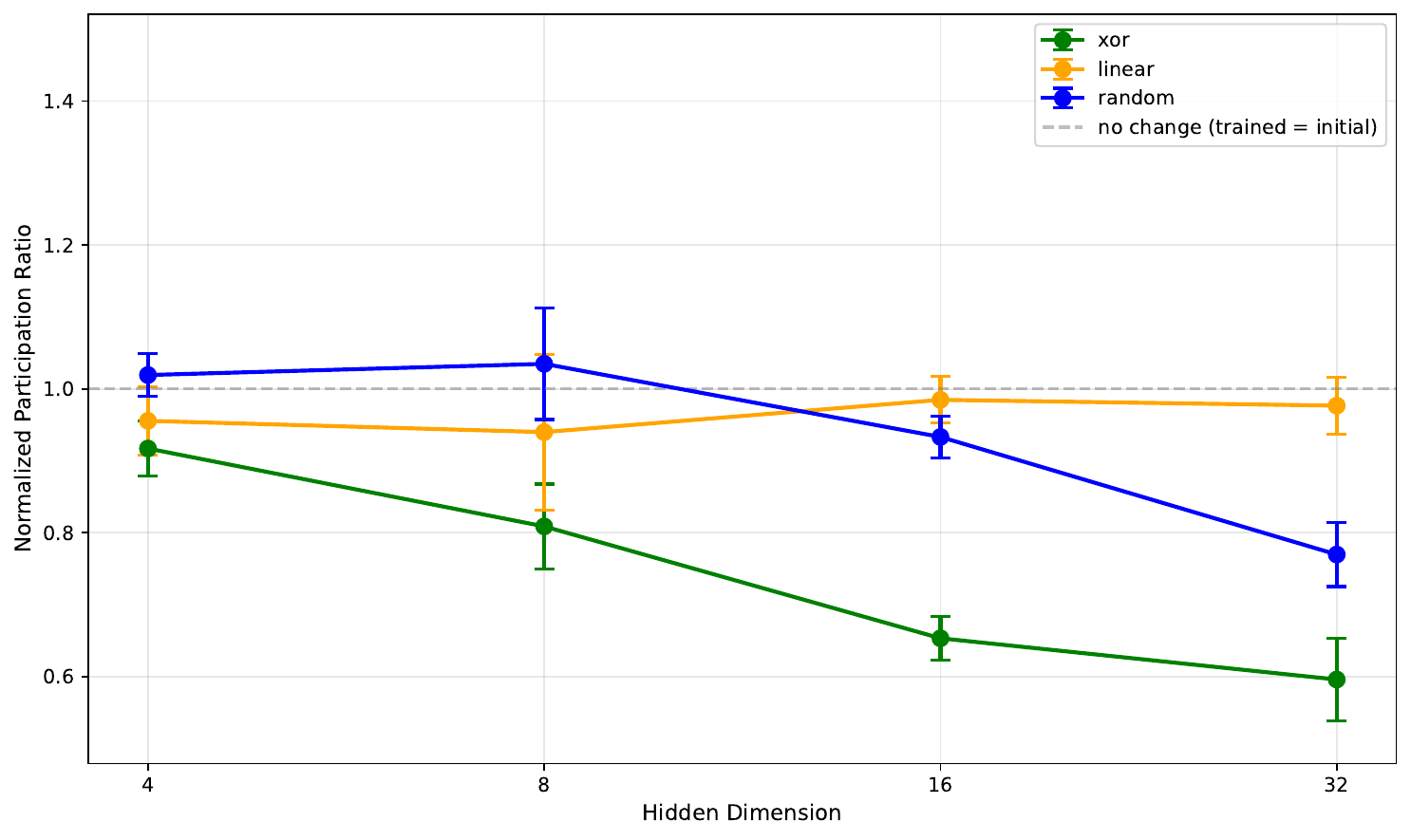}
    \caption{$k=3$}
    \label{fig:panel_c}
\end{subfigure}
\caption{
Normalized participation ratios (mean $\pm$ standard error) across hidden dimensions.
}
\label{normalizedparticipationratiocomparison}
\end{figure}
We can better highlight the effect of training on the participation ratios by normalizing them, i.e., we divide the participation ratio of a trained model by its value at initialization. 
This provides a common baseline, since all models are trained from identical initializations.
Fig.~\ref{normalizedparticipationratiocomparison} shows how these normalized participation ratios vary with the size of the hidden dimension $m$.
The noticeable decrease in the normalized ${\rm PR}^{(k)}$ of the \texttt{xor}$(32)$ model occurs for training $R^2 > 0.5$ (see Table \ref{train_r2_table}).

\subsubsection{Random Rotation Ratios}
\label{maxpermutationratiosection}

To what extent are the tail behaviors of the minor distributions exceptional?
If the minors are disrupted by an arbitrary rotation, will their tails contract?
We probe this alignment property via the random rotation ratio $\overline r^{(k)}$, defined as follow.

For each Jacobian $J_{ji}({\bf x})$, find the largest elements of the corresponding size-$k$ minor matrices: $M^{(k)}({\bf x}) = {\rm max}_{\mathbf{j} \mathbf{i}} \big(J_{\mathbf{j} \mathbf{i}}^{(k)}(\mathbf{x}) \big)$.
$M^{(k)}({\bf x})$ represents the strongest response of the size-$k$ minor of the model to example ${\bf x}$.
Next, generate a set of $m \times m$ orthogonal matrices $R_p$, $p = 1, \cdots, N$ (we took $N=400$), uniformly distributed in $SO(m)$.
For each matrix, compute the randomly-rotated Jacobian:
\begin{align}
J_p(\mathbf{x}) = R_p \cdot J(\mathbf{x}).
\end{align}
From the rotated Jacobians, obtain randomly-rotated size-$k$ minor matrices $J_{p}^{(k)}({\bf x})$.
The elements of these matrices are simply the absolute values of the $k \times k$ minors of $J_{p}({\bf x})$.
Determine the largest elements $m^{(k)}_{p}({\bf x}) = {\rm max}_{\mathbf{j} \mathbf{i}}  \Big( \big( J^{(k)}_{p}({\bf x}) \big)_{\mathbf{j} \mathbf{i}} \Big)$ and compute the average over rotations: $m^{(k)}({\bf x}) = {1 \over N} \sum_p m^{(k)}_{p}({\bf x})$.
This gives the fraction, $r^{(k)}({\bf x)} = M^{(k)}({\bf x})/m^{(k)}({\bf x})$.
The random rotation ratios $\overline r^{(k)}$ are simply the averages across all examples of this fraction:
\begin{align}
\label{randomrotationratio}
\bar r^{(k)} = {1 \over B} \sum_{\bf x} r^{(k)}({\bf x}).
\end{align}
We used the same set of rotations for all examples; this should be OK since the rotations are uniformly distributed in $SO(m)$. 
A large random rotation ratio suggests that the largest minors are not arising as a statistical accident, but rather reflect a conspiracy: A group of $k$ input neurons is truly attempting to communicate with a specific group of $k$ hidden-layer neurons, and this is being thwarted by our random rotations. 
This results in a contraction of the tail of the minor distribution of randomly rotated Jacobians.

To get a feeling for this, it's helpful to consider the random rotation ratios of a few idealized Jacobians for a model with $n=3$ input dimensions and $m=7$ hidden neurons:
\begin{align}
\label{aligned}
  J_1 &= \begin{pmatrix} \mathbf{I}_3 \\ \mathbf{0}_{4 \times 3} \end{pmatrix},  \\[0.5em]
  \label{gaussian}
  J_2 &= (J_{ji})_{7 \times 3}, \quad J_{ji} \overset{\text{i.i.d.}}{\sim} \mathcal{N}(0,1),  \\[0.5em]
  \label{uniform}
  J_3 &= (J_{ji})_{7 \times 3}, \quad J_{ji} \overset{\text{i.i.d.}}{\sim} \mathcal{U}(-1,1), 
\end{align}
where $\mathbf{I}_3$ is the $3 \times 3$ identity matrix and $\mathbf{0}_{4 \times 3}$ is the $4 \times 3$ zero matrix.
$J_1$ represents a Jacobian with perfect alignment (and normalized columns).
$J_2$ and $J_3$ represent ``maximally-unaligned" Jacobians with random entries of either bounded or decaying character.
\begin{table}[h]
    \centering
    \begin{tabular}{lccc}
    \hline
    Model & $k=1$ & $k=2$ & $k=3$ \\
    \hline
    $J_1$ (Aligned) & 1.483 & 1.803 & 2.607 \\
    $J_2$ (Gaussian) & 0.993 & 0.821 & 0.841 \\
    $J_3$ (Uniform) & 0.768 & 0.958 & 0.955 \\
    \hline
    \end{tabular}
    \caption{Random rotation ratios of the toy Jacobians \eqref{aligned}, \eqref{gaussian}, \eqref{uniform}. 
    }
    \label{toymmrratios}
\end{table}
Table \ref{toymmrratios} reports the corresponding random rotation ratios.
We see that the perfectly aligned Jacobian sees a greater than $1.0$ random rotation ratio that gets more pronounced with $k$.
On the other hand, the uniform and Gaussian Jacobians are relatively unaffected by the random rotations, although there's some weak dependence on $k$.
Notice that an overall rescaling of the Jacobian does not modify the corresponding random rotation ratios; a nonuniform scaling of the entries of a Jacobian can affect the ratios, e.g., a nonuniform scaling of the three nonzero values of $J_1$, can modify the $k=1$ and $k=2$ ratios.

\begin{figure}[h]
\centering
\begin{subfigure}[b]{0.32\textwidth}
    \centering
    \includegraphics[width=\textwidth]{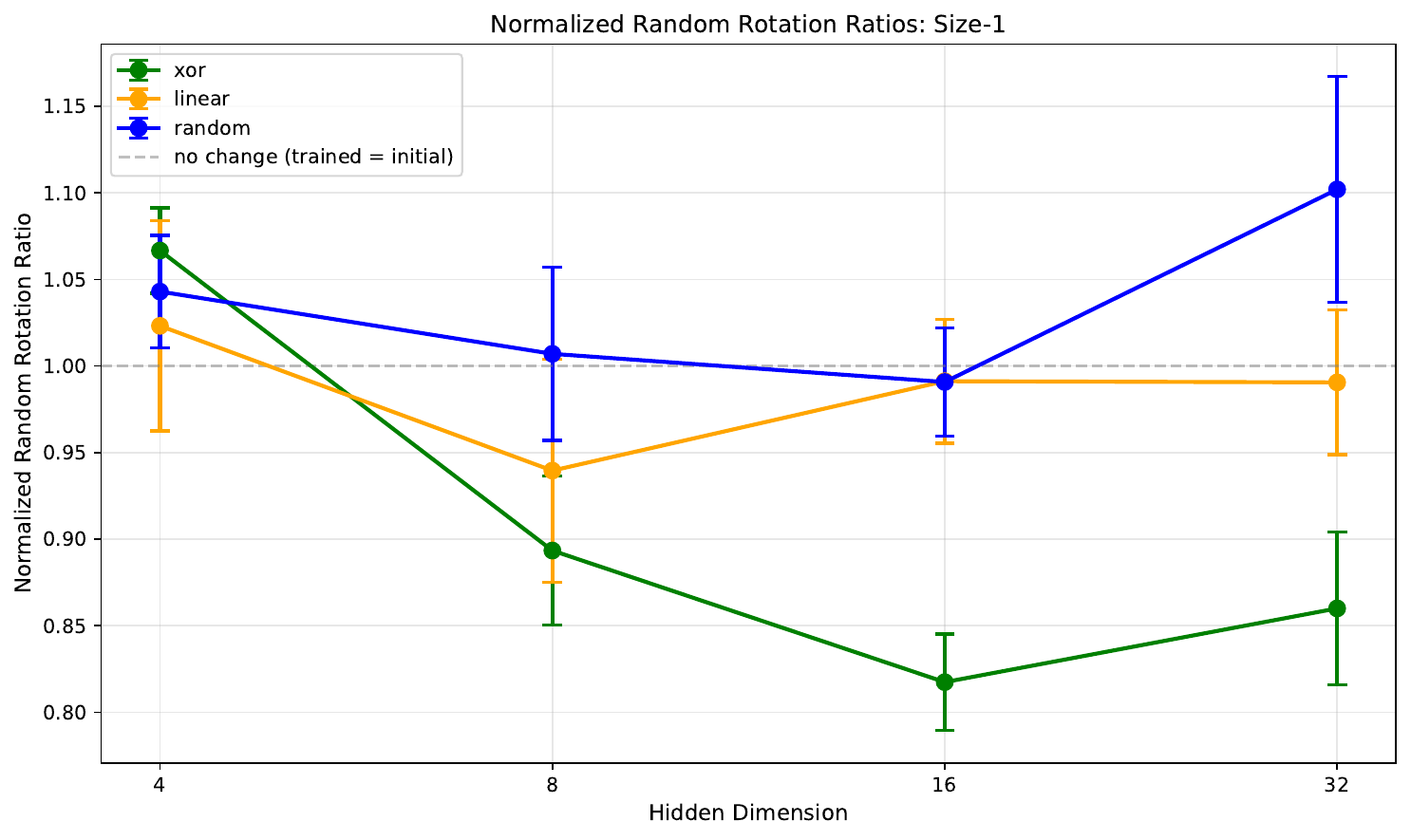}
    \caption{$k=1$}
    \label{fig:panel_a}
\end{subfigure}
\hfill
\begin{subfigure}[b]{0.32\textwidth}
    \centering
    \includegraphics[width=\textwidth]{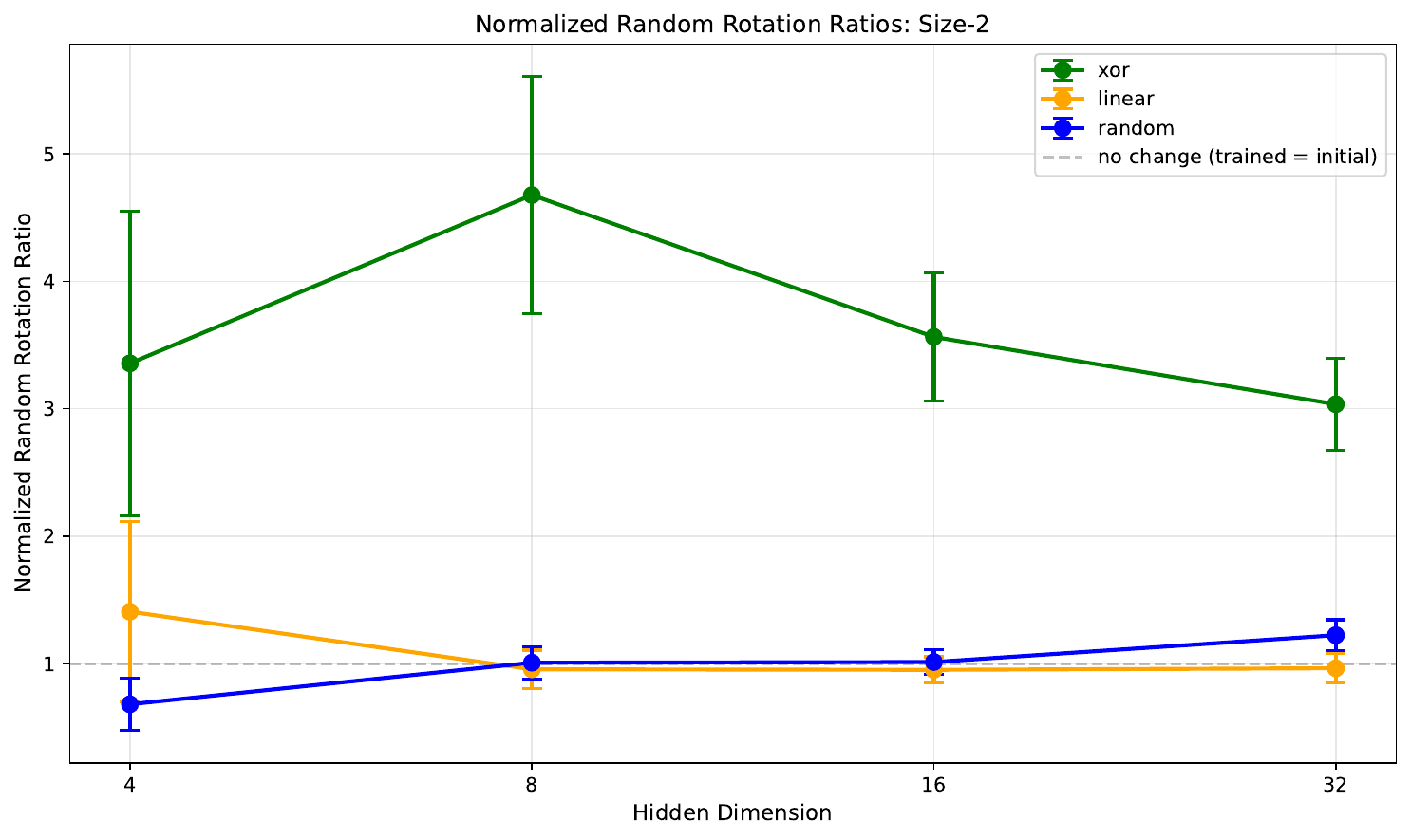}
    \caption{$k=2$}
    \label{fig:panel_b}
\end{subfigure}
\hfill
\begin{subfigure}[b]{0.32\textwidth}
    \centering
    \includegraphics[width=\textwidth]{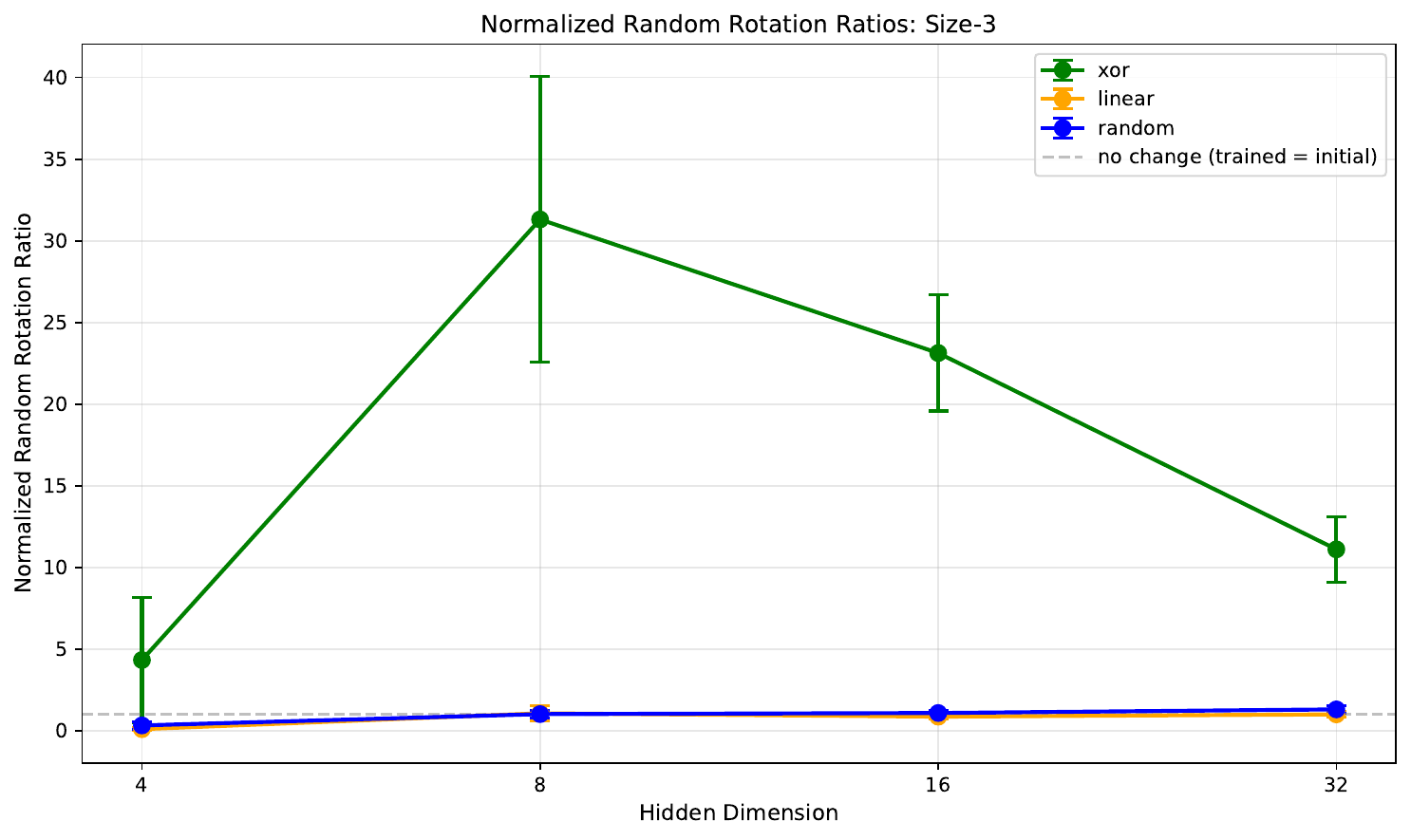}
    \caption{$k=3$}
    \label{fig:panel_c}
\end{subfigure}
\caption{
Normalized random rotation ratios across hidden dimensions.
%  $m \in \{4, 8, 16, 32 \}$.
% Plotted are the normalized ratios averaged over model seeds, for each $m$, with error bars given by the estimated standard error.
}
\label{normalizedmmrratiocomparison}
\end{figure}

We have found the initial random rotation ratios to scale as $c k$ for some constant $c \approx 1.0$.
Fig.~\ref{normalizedmmrratiocomparison} plots the random rotation ratios of the \texttt{xor}$(m)$, \texttt{linear}$(m)$, and \texttt{random}$(m)$ models, normalized by the initial random rotation ratios.
We see that all models have random rotation ratios near $1.0$ for $k=1$.
As $k$ increases, the \texttt{xor} model is distinguished from the \texttt{linear} and \texttt{random} models, displaying noticeably larger random rotation ratios. 
This provides strong evidence that the \texttt{xor} model's inner map exhibits significant alignment.
It's notable that the \texttt{xor}$(m)$ model's $k>1$ random rotation ratio is greater than one for all $m$.

\subsubsection{Column Divergence}
\label{columndivergencesection}

The random rotation ratio measures the alignment of exceptional minors.
What about generic ones?
In particular, how are typical input $k$-planes aligned within the hidden dimension?
For example, consider column $(i', j')$ of the size-$2$ minor matrix $J^{(2)}({\bf x})$.
If the 2-plane defined by $(i', j')$ gets mapped entirely along, say, hidden 2-plane $(1, 2)$ then $J^{(2)}_{(i, j), (i', j')}({\bf x})$ would be nonzero only when $(i, j) = (1,2)$.
On the other hand, the $(i', j')$ plane might be evenly distributed across all hidden 2-planes, with each element of the $(i', j')$ column $J^{(2)}_{(i, j), (i', j')}({\bf x})$ taking roughly the same value.

To measure the degree to which alignment occurs in general, we view the columns of $J^{(k)}({\bf x})$ as logits and form column probabilities using softmax (at unit temperature):
% (recall the elements of $J^{(k)}({\bf x})$ are nonnegative).
% There are various ways to form a probability distribution out of these logits.
% We can form a probability by normalizing the elements in a given column by the sum of the corresponding column elements: 
% This $L_1$ normalization preserves the relative magnitudes of column entries; alternatively, using a nonlinearity such as softmax (at unit temperature) to form a probability can greatly accentuate small differences in values, especially in higher-$k$ minors in which sparsity increases.
% Denoting minor matrix index tuples by $\mathbf{j}, \mathbf{i}$, w
% We have the column probabilities under $L_1$ normalization:
\begin{align}
\label{columnprobsdef}
% P^{(k)}_{\mathbf{j} \mathbf{i}}({\bf x}) & = {J^{(k)}_{\mathbf{j} \mathbf{i}} ({\bf x})\over \sum_{\mathbf{l}} J^{(k)}_{\mathbf{l} \mathbf{i}}({\bf x}) }.
%  \\
P^{(k)}_{\mathbf{j} \mathbf{i}}({\bf x}) & = {\exp\big(J^{(k)}_{\mathbf{j} \mathbf{i}} ({\bf x}) \big) \over \sum_{\mathbf{l}} \exp\big( J^{(k)}_{\mathbf{l} \mathbf{i}} ({\bf x}) \big)}.
\end{align}
The softmax nonlinearity accentuates small differences in column entries and makes the effects of training more visible.
An alternative choice is to normalize the elements in a given column by the sum of the corresponding column elements (recall the elements of $J^{(k)}({\bf x})$ are nonnegative), thereby preserving the relative magnitudes of elements.
% For a softmax normalization we replace $P^{(k)}_{\mathbf{j} \mathbf{i}}({\bf x})$ above with ${\rm Softmax}\big(J^{(k)}_{\mathbf{j} \mathbf{i}} ({\bf x}) \big) = {\exp\big(J^{(k)}_{\mathbf{j} \mathbf{i}} ({\bf x}) \big) \over \sum_{\mathbf{l}} \exp\big( J^{(k)}_{\mathbf{l} \mathbf{i}} ({\bf x}) \big)}$.
\eqref{columnprobsdef} gives us per-column ($\mathbf{i}$) and per-example ($\mathbf{x}$) probabilities, from which we can compute column entropies or column divergences of or between the initial and trained models.
The column entropy is the Shannon entropy of a column of a size-$k$ minor matrix:
\begin{align}
\label{shannondef}
S_{\mathbf{i}}^{(k)}({\bf x}) = - \sum_{\mathbf{j}} P^{(k)}_{\mathbf{j} \mathbf{i}} ({\bf x}) \log P^{(k)}_{\mathbf{j} \mathbf{i}} ({\bf x}).
\end{align}
The column divergence is the KL divergence between the trained $P^{(k)}_{\mathbf{j} \mathbf{i}} ({\bf x})$ and initial $Q^{(k)}_{\mathbf{j} \mathbf{i}} ({\bf x})$ models for the size-$k$ minor matrix:
\begin{align}
\label{kldef}
KL_{\mathbf{i}}^{(k)}({\bf x}) = \sum_{\mathbf{j}} P^{(k)}_{\mathbf{j} \mathbf{i}} ({\bf x}) \log \big( P^{(k)}_{\mathbf{j} \mathbf{i}} ({\bf x}) / Q^{(k)}_{\mathbf{j} \mathbf{i}} ({\bf x}) \big).
\end{align}
For both quantities, we choose the base of the logarithm equal to $\binom{m}{k}$,  the number of rows in the size-$k$ minor matrix.
This ensures the entropy lies in $[0,1]$, with an entropy of $1.0$ signifying uniform distribution of values and an entropy of $0.0$ signifying complete alignment.
The KL divergence is a nonnegative function that increases as the trained model diverges from its initialization.

\begin{figure}[h]
\centering
\begin{subfigure}[b]{0.32\textwidth}
    \centering
    \includegraphics[width=\textwidth]{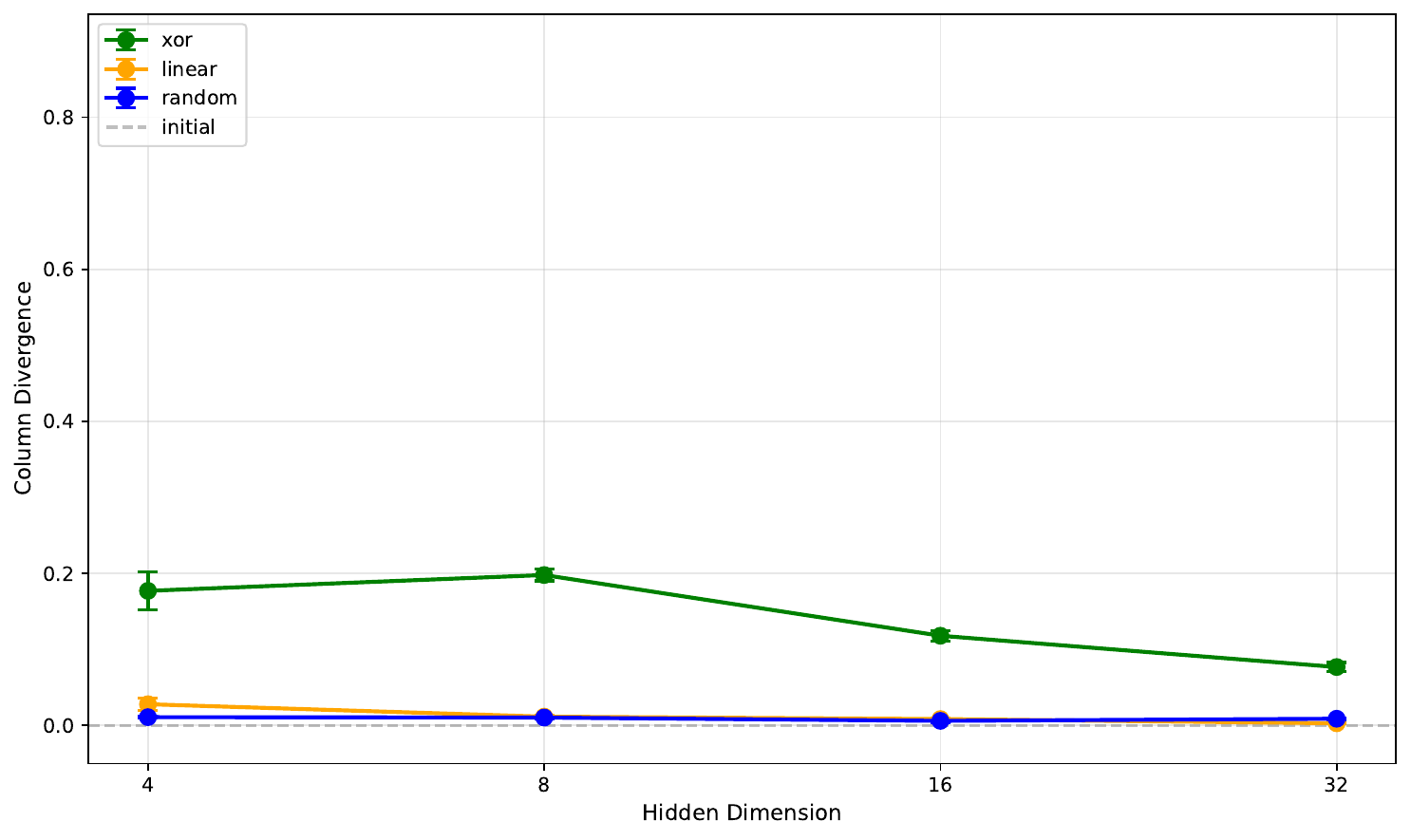}
    \caption{$k=1$}
    \label{fig:panel_a}
\end{subfigure}
\hfill
\begin{subfigure}[b]{0.32\textwidth}
    \centering
    \includegraphics[width=\textwidth]{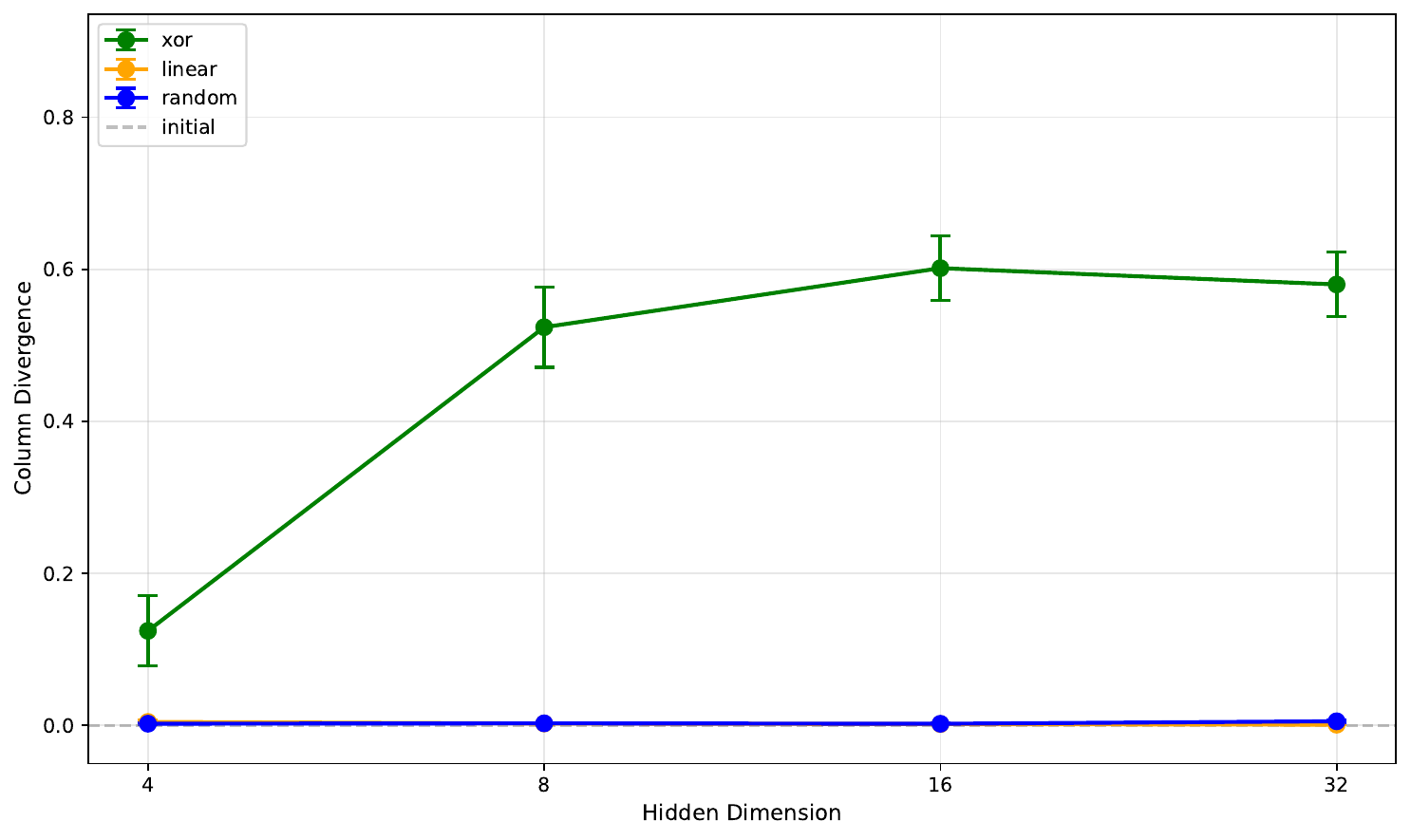}
    \caption{$k=2$}
    \label{fig:panel_b}
\end{subfigure}
\hfill
\begin{subfigure}[b]{0.32\textwidth}
    \centering
    \includegraphics[width=\textwidth]{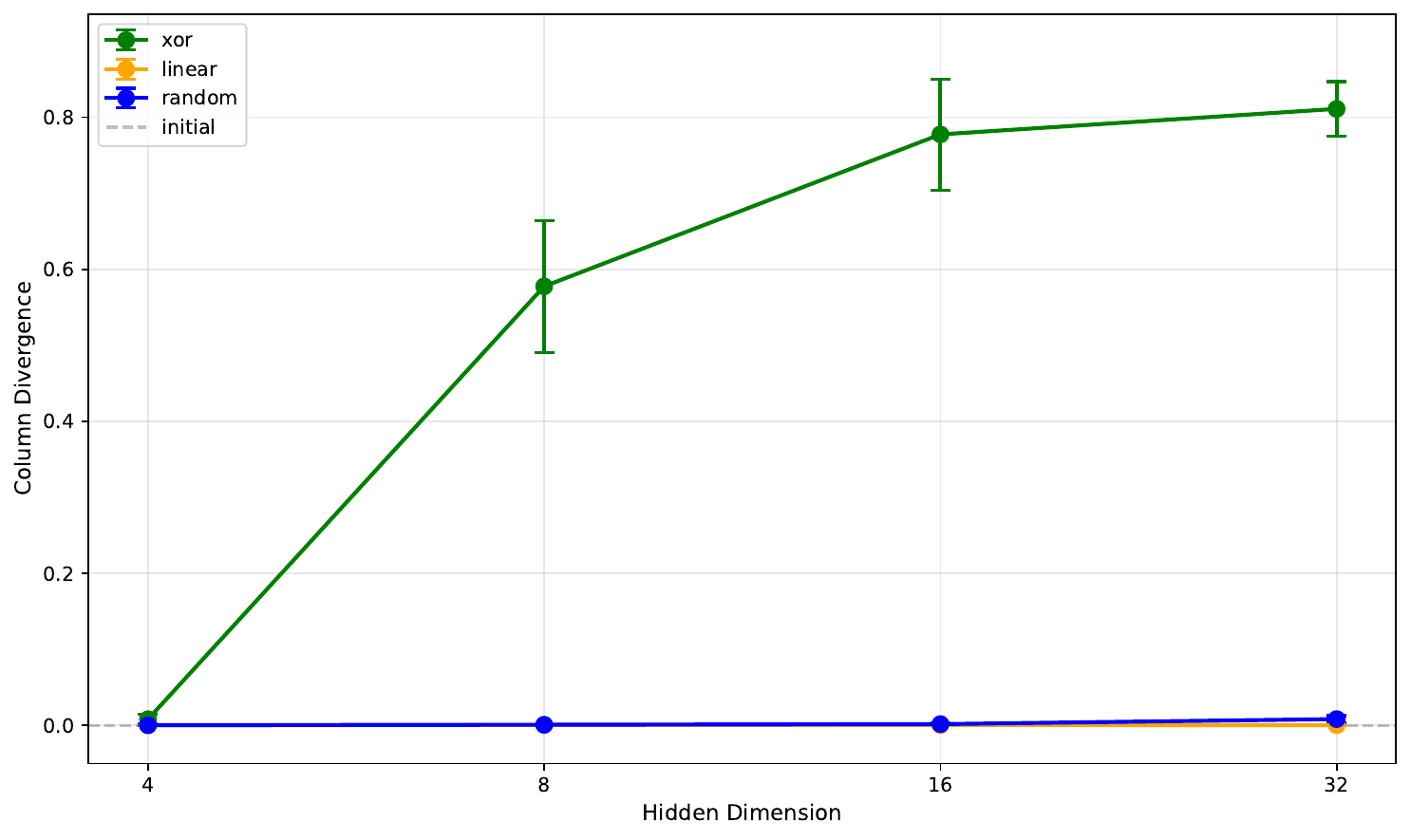}
    \caption{$k=3$}
    \label{fig:panel_c}
\end{subfigure}
\caption{
Column divergences of size-$k$ minor matrices across hidden dimensions.
}
\label{columnentropyocomparison}
\end{figure}
Fig.~\ref{columnentropyocomparison} plots the column divergences of size-$k$ minor matrices, averaged over columns and examples.
To interpret these plots, we need to know the column entropies of the model at initialization.
We find these initial entropies to be very close to $1.0$, reflective of the relatively evenly distributed initial minor values.
The column divergences then indicate that both the (trained) \texttt{linear}$(m)$ and \texttt{random}$(m)$ models stay close to their initial even-distributed values (similar to what we've seen in other metrics).
On the other hand, the \texttt{xor}$(m)$ model differs noticeably from the initial model; this difference is monotonically increasing with hidden dimension $m$ for $k=2$ and $k=3$, while relatively constant for $k=1$.
This indicates that minor matrices in \texttt{xor}$(m)$ develop increasing alignment through training and this alignment becomes more pronounced as model size increases.

\subsection{Dynamics}
\label{interpolatesection}

In the previous sections, we compared zero row and minor concentration metrics between initial and trained models.
We now consider the dynamics of these metrics as models train or interpolate between target functions of tunable complexity, focusing on models with $m=32$ hidden dimensions.
The primary conclusion will be that KA geometry is positively correlated with model performance on sufficiently complex functions.
This conclusion is already supported by the earlier experiments tracking KA geometry with the number of hidden neurons---recall from Table \ref{train_r2_table} that the \texttt{xor}$(m)$ model's performance steadily increases with $m$.

\begin{figure}[h]
\centering
\begin{subfigure}[b]{0.49\textwidth}
    \centering
    \includegraphics[width=\textwidth]{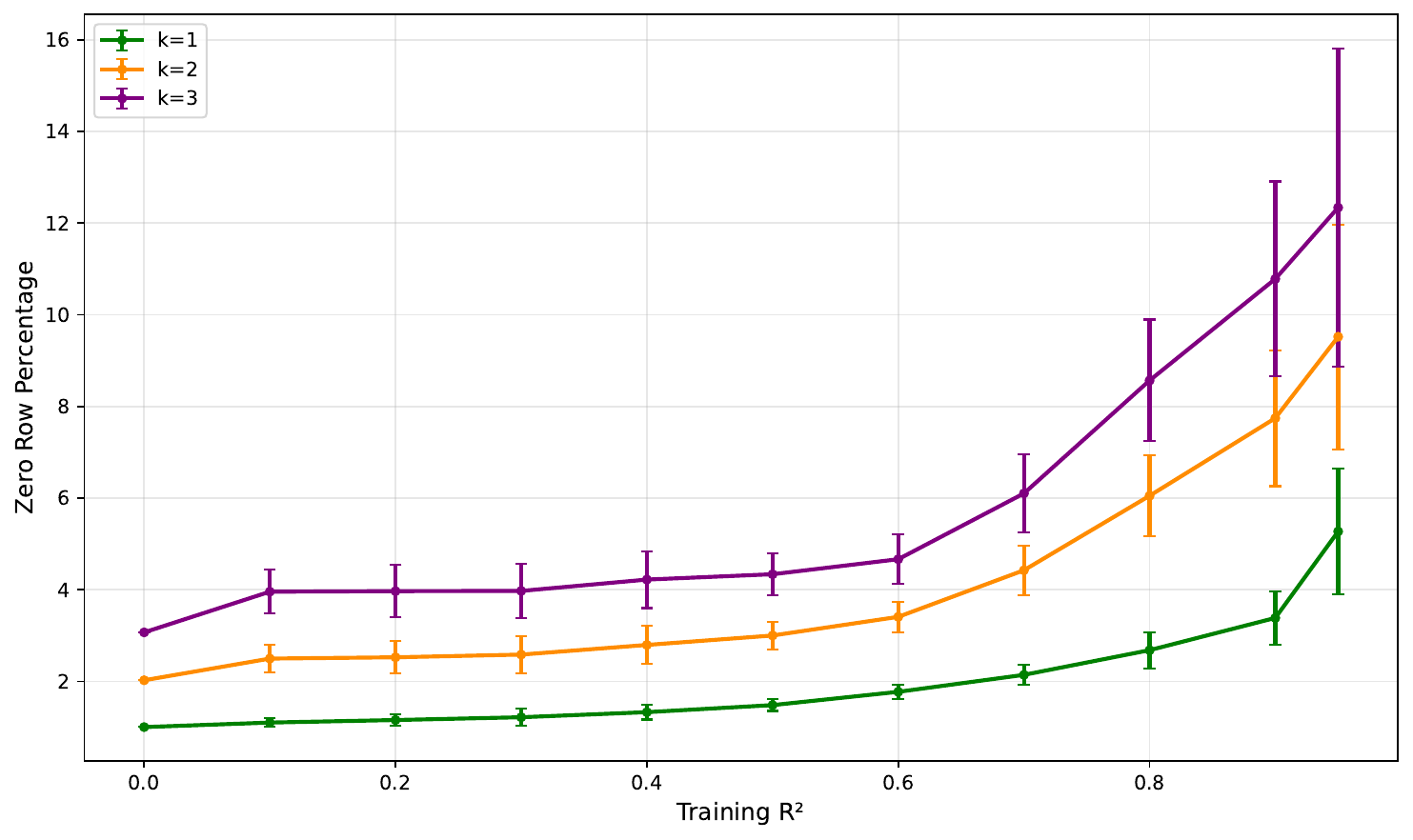}
    \caption{Zero Rows}
    \label{zeroevolution}
\end{subfigure}
\hfill
\begin{subfigure}[b]{0.49\textwidth}
    \centering
    \includegraphics[width=\textwidth]{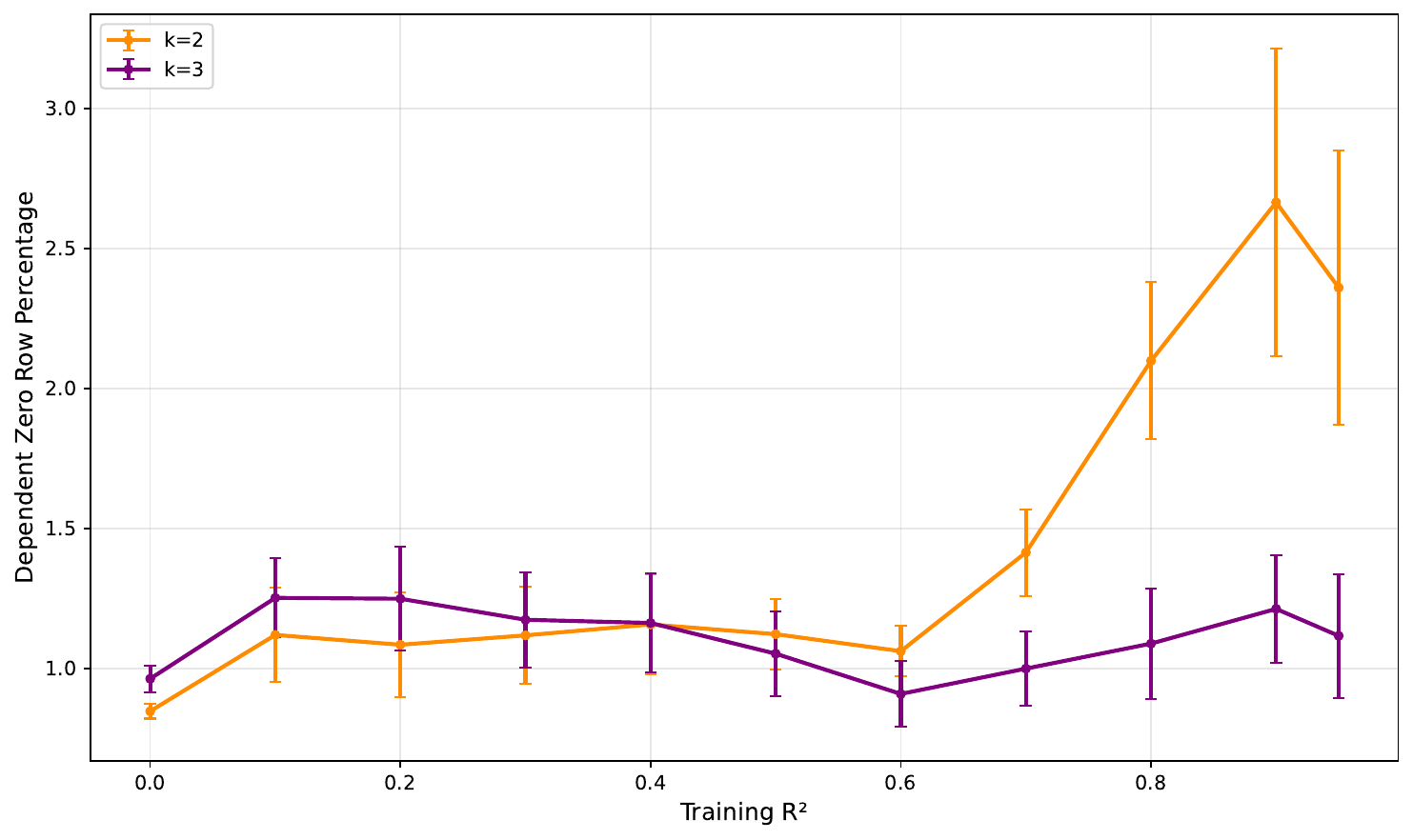}
    \caption{Dependent Zero Rows}
    \label{dependentevolution}
\end{subfigure}
\caption{
Evolution over training of size-$k$ zero row and dependent zero row percentages in the \texttt{xor}$(32)$ model.
}
\label{zerorowmetricsevolution}
\end{figure}
We begin with time-evolution experiments that track KA geometry over training.
For this, we will monitor the model at regular intervals in training $R^2$ and record the KA metrics.
Note that this gives the false impression that training proceeds regularly; instead, the model makes a great deal of progress in learning the function during the initial training epochs and then asymptotes towards its optimal solution later in training.

Fig.~\ref{zerorowmetricsevolution} shows how zero row metrics vary with $R^2$ using a ``false-positive rate" $q^{(1)} = 0.01$.
Around $R^2 \approx 0.5$, we observe in Fig.~\ref{zeroevolution} the development and steady increase with $R^2$ of the number of zero rows.
The nonzero percentages of zero rows for $R^2 < 0.5$ should be interpreted as ``false-positives" because the fraction of zero rows is approximately equal to $q^{(k)}$.
We likewise see the development of size-$2$ dependent zero rows in Fig.~\ref{dependentevolution} around training $R^2 \approx 0.5$; we do not discern a significant signal in size-$3$ dependent zero rows.

\begin{figure}[h]
\centering
\begin{subfigure}[b]{0.32\textwidth}
    \centering
    \includegraphics[width=\textwidth]{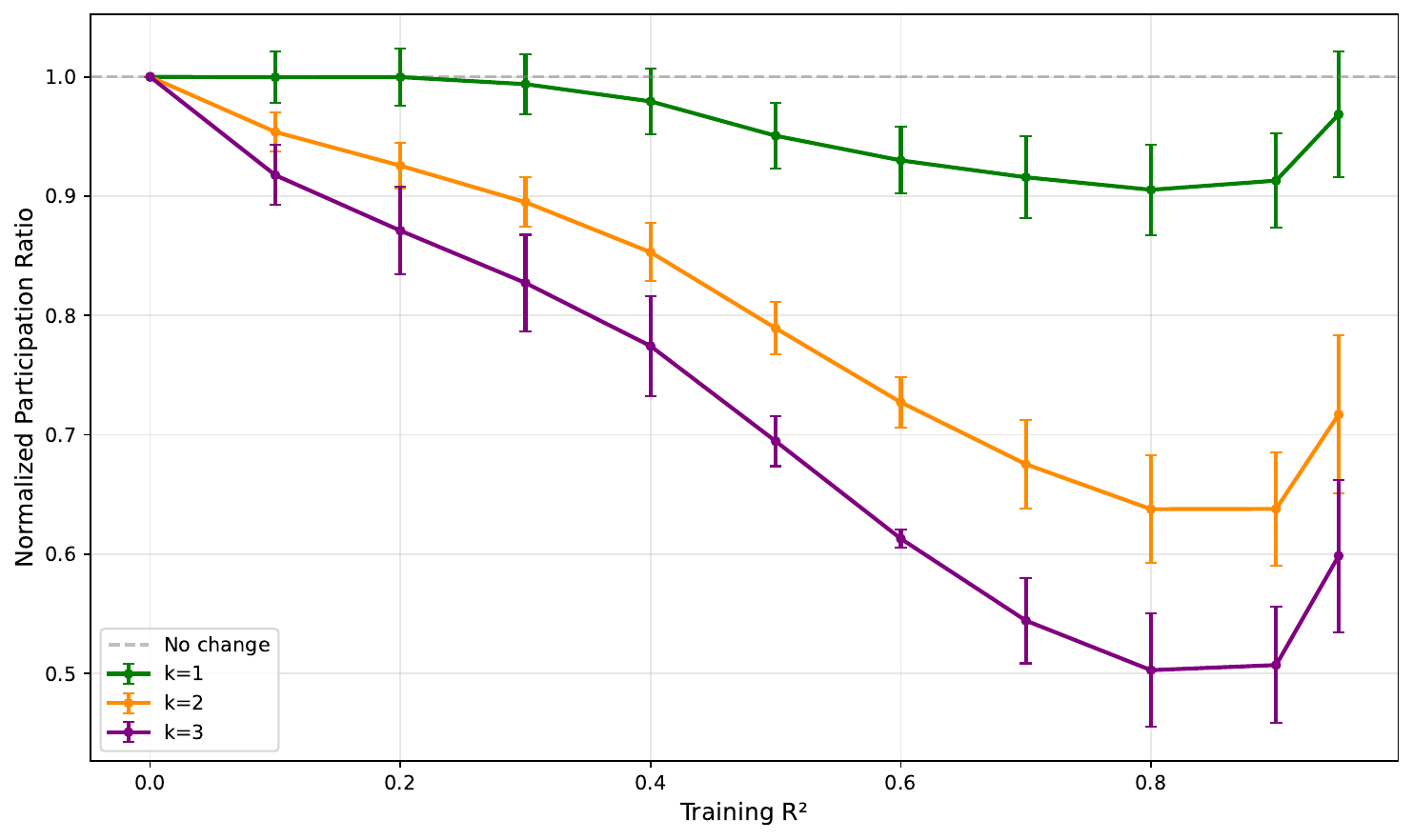}
    \caption{Participation Ratios}
    \label{participationratioevolution}
\end{subfigure}
\hfill
\begin{subfigure}[b]{0.32\textwidth}
    \centering
    \includegraphics[width=\textwidth]{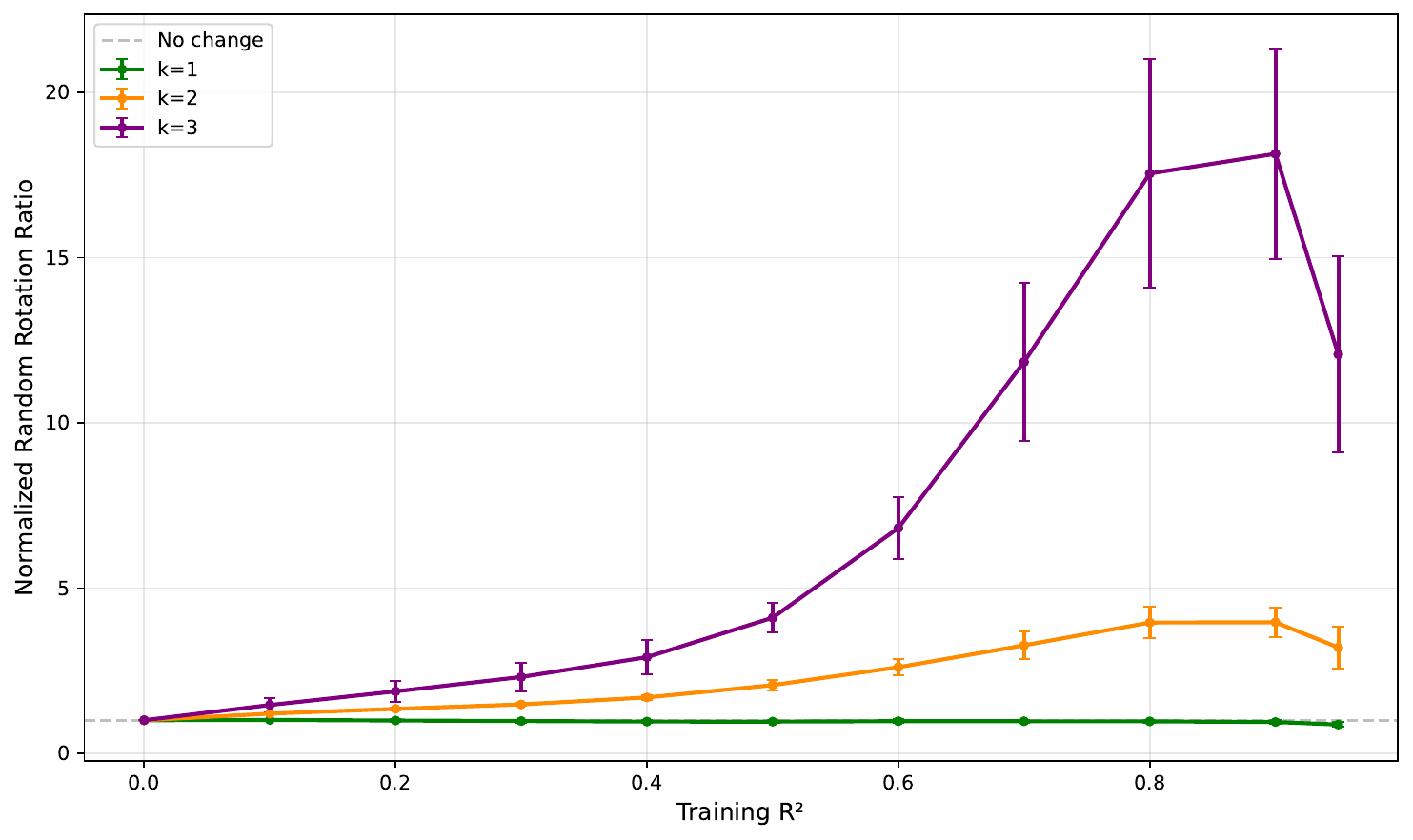}
    \caption{Random Rotation Ratios}
    \label{randomrotationratioevolution}
\end{subfigure}
\hfill
\begin{subfigure}[b]{0.32\textwidth}
    \centering
    \includegraphics[width=\textwidth]{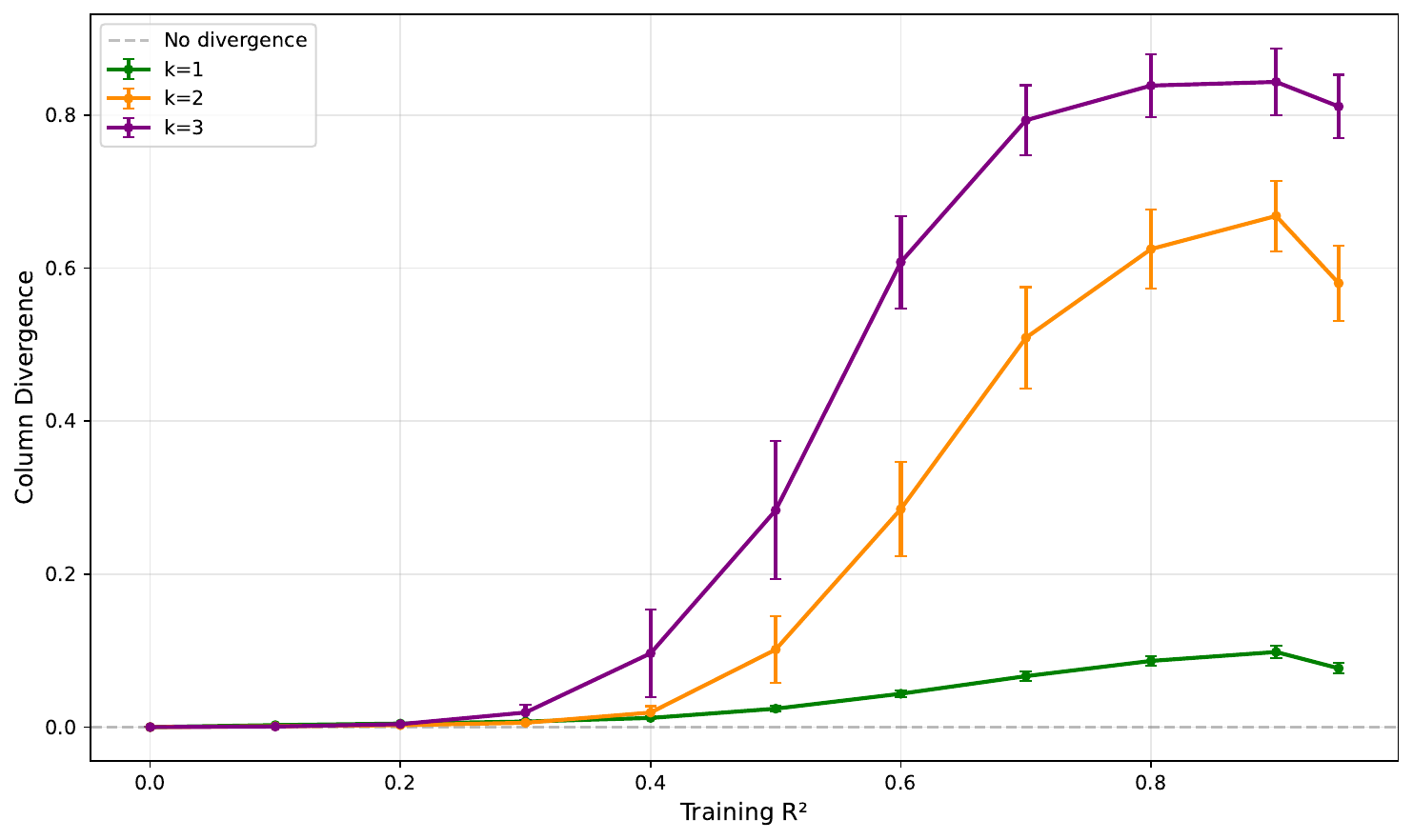}
    \caption{Column Divergences}
    \label{columndivergenceevolution}
\end{subfigure}
\caption{
Evolution over training of size-$k$ participation ratios, random rotation ratios, and column divergences in the \texttt{xor}$(32)$ model.}
\label{participation_permutation_entropy_evolution}
\end{figure}
What about the minor concentration metrics?
Fig.~\ref{participation_permutation_entropy_evolution} shows how the size-$k$ participation ratios, random rotation ratios, and column divergences vary over training.
All three metrics evidence distinct measures for the transition toward a solution with KA geometry around or even slightly before training $R^2 \approx 0.5$.
The decrease with training success of the participation ratio and the increase of the column divergence are (each) reflective of both scale and alignment. 
On the other hand, the increase of the random rotation ratio reflects alignment alone.
Taken together these observations strongly suggest that zero rows and minor concentration are correlated; further, the different aspects of minor concentration---scale and alignment---seem to be likewise correlated.

Let's next see how KA geometry varies with the complexity of the function.
For this we consider the family of functions:
\begin{align}
\label{lambdaxorfunction}
\lambda\texttt{-xor}({\bf x}) & = \prod_{i=1}^n \sin \big(\lambda \pi x_i \big), \quad \lambda \in [0.5, 1.5],
\end{align}
with $n=3$ input dimensions.
$\lambda\texttt{-xor}$ with $\lambda=1.0$ is our original \texttt{xor} function \eqref{xorfunction};
$\lambda\texttt{-xor}$ is easily learned for $\lambda \lessapprox 1.2$ and becomes increasingly difficult to learn for $\lambda > 1.2$ (see Fig.~\ref{training_performance_interp}).
We therefore expect KA geometry to occur when $\lambda \lessapprox 1.2$.
\begin{figure}[h]
\centering
\includegraphics[width=.50\textwidth]{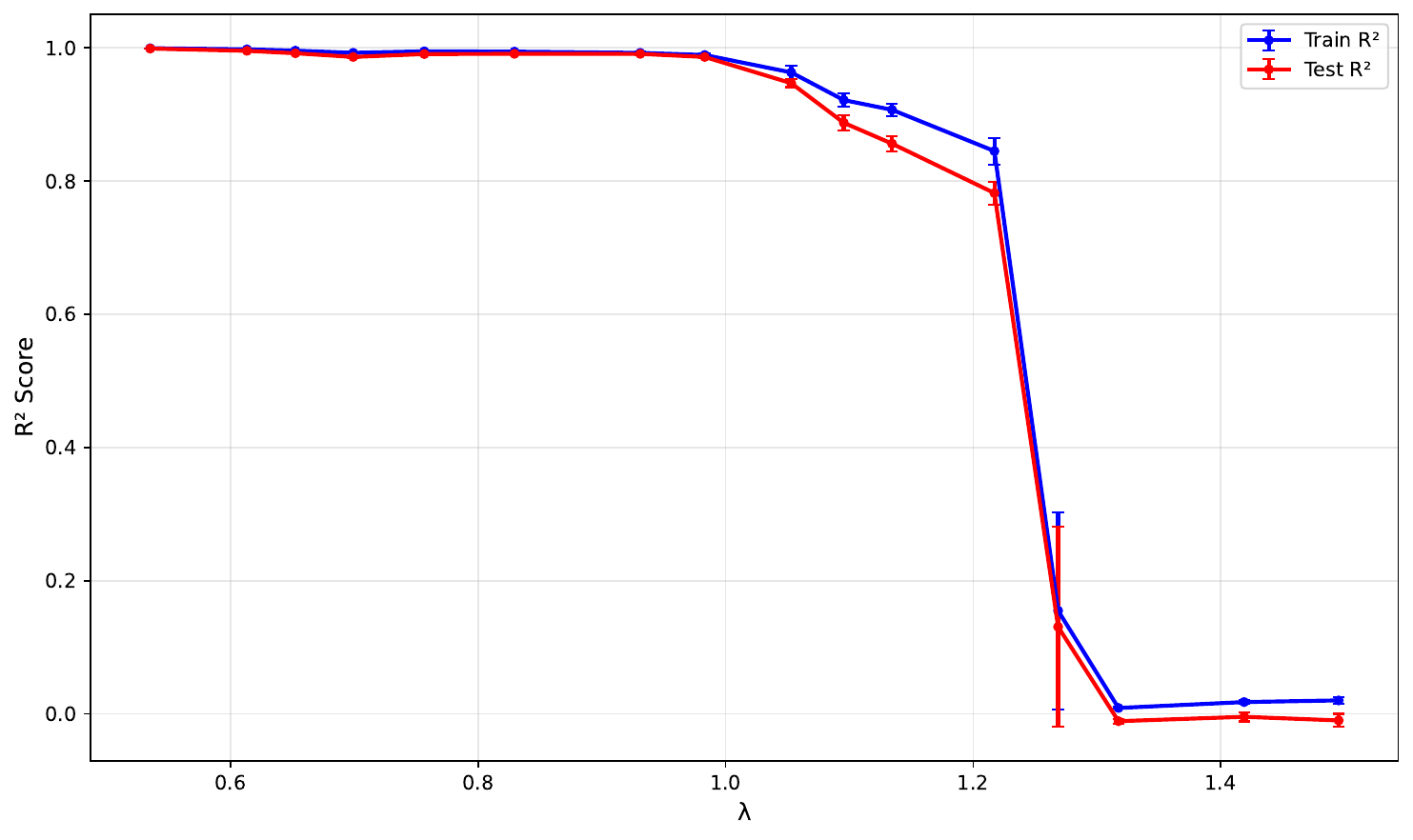}
\caption{
Best training and test $R^2$ of the $\lambda\texttt{-xor}(32)$ models for $\lambda \in [0.5, 1.5]$ achieved through training, averaged over 4 model seeds per $\lambda$.
}
\label{training_performance_interp}
\end{figure}

\begin{figure}[h]
\centering
\begin{subfigure}[b]{0.49\textwidth}
    \centering
    \includegraphics[width=\textwidth]{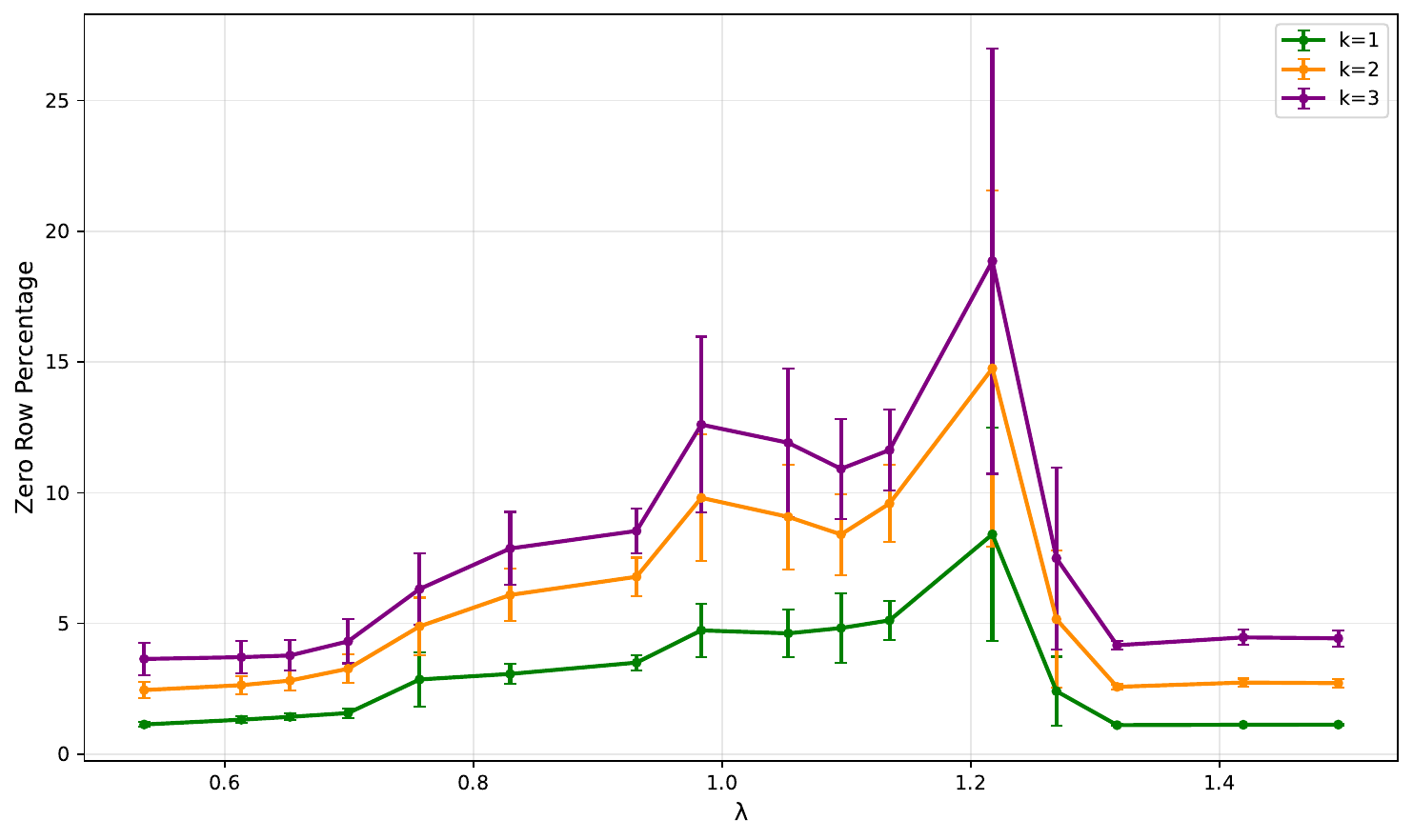}
    \caption{Zero Rows}
    \label{zerointerpolation}
\end{subfigure}
\hfill
\begin{subfigure}[b]{0.49\textwidth}
    \centering
    \includegraphics[width=\textwidth]{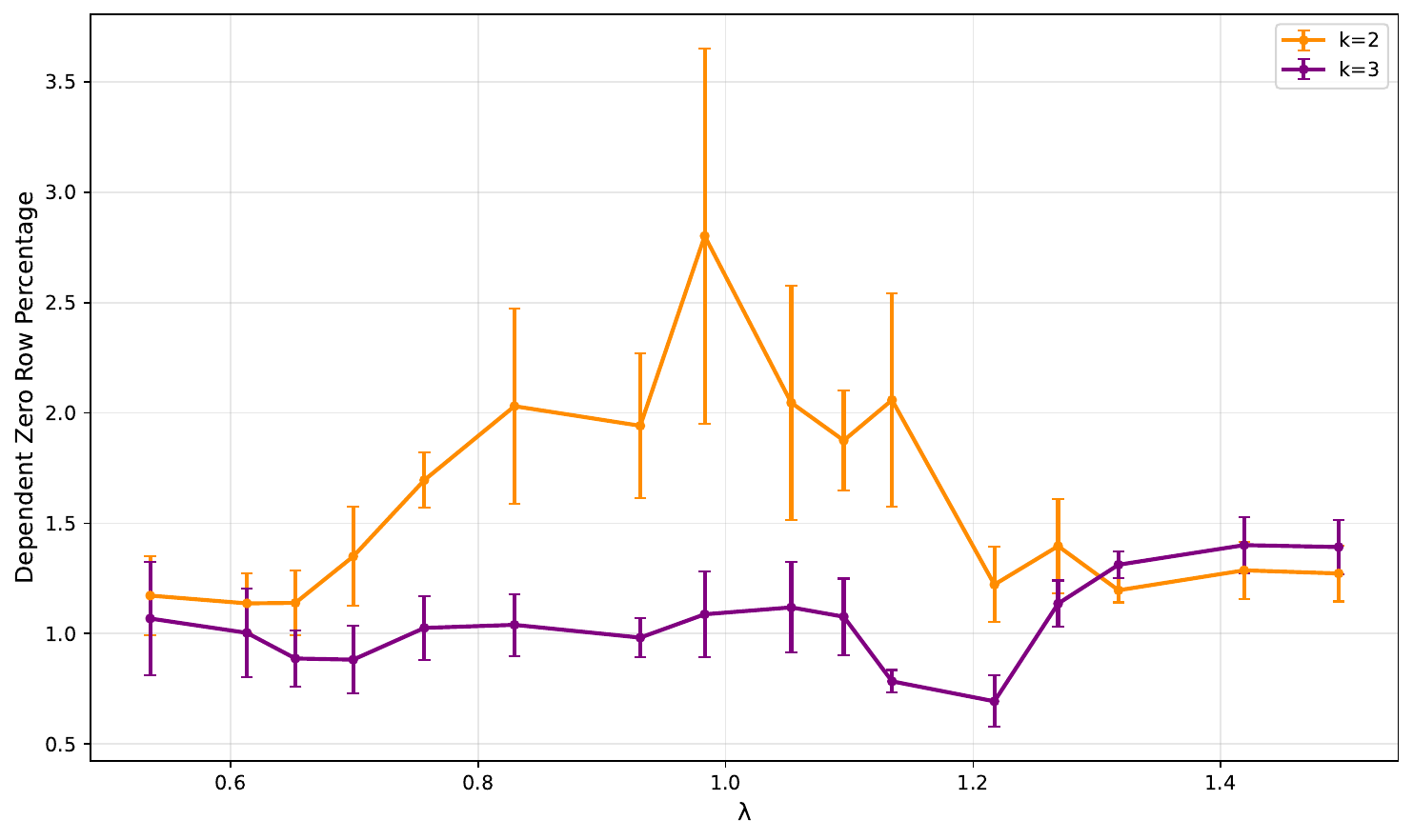}
    \caption{Dependent Zero Rows}
    \label{dependentinterpolation}
\end{subfigure}
\caption{
Interpolation over $\lambda \in [0.5, 1.5]$ of zero row and dependent zero row percentages in the $\lambda\texttt{-xor}(32)$ models.
}
\label{zerorowmetricsinterpolation}
\end{figure}

In Fig.~\ref{zerorowmetricsinterpolation} we plot the zero row metrics. 
We see the development of a noticeable fraction of zero rows for $0.7 \lessapprox \lambda \lessapprox 1.2$ in Fig.~\ref{zerointerpolation}; we likewise notice size-$2$ dependent zero rows appearing in this range of $\lambda$ in Fig.~\ref{dependentinterpolation}, without any discernible size-$3$ dependent zero rows, similar to evolution plots.
The region $0.7 \lessapprox \lambda \lessapprox 1.2$ is a sort of ``Goldilocks regime" for the  $\lambda\texttt{-xor}(32)$ family of models.
The locations of the maxima of the zero and size-$2$ dependent zero row percentages differ, occurring near $\lambda \approx 1.2$ and $\lambda \approx 1.0$.
The large error bars at these maxima make it unclear whether there's definitive maxima or simply broad humps of zero rows about these points.

\begin{figure}[h]
\centering
\begin{subfigure}[b]{0.32\textwidth}
    \centering
    \includegraphics[width=\textwidth]{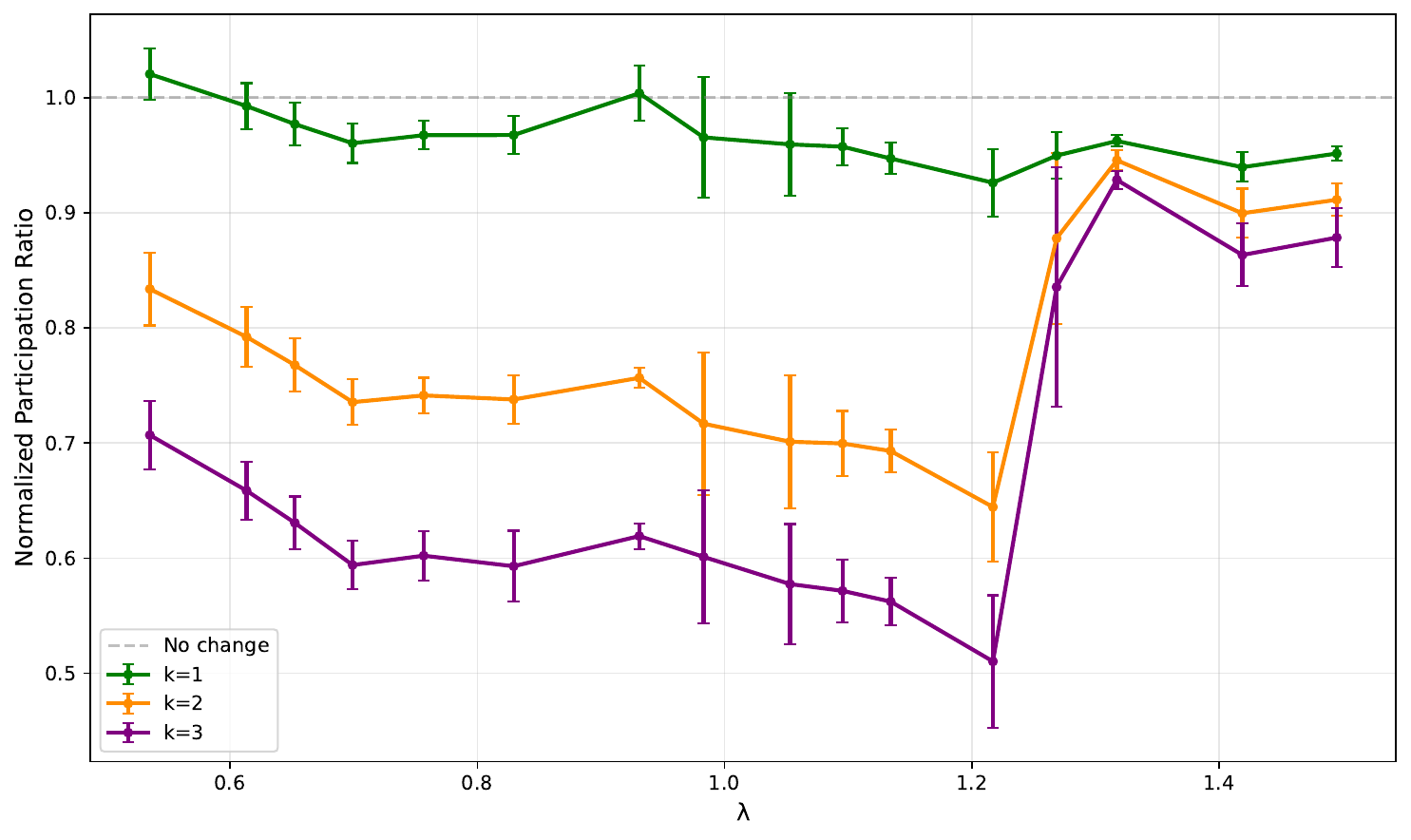}
    \caption{Participation Ratios}
    \label{participationratiosinterpolation}
\end{subfigure}
\hfill
\begin{subfigure}[b]{0.32\textwidth}
    \centering
    \includegraphics[width=\textwidth]{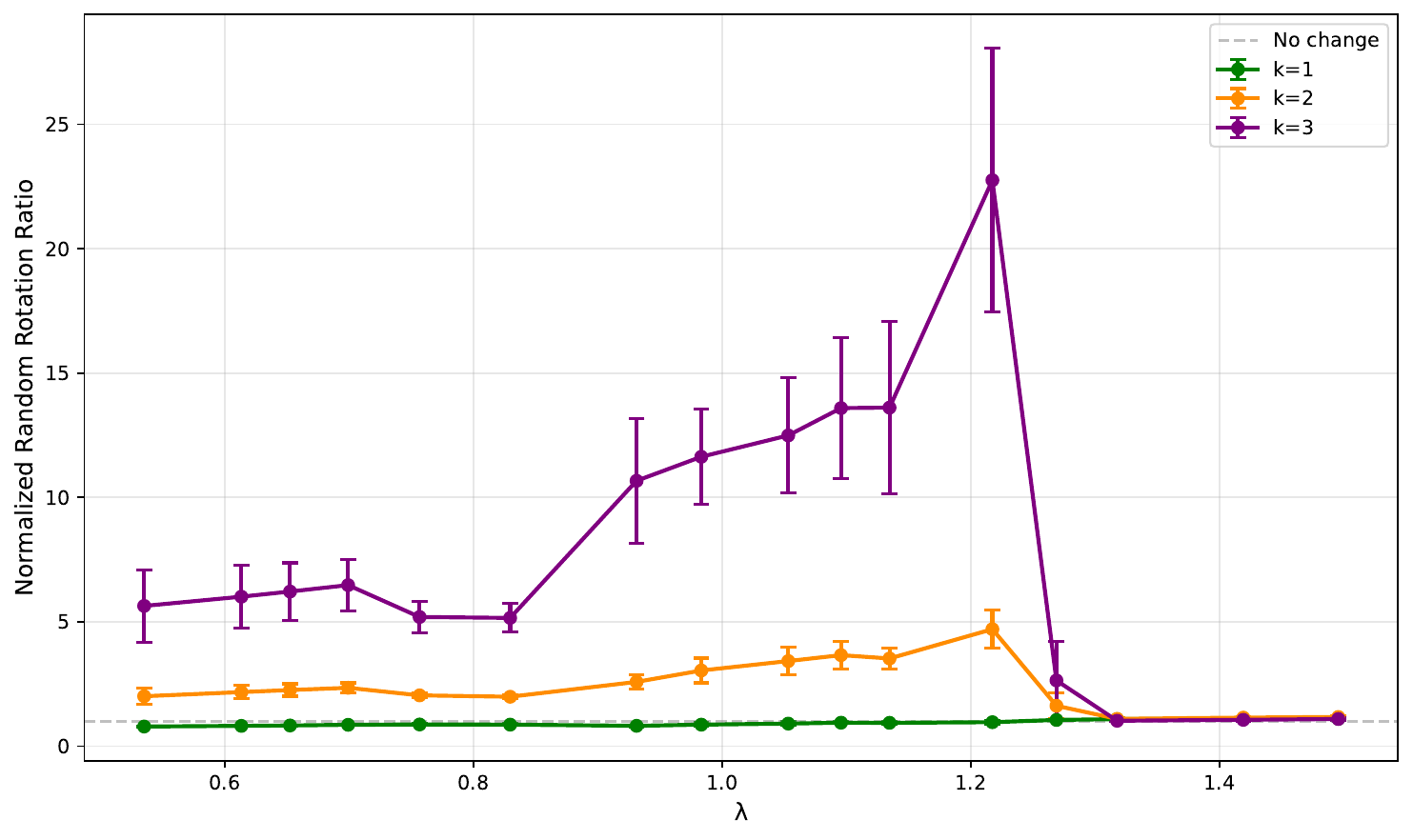}
    \caption{Random Rotation Ratios}
    \label{randomrotationratiosinterpolation}
\end{subfigure}
\hfill
\begin{subfigure}[b]{0.32\textwidth}
    \centering
    \includegraphics[width=\textwidth]{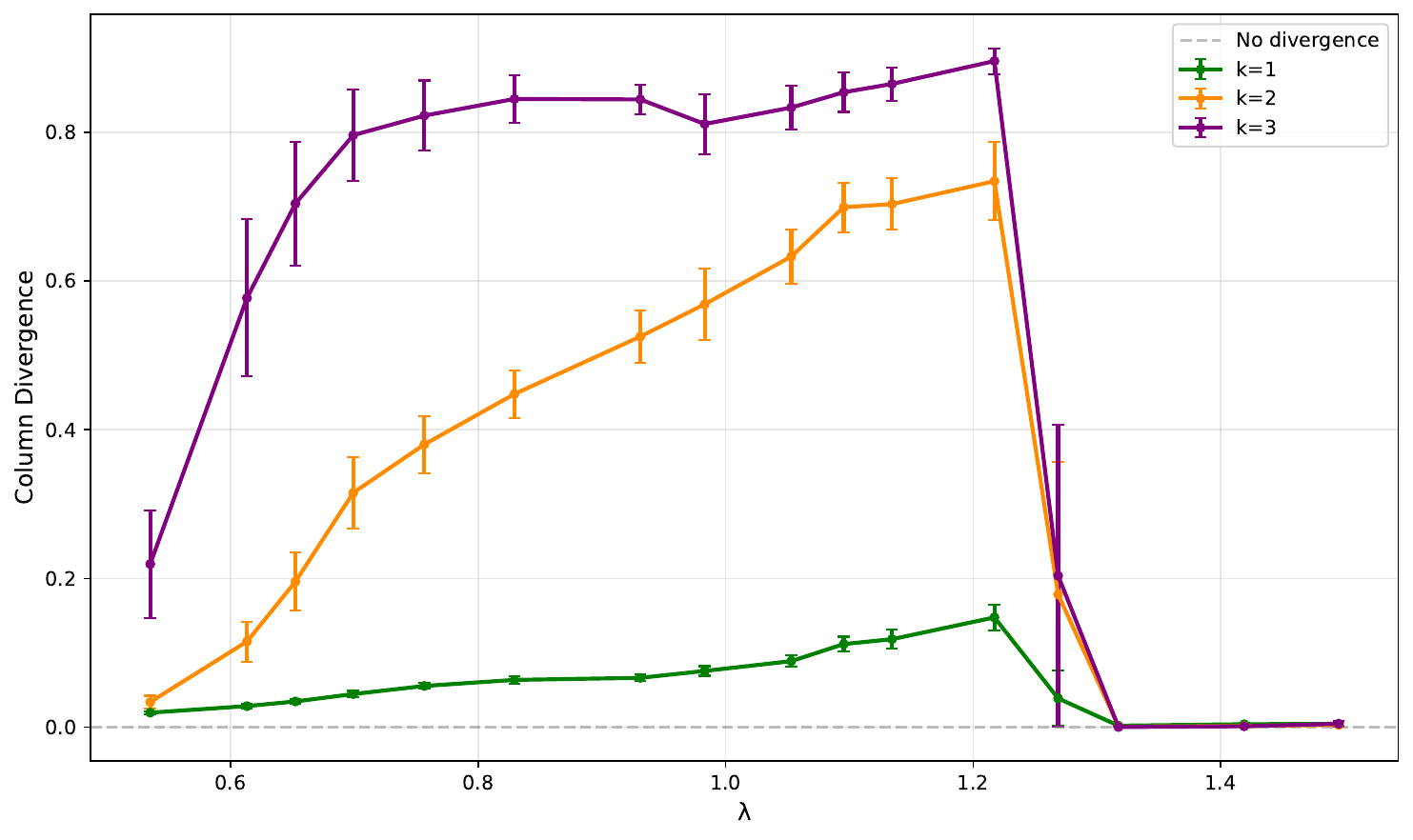}
    \caption{Column Divergences}
    \label{columndivergencesinterpolation}
\end{subfigure}
\caption{
Interpolation over $\lambda \in [0.5, 1.5]$ of size-$k$ participation ratios, random rotation ratios, and column divergences in the $\lambda\texttt{-xor}(32)$ models.}
\label{participation_permutation_entropy_interpolation}
\end{figure}

The minor concentration metrics shown in Fig.~\ref{participation_permutation_entropy_interpolation} seem to be broadly consistent with the zero row metrics: All three metrics suggest noticeable KA geometry for $0.7 \lessapprox \lambda \lessapprox 1.2$.
For $\lambda \lessapprox 0.7$ the minor concentration metrics display modest indications of KA geometry, which decrease as $\lambda \rightarrow 0.5$.
This may be related to the early signal onsets of the participation ratios and column divergences shown in the evolution plots in Figs.~\ref{participationratioevolution} and \ref{columndivergenceevolution} that occurred before the onsets of the zero row metrics in Figs.~\ref{zerorowmetricsevolution}.
We have found in other experiments (not shown) that as $\lambda \rightarrow 0$, KA geometry entirely disappears; in this limit, $\lambda\texttt{-xor}(\mathbf{x}) \rightarrow (\lambda \pi)^n \prod_i x_i$, with the only function nodes appearing when $x_i = 0$.
The extremum near $\lambda \approx 1.2$---minimum for the participation ratio and maximum for the random rotation ratio and column divergence---is seen in all three plots.
It's tempting to correlate these extrema with the extrema found in the minor concentration evolution plots in Fig.~\ref{participation_permutation_entropy_evolution} near training $R^2 \approx 0.85$, roughly the same training $R^2$ achieved at $\lambda \approx 1.2$ (see Fig.~\ref{training_performance_interp}).
We don't understand what causes this extremum away from the location of peak performance.
(KA geometry does not guarantee future performance.)
This structure may indicate that the model is attempting to develop a universal inner map that the linear outer function (for our 1-hidden layer model) is unable to capitalize upon; see Appendix \ref{batchsizeappendix} for a related discussion.

\section{Discussion}

We have laid foundations for observing KA-like geometry as it evolves in laptop-sized models---where (nearly) every relevant piece of data can be processed.
Our goal, only just begun at the time of writing, is to then validate these geometric findings in large models.
Large models, of course, present a combinatorial explosion of places to look for KA: Jacobians between which pair of layers; minors of which size $k$; training may be in-parallel or layer by layer.
In large models, data can only be sampled, not exhausted, so one must guard against possible selection biases.
Also large models come in a plethora of flavors: vanilla fully-connected neural networks, convolutional nets, recurrent neural nets, attention-enhanced nets, neural nets coupled to tools, etc.
Despite these difficulties of scale and complexity, we expect to begin the work of mapping out the ``KA phase diagram'' showing the how, when, and where of KA-like geometry emerging through training.

Of course, our ultimate goal is not merely observation, but intervention.
As we learn how KA geometric patterning correlates with learning, it will be tempting to accelerate learning by nudging such patterning forward---at the appropriate layers and stages in the training process.
It appears on first sight fantastically costly to compute the ``gradient of KA-ness'' and add it to the loss function during training.
So, even if we figure out when and where we should be pushing toward KA geometry, it might be too costly to carry out the necessary computations as we train.
This is our greatest concern.

But there may be cheap tools already in the community's tool kit.
Oscillatory activation functions, e.g., $x+0.1\sin(10x)$, have been used \cite{2021arXiv210812943M} with mixed results.
It is easy to see that the coordinate-wise stretching/compressing such functions induce, on average, increase Jacobian coordinate alignment, and thus minor concentration.
But forcing coordinate alignment will have its time and place; doing so prematurely may thwart the evolution of a useful abstraction; there may not yet be the requisite data to support the abstraction.
We expect that if one we can answer the ``where" and ``when" for the spontaneous emergence of KA-like geometry, we can use computationally cheap tools (like activation functions with periodic terms) to accelerate such emergence in the appropriate places.
If, as we believe, KA-like geometry emerges to codify a developing abstraction, such as recognizing a feature, or correlating two phrases, it will have a time and place.
It may be that a simple trick such as oscillatory activation is all we need---provided we know when and where to apply it (i.e., know the phase diagram).
We hope to explore both the emergence of KA in large models, and the effectiveness of such interventions in a sequel.

\acknowledgements

We thank Vitaly Aksenov, Eve Bodnia, and Vlad Isenbaev for useful discussions.
We are particularly grateful to Vitaly Aksenov for extensive comments on an early draft.
% This work was supported by the U.S.~Department of Energy, Office of Science, Office of Basic Energy Sciences, under Award No.~DE-SC0020007. 

\appendix 

\section{Fitting the Distribution Tails}
\label{tailfittingappendix}

Recall Fig.~\ref{minordistributioncomparison}, which compares the size-$3$ minor distributions of the \texttt{xor}$(32)$, \texttt{linear}$(32)$, and \texttt{random}$(32)$ models.
Clearly, the (trained) \texttt{xor}$(32)$ model has larger magnitude outliers than at initialization (in addition to its greater peakedness near zero).
For example, the ratio of the $0.99$ to $0.01$ quantiles of size-$3$ minors in \texttt{xor}$(32)$ is astronomical, $10^{45}$ vs $10^4$ at initialization.
(We interpret this as blow-up to the behavior of the KA inner function, which requires quite delicate construction to have better than H\"older regularity.)
Might the relatively heavy tail behavior of the \texttt{xor}$(32)$ model be described by a universal scaling, e.g., power-law or exponential?
Our answer, explained in this appendix, is no. 

\begin{figure}[h]
\centering
\begin{subfigure}[b]{0.32\textwidth}
    \centering
    \includegraphics[width=\textwidth]{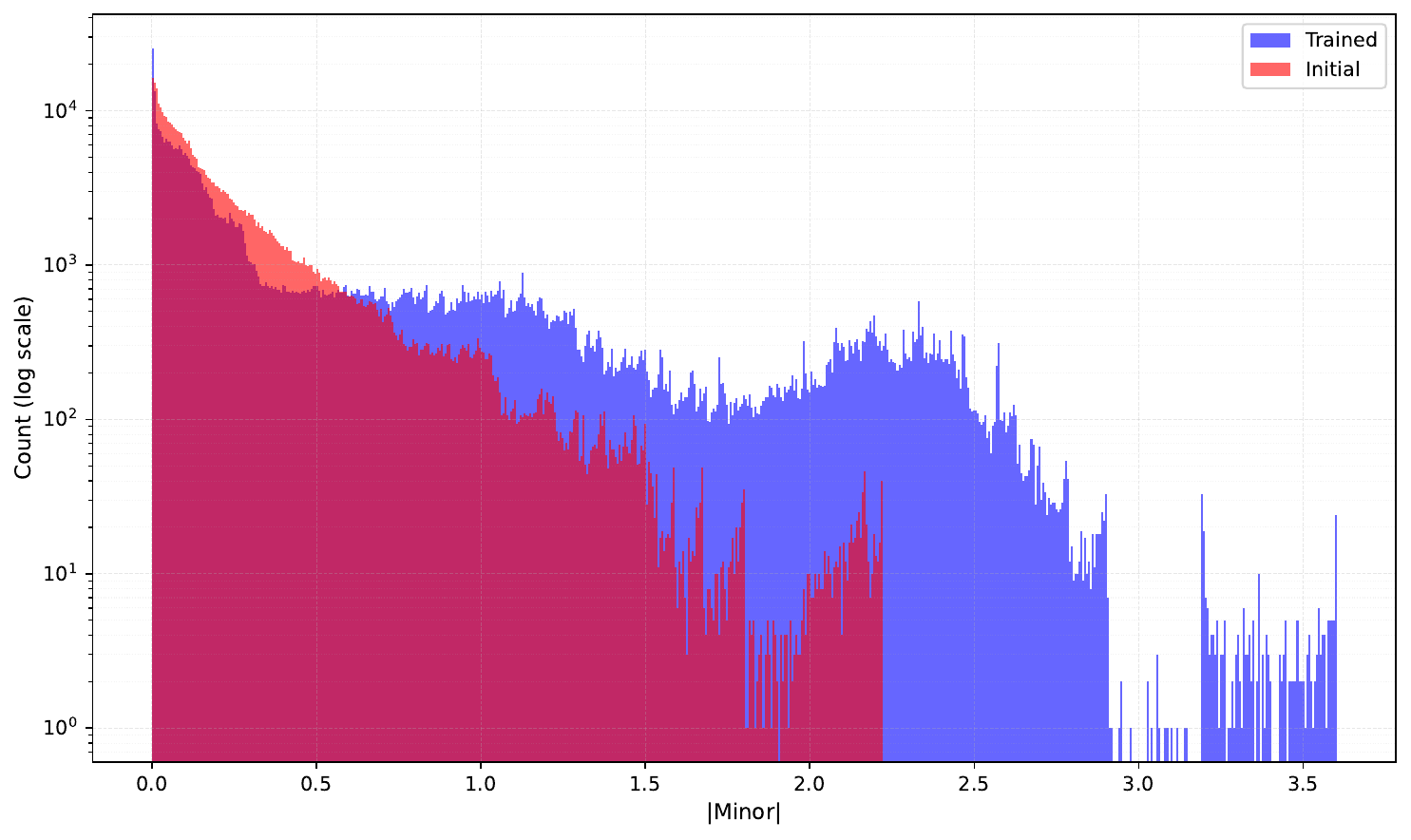}
    \caption{$k=1$}
    \label{xork1distribution}
\end{subfigure}
\hfill
\begin{subfigure}[b]{0.32\textwidth}
    \centering
    \includegraphics[width=\textwidth]{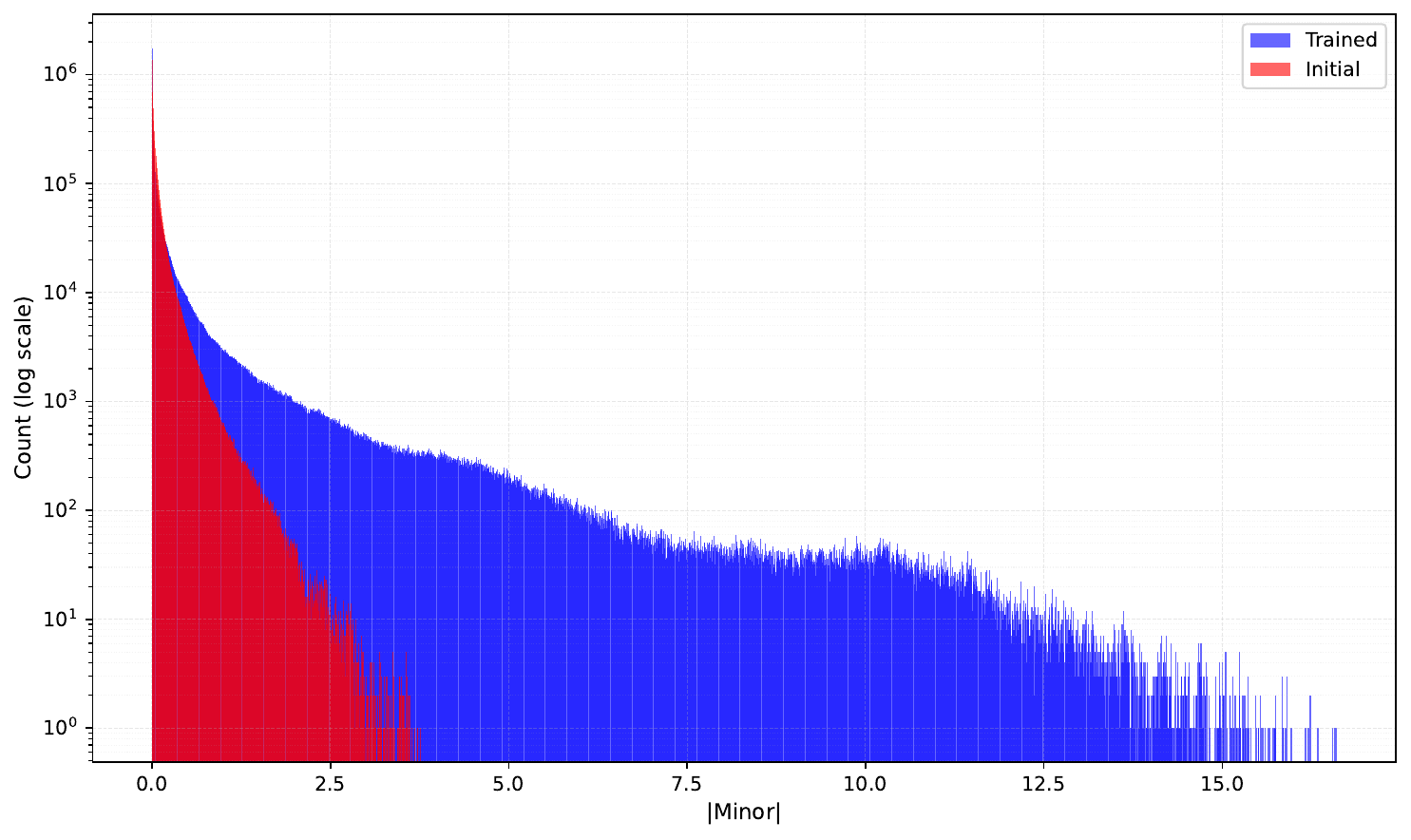}
    \caption{$k=2$}
    \label{xork2distribution}
\end{subfigure}
\hfill
\begin{subfigure}[b]{0.32\textwidth}
    \centering
    \includegraphics[width=\textwidth]{minor_distributions_xor_h32_k3_train}
    \caption{$k=3$}
    \label{xork3distribution}
\end{subfigure}
\caption{
Size-$k$ minor distributions of the \texttt{xor}$(32)$ model.}
\label{xorminordistributions}
\end{figure}
Our reasoning goes as follows.
We consider the distributions of size-$k$ minors of the \texttt{xor}$(32)$ model in Fig.~\ref{xorminordistributions}.  
Given their similarity, we take the model at initialization to nominally represent the \texttt{linear}$(32)$ model and \texttt{random}$(32)$ models as well (see Fig.~\ref{minordistributioncomparison}).
The tail behaviors of these distributions can be inferred with the mean excess function (MEF). 
For a random variable $X$, the mean excess function is the conditional expectation, ${\rm MEF}(u) = E(X - u | X > u)$, where ``$E$" denotes expectation and $u \in \mathbb{R}$.
An exponential distribution has constant MEF; a power-law distribution has increasing MEF.

\begin{figure}[h]
\centering
\begin{subfigure}[b]{0.32\textwidth}
    \centering
    \includegraphics[width=\textwidth]{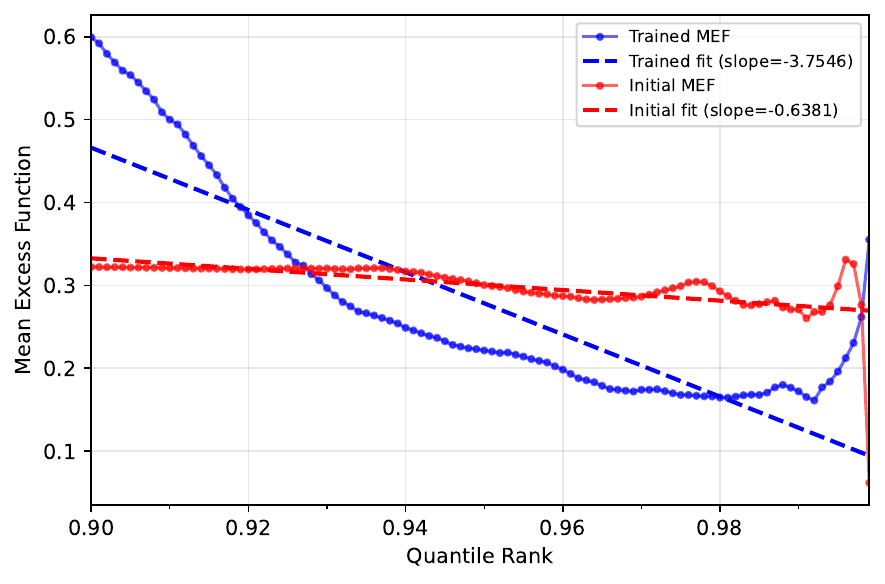}
    \caption{$k=1$}
    \label{xork1distribution}
\end{subfigure}
\hfill
\begin{subfigure}[b]{0.32\textwidth}
    \centering
    \includegraphics[width=\textwidth]{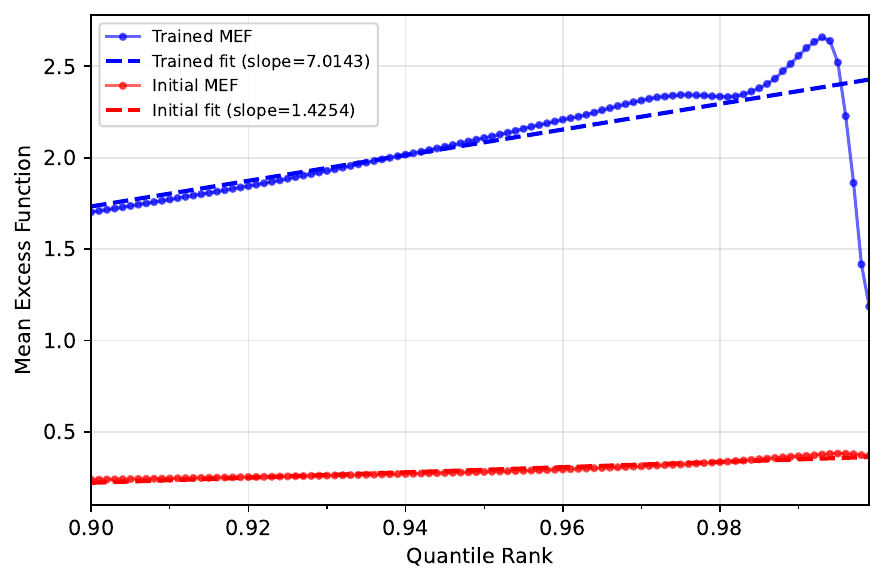}
    \caption{$k=2$}
    \label{xork2distribution}
\end{subfigure}
\hfill
\begin{subfigure}[b]{0.32\textwidth}
    \centering
    \includegraphics[width=\textwidth]{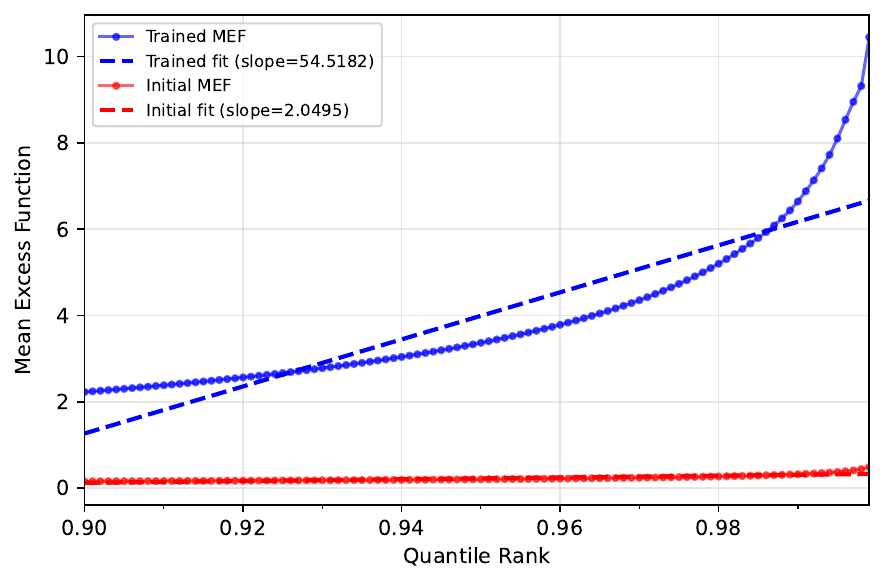}
    \caption{$k=3$}
    \label{xork3distribution}
\end{subfigure}
\caption{
Mean excess functions (MEFs) of size-$k$ minors of the \texttt{xor}$(32)$ model.
The dashed lines are linear fits.
}
\label{mefplots}
\end{figure}
We computed the MEFs of the \texttt{xor}$(32)$ model empirically, with example results given in Fig.~\ref{mefplots}.
Note that in order to compare the trained and initial distributions of each size-$k$ minor, which have differing spans (e.g., the trained model's size-$3$ minor maximum is approximately an order of magnitude larger than the maximum at initialization), the MEFs are plotted with respect to quantile rank $q$, i.e., $u = Q(q)$ for $q \in [.90, 1.0)$ and we plot ${\rm MEF}\big(Q(q)\big)$, where $Q(q)$ denotes the value of the quantile.

Fig.~\ref{mefplots} indicates that the tail behaviors vary with $k$ and the precise range over which the analysis is performed (e.g., top $1\%$ or top $10\%$ of values). 
In general, both distributions are lighter-than-exponential tailed at $k=1$; the initial size-$k$ minors are somewhat heavy-tailed for $k=2,3$ with polynomial-decaying tails, while the size-$k$ minors of the trained model are heavier-tailed than at initialization for $k >1$.
(This assumes we excise the highest quantile ranks at $k=2$; we believe the decay at high rank is a finite sample effect.)

\section{Batch Size Dependence and Latent KA Geometry in the \texttt{random} Model}
\label{batchsizeappendix}

In this appendix, we look at how KA geometry depends on batch size.
All the results in the main sections of the paper were performed using the approximate critical batch size \cite{2018arXiv181206162M} of the \texttt{xor} model.
(We took batch size $B=250$, the upper critical batch size across all hidden dimensions $m \in \{4, 8, 16, 32 \}$.)
While unnecessary when training these small models, the critical batch size is useful when training larger models and optimizing both compute and time.

We have found the KA metrics to be relatively uniform across batch sizes in the \texttt{xor}$(32)$ and \texttt{linear}$(32)$ models.
The metrics are nearly exactly flat in the \texttt{linear}$(32)$ model, while there are larger variations across batch sizes in the \texttt{xor}$(32)$ model.
As far as we can tell, the only noteworthy effect that occurs in the \texttt{xor}$(32)$ model is the appearance of consistently zero rows, i.e., a row that's zero across at least $99\%$ of examples, for small batch sizes $B=32$ and $B=64$. 
The percentage of not-consistently zero rows (the example-dependent patterning of the Jacobian that reflects KA geometry), however, is about the same across all batch sizes.

\begin{figure}[h]
\centering
\begin{subfigure}[b]{0.45\textwidth}
\centering
\includegraphics[width=\textwidth]{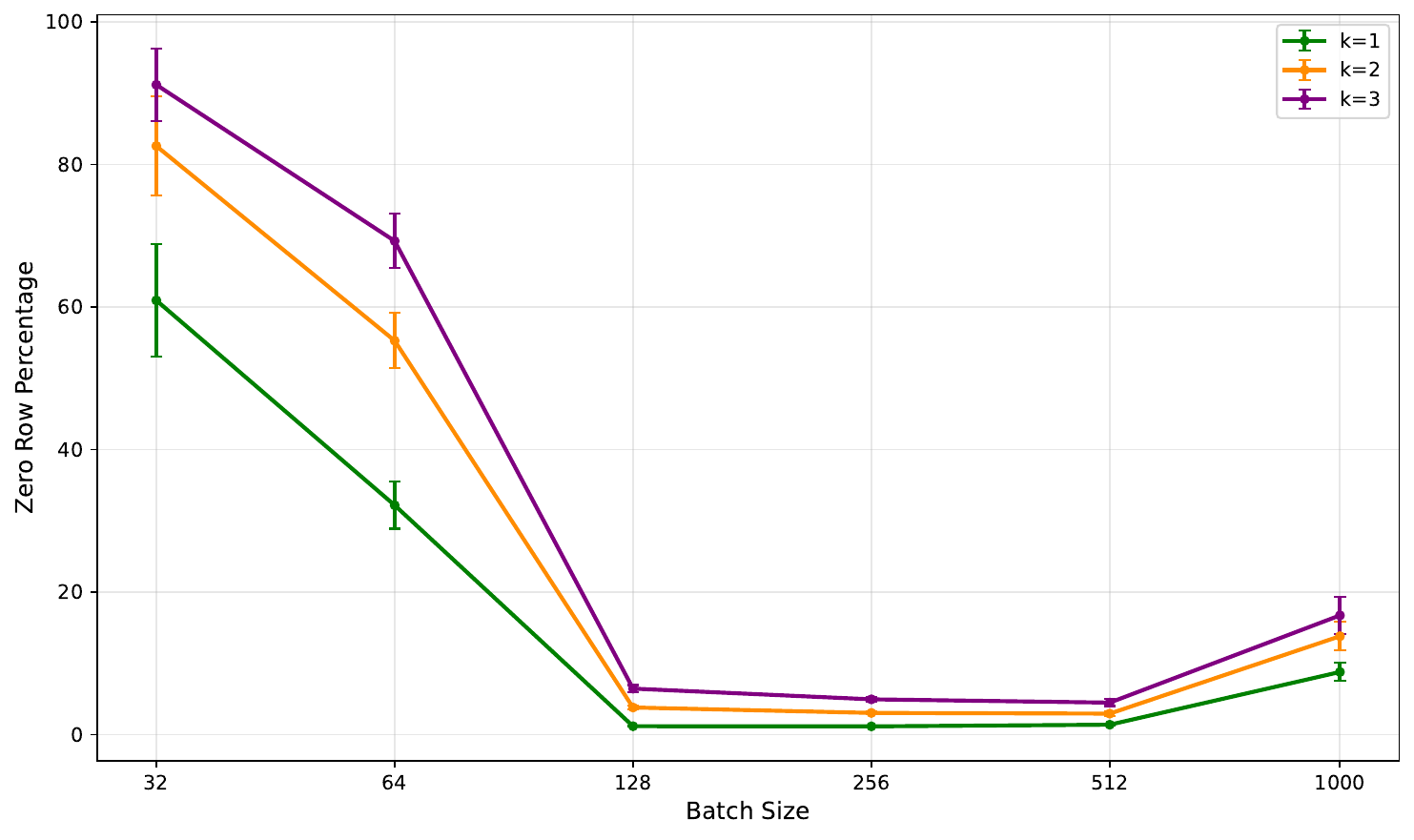}
\caption{Zero Rows}
\label{random_batch_zero_rows}
\end{subfigure}
\hfill
\begin{subfigure}[b]{0.45\textwidth}
\centering
\includegraphics[width=\textwidth]{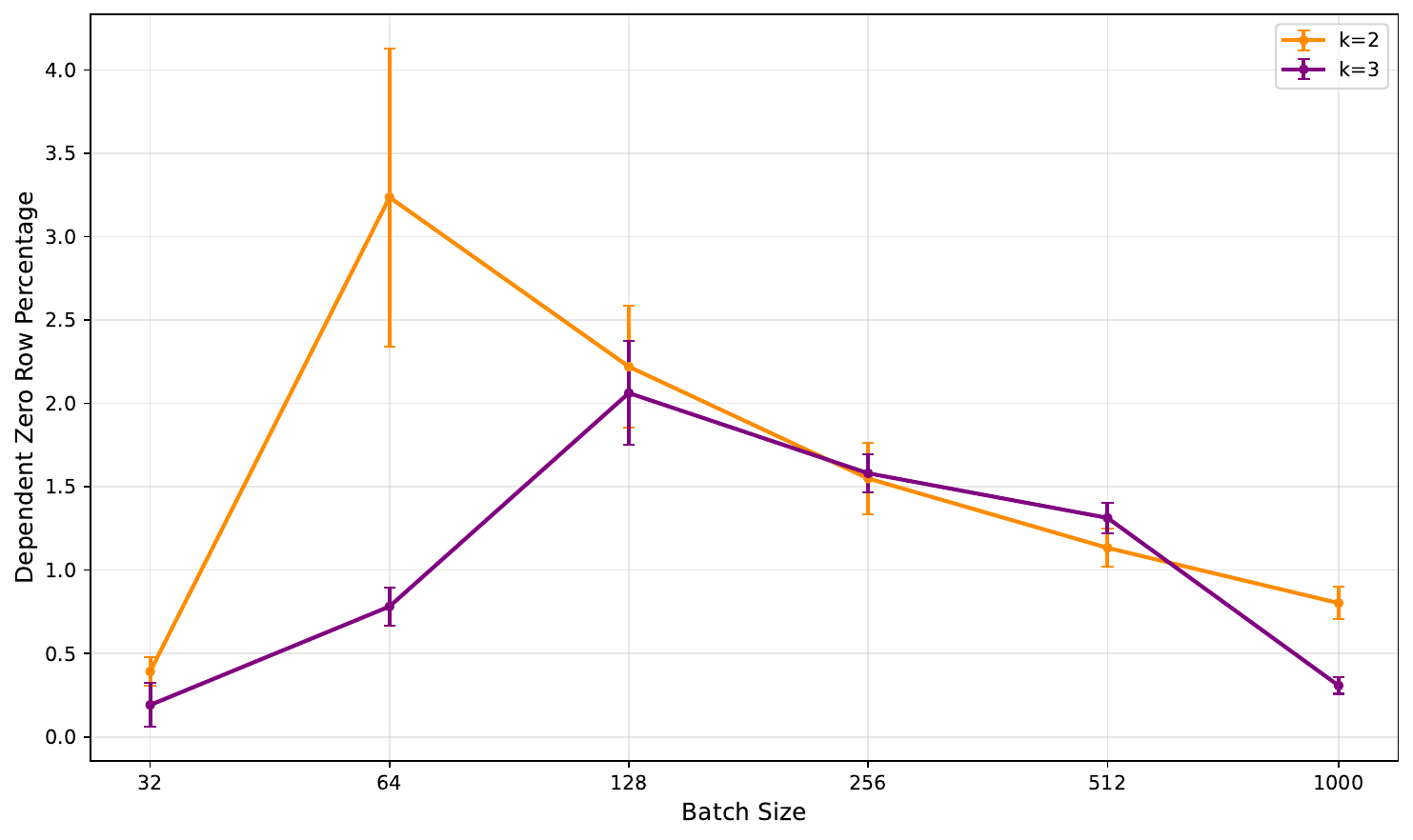}
\caption{Dependent Zero Rows}
\label{random_batch_dependent_zeros}
\end{subfigure}
\vspace{1em} % Adds a little vertical space between the rows
\begin{subfigure}[b]{0.32\textwidth}
    \centering
    \includegraphics[width=\textwidth]{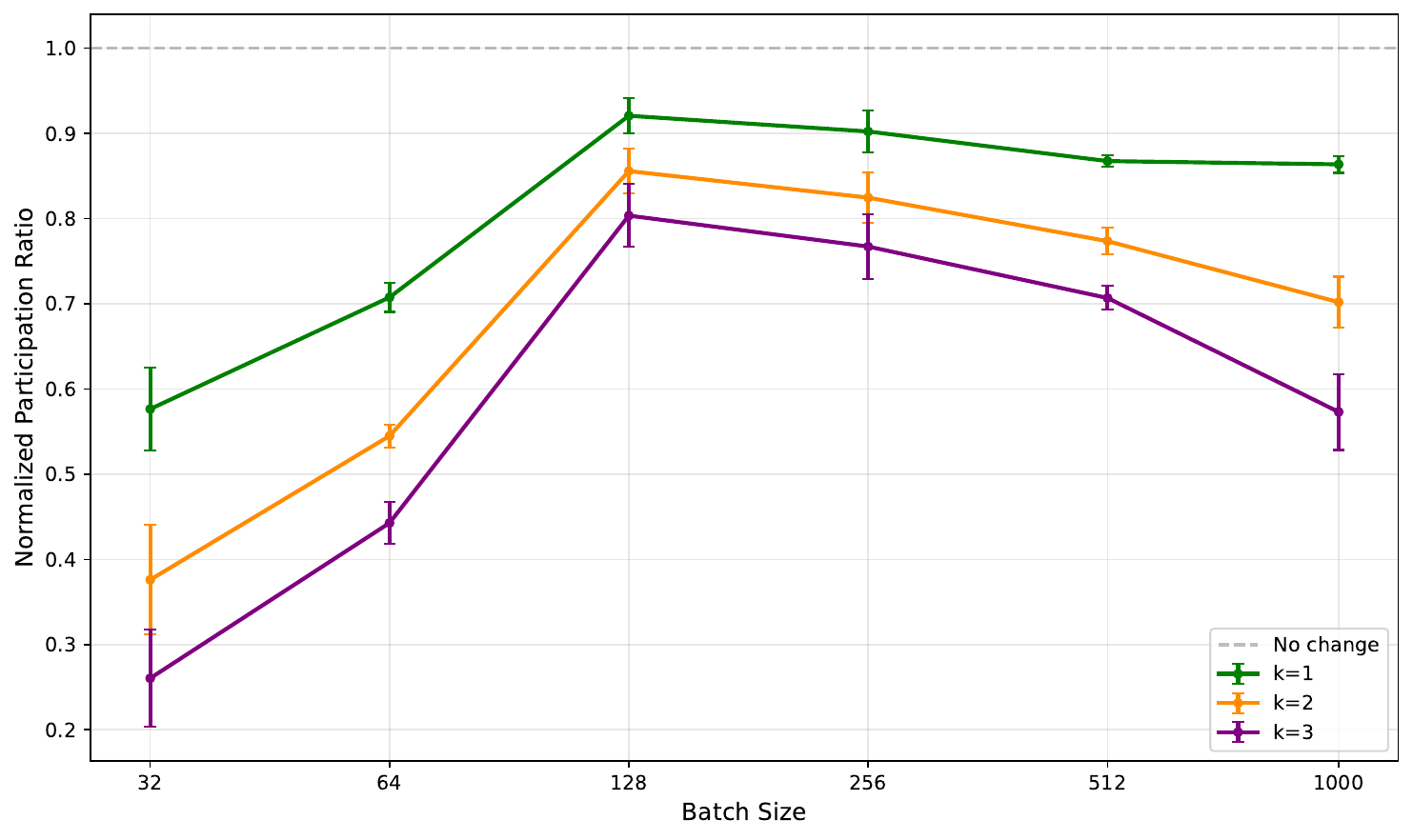}
    \caption{Participation Ratios}
    \label{random_batch_participation_ratios}
\end{subfigure}
\hfill
\begin{subfigure}[b]{0.32\textwidth}
    \centering
    \includegraphics[width=\textwidth]{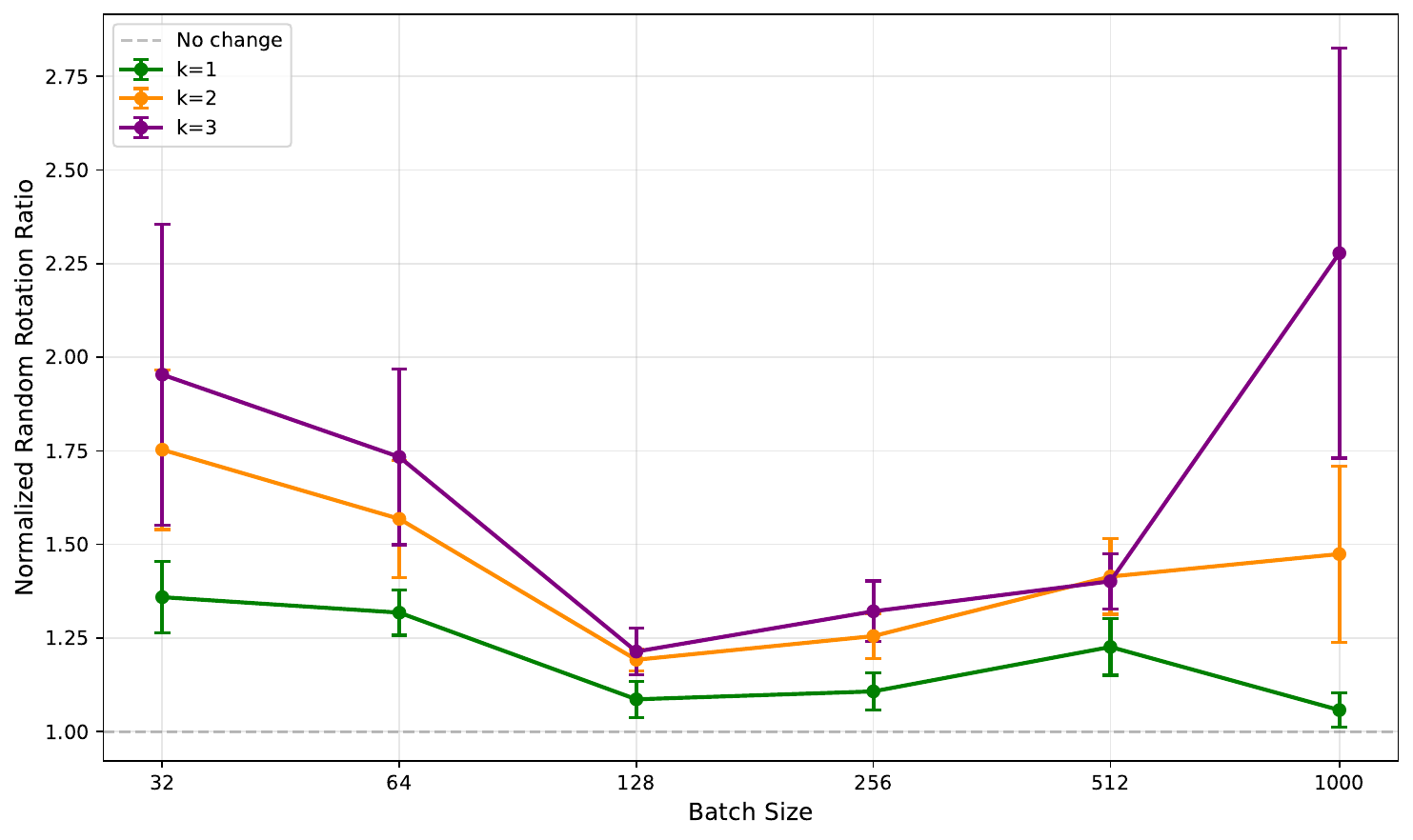}
    \caption{Random Rotation Ratios}
    \label{random_batch_random_rotation_ratios}
\end{subfigure}
\hfill
\begin{subfigure}[b]{0.32\textwidth}
    \centering
    \includegraphics[width=\textwidth]{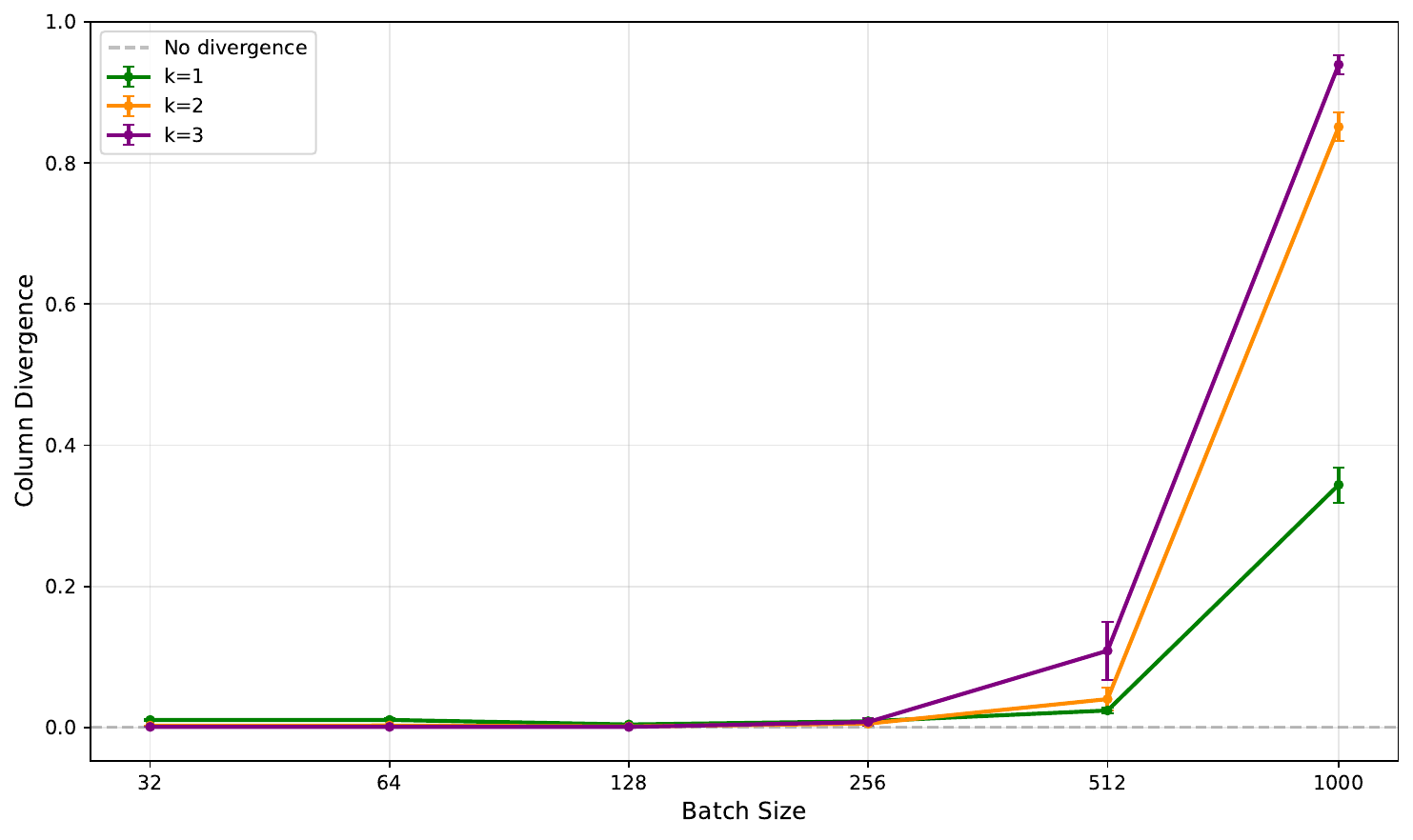}
    \caption{Column Divergences}
    \label{random_batch_column_divergences}
\end{subfigure}
\caption{
Zero row and minor concentration metrics in the \texttt{random}$(32)$ model for batch sizes $B = \lambda \in \{32, 64, 128, 256, 512, 1000\}$.
}
\label{random_batch_interpolation}
\end{figure}

The behavior of the \texttt{random}$(32)$ model is more interesting (see Fig.~\ref{random_batch_interpolation}): There is a non-monotonic dependence in which the \texttt{random}$(32)$ model seems to exhibit a weak KA geometry signal for small batch sizes, zero signal for the critical batch size (as already shown in the main sections of the paper), and nearly indistinguishable signal (from the \texttt{xor}$(32)$ model) at full batch.
Why do we call the signal at small batch sizes weak?
Consider the zero row percentages in Fig.~\ref{random_batch_zero_rows}, for example, in which the measured number of zero rows is highest for the smallest batch size $B=32$.
This signal is a weak expression of KA geometry because almost all of the measured zero rows are consistently zero:
About $97\%$ and $95\%$ of zero rows in the size-$1$ minor matrix are consistently zero when $B=32$ and $B=64$; and there are no consistently zero rows for $B \in \{128, 256, 512, 1000 \}$.

We don't really understand why training, both \texttt{random}$(32)$ and \texttt{xor}$(32)$ models, at small batch sizes produces so many consistently zero rows, however, it seems like the models have abandoned the resources they would need to learn.
We find that the rows that are consistently zero exactly coincide with the smallest rows of the weight matrix.
Here, we're defining the size of a row in the weight matrix as the mean absolute value of a row element.
The Jacobian formula \eqref{jacobianformula} and this coincidence implies that the consistently zero rows are caused by small rows in the weight matrix.
These consistently zero rows therefore correspond to ``dead neurons."
(As a check, we have also looked at the biases of the first-layer map.
Recall the ${\rm GeLU}$ activation has approximately vanishing value and slope when its argument is negative.
Large in magnitude negative biases could be an alternative source of consistently zero rows. 
We find such biases differ in count from the number of consistently zero rows and generally occur on different neurons, e.g., about $50\%$ of the consistently zero rows coincide with neurons receiving a negative bias. This means that such biases could at most only support the weight matrix in creating consistently zero rows.)

How can we understand the appearance of KA geometry in the zero row and minor concentration metrics at the largest batch size?
Our hypothesis is that the inner map of the \texttt{random}$(32)$ model is developing KA geometry (since our metrics are explicitly showing this), however, the linear outer map of the 1-hidden layer model lacks the capacity to memorize the target function; if we endow the outer map with greater capacity, the enhanced model (referred to as the bootstrapped model below) should have a better chance at memorization.
(For a fixed set of training points, the \texttt{random} function can be thought of as a sufficiently high degree polynomial.)

To test this hypothesis, we train a base $1$-hidden layer MLP \eqref{modeldef} (with $m=32$ hidden neurons) in our usual way, extract its first-layer map, $\sigma\big( \mathbf{x} \cdot A + a \big)$, and use it as the fixed $m$-dimensional input to a second $1$-hidden layer MLP, which we then proceed to train from scratch on (input, output) pairs $\big( \sigma\big( \mathbf{x} \cdot A + a \big), \texttt{random}(\mathbf{x}) \big)$.
This second $1$-hidden layer MLP is the replacement of our original linear outer map; the first-layer map of the base model plays the role of a feature extractor.
We call this composite model the bootstrapped $1$-hidden layer model.
It's important to stress that at no time is there more than a $1$-hidden layer model being optimized.

\begin{figure}[h]
\centering
\includegraphics[width=.50\textwidth]{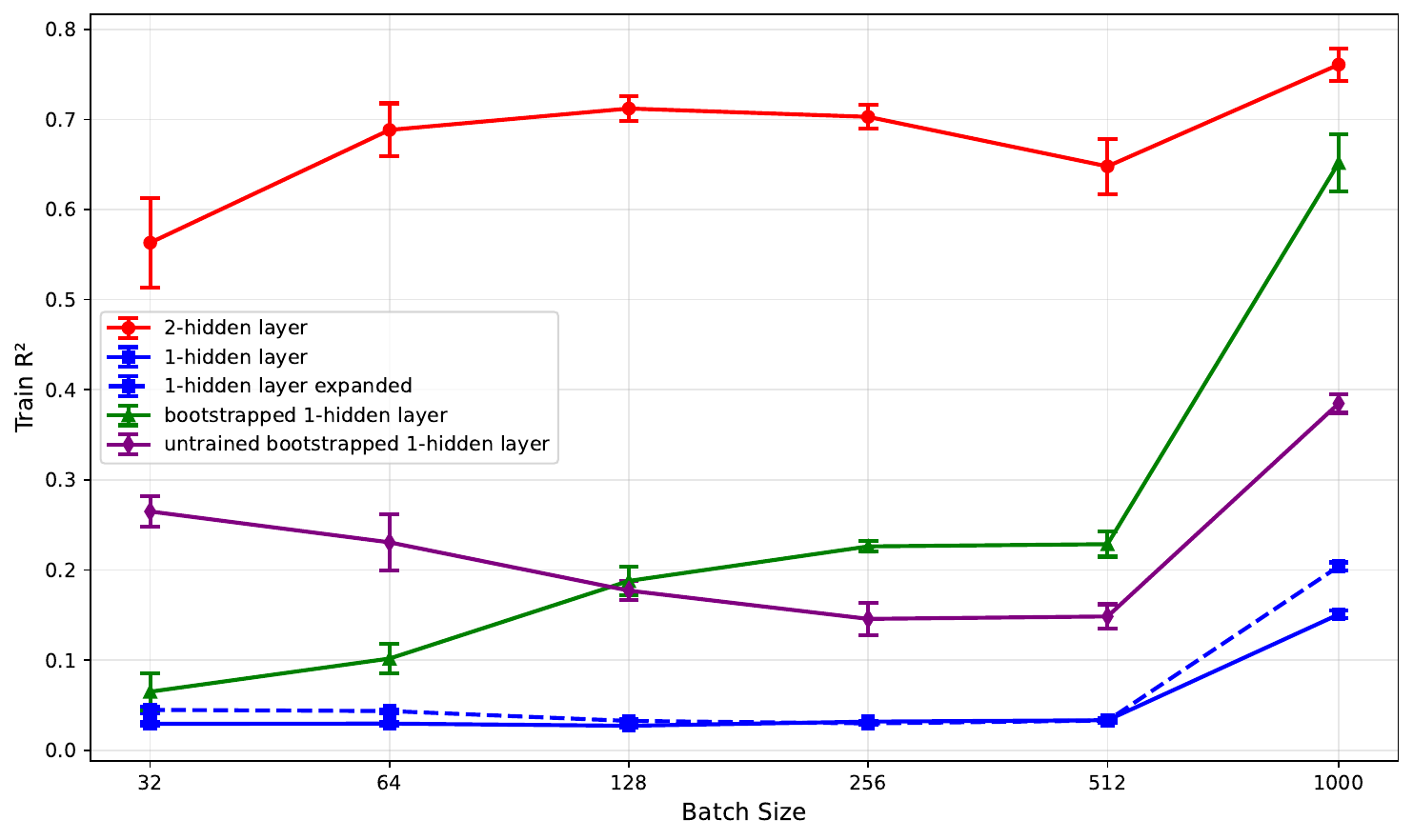}
\caption{
Best training $R^2$ of the \texttt{random}$(32)$ model over batch sizes $B \in \{ 32, 64, 128, 256, 512, 1000 \}$.
The test $R^2 \lessapprox 0.0$ for all models.
}
\label{training_performance_interp}
\end{figure}
The training $R^2$ of this protocol is shown in Fig.~\ref{training_performance_interp}.
We find that only at full batch does the bootstrapped model exhibit noteworthy training $R^2$. 
Interestingly, this coincides with the appearance of KA geometry in Fig.~\ref{random_batch_interpolation}.
For comparison we have also displayed the training $R^2$ of: a standard $2$-hidden layer MLP (where both hidden layers are $32$-dimensional), the base $1$-hidden layer MLP, an expanded $1$-hidden layer MLP with $264$ hidden neurons, and an untrained bootstrapped $1$-hidden layer MLP, in which the would-be feature extractor from the base model is untrained, i.e., the $A$, $a$ are fixed at their random initializations.
The $2$-hidden layer MLP shows that a second hidden layer is sufficient to reasonably well memorize the \texttt{random} function.
The untrained bootstrapped model is a ``random feature extractor" control that examines whether the extra parameters, that arise in the second $1$-hidden layer MLP due to the $m$-dimensional input (i.e., the $m \times m$ linear map vs. the $n \times m$ linear map, where $m > n$), increase the model's ability to memorize.
The expanded $1$-hidden layer MLP indicates that extra parameters in the base model are insufficient to match the ``random feature extractor" of the untrained bootstrapped model.
Our conclusion from this is that the \texttt{random}$(32)$ model is exhibiting genuine KA geometry when trained at full batch.
It is interesting that a standard metric, like $R^2$, misses this feature of model development.

\section{Other Interpolations}
\label{otherinterpolationsappendix}

In \S \ref{interpolatesection}, we used the $\lambda$-\texttt{xor} family of target functions \eqref{lambdaxorfunction} to study how KA geometry varies under interpolation between functions of differing complexity.
In this appendix, we will study two other families of functions.

\begin{figure}[h]
\centering
\begin{subfigure}[b]{0.45\textwidth}
\centering
\includegraphics[width=\textwidth]{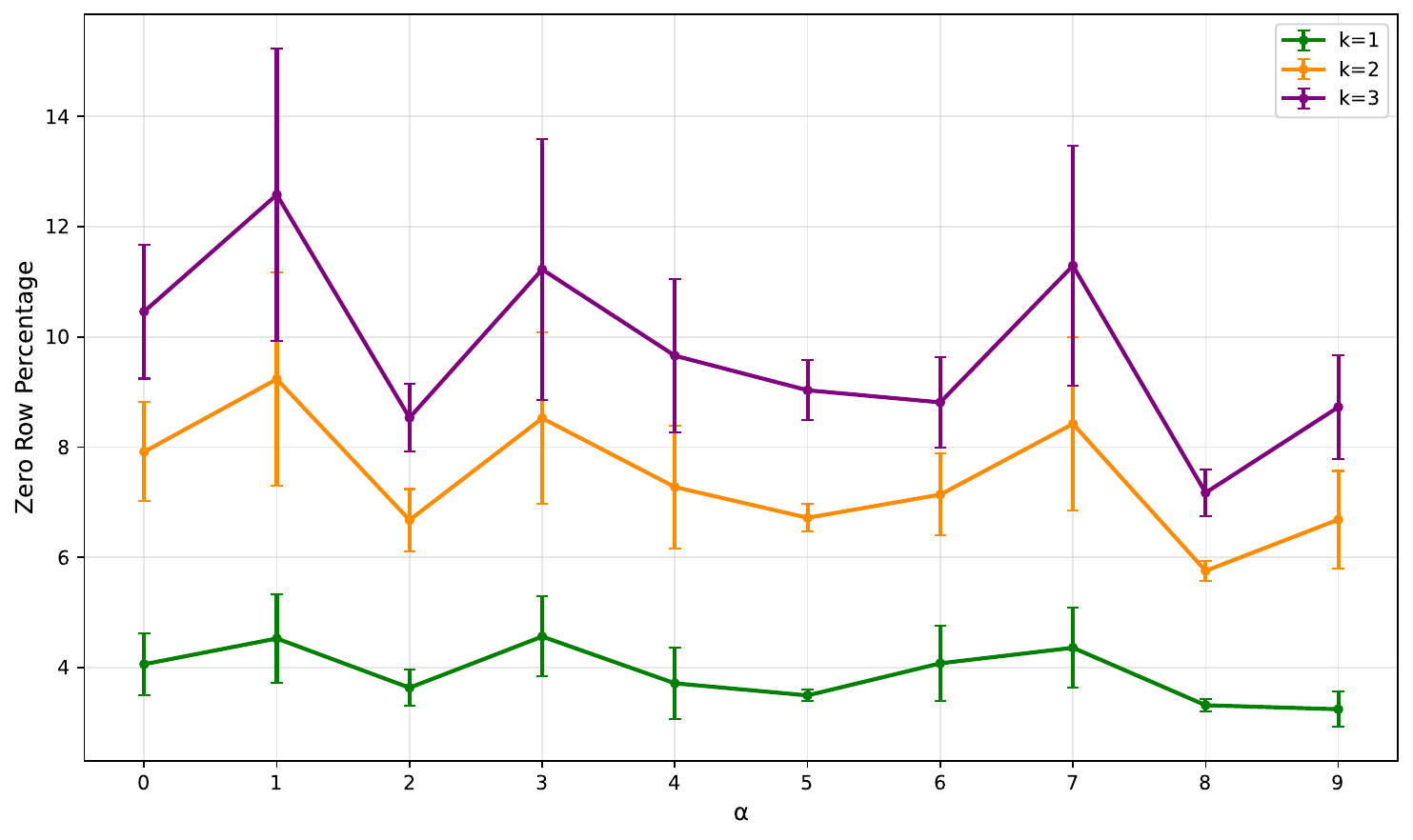}
\caption{Zero Rows}
\label{SO_zero_rows}
\end{subfigure}
\hfill
\begin{subfigure}[b]{0.45\textwidth}
\centering
\includegraphics[width=\textwidth]{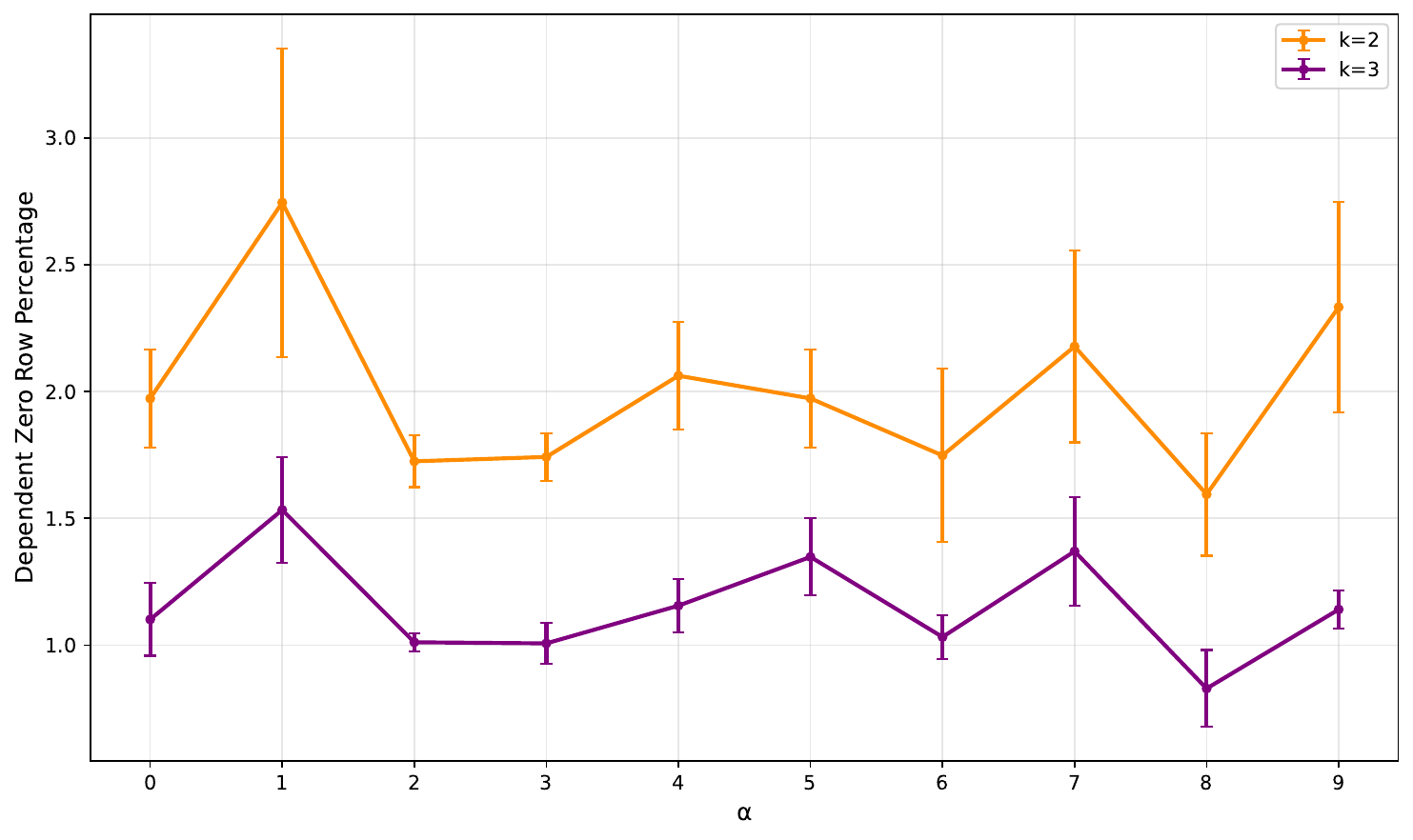}
\caption{Dependent Zero Rows}
\label{SO_dependent_zeros}
\end{subfigure}

\vspace{1em} 
\begin{subfigure}[b]{0.32\textwidth}
    \centering
    \includegraphics[width=\textwidth]{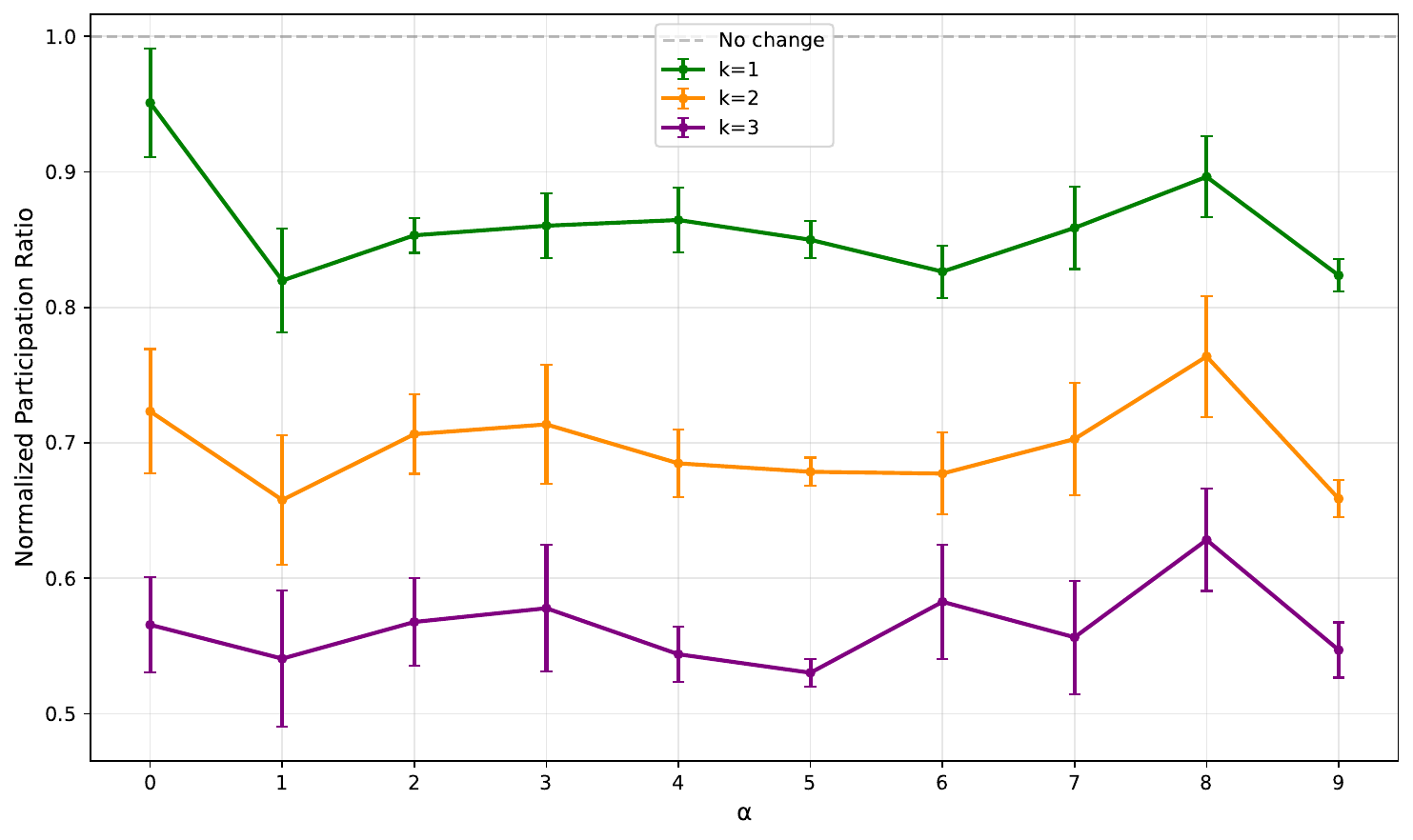}
    \caption{Participation Ratios}
    \label{SO_participation_ratios}
\end{subfigure}
\hfill
\begin{subfigure}[b]{0.32\textwidth}
    \centering
    \includegraphics[width=\textwidth]{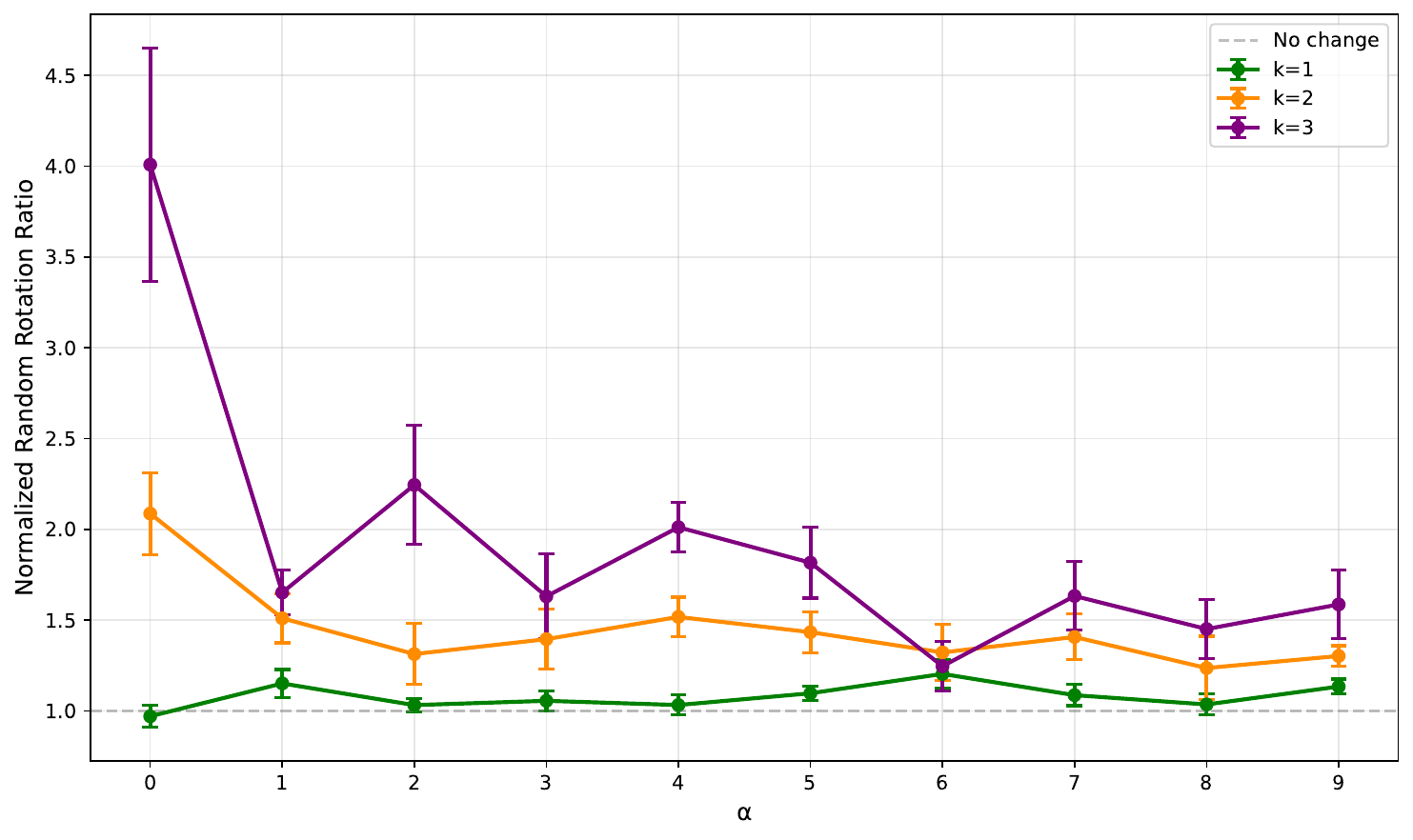}
    \caption{Random Rotation Ratios}
    \label{SO_random_rotation_ratios}
\end{subfigure}
\hfill
\begin{subfigure}[b]{0.32\textwidth}
    \centering
    \includegraphics[width=\textwidth]{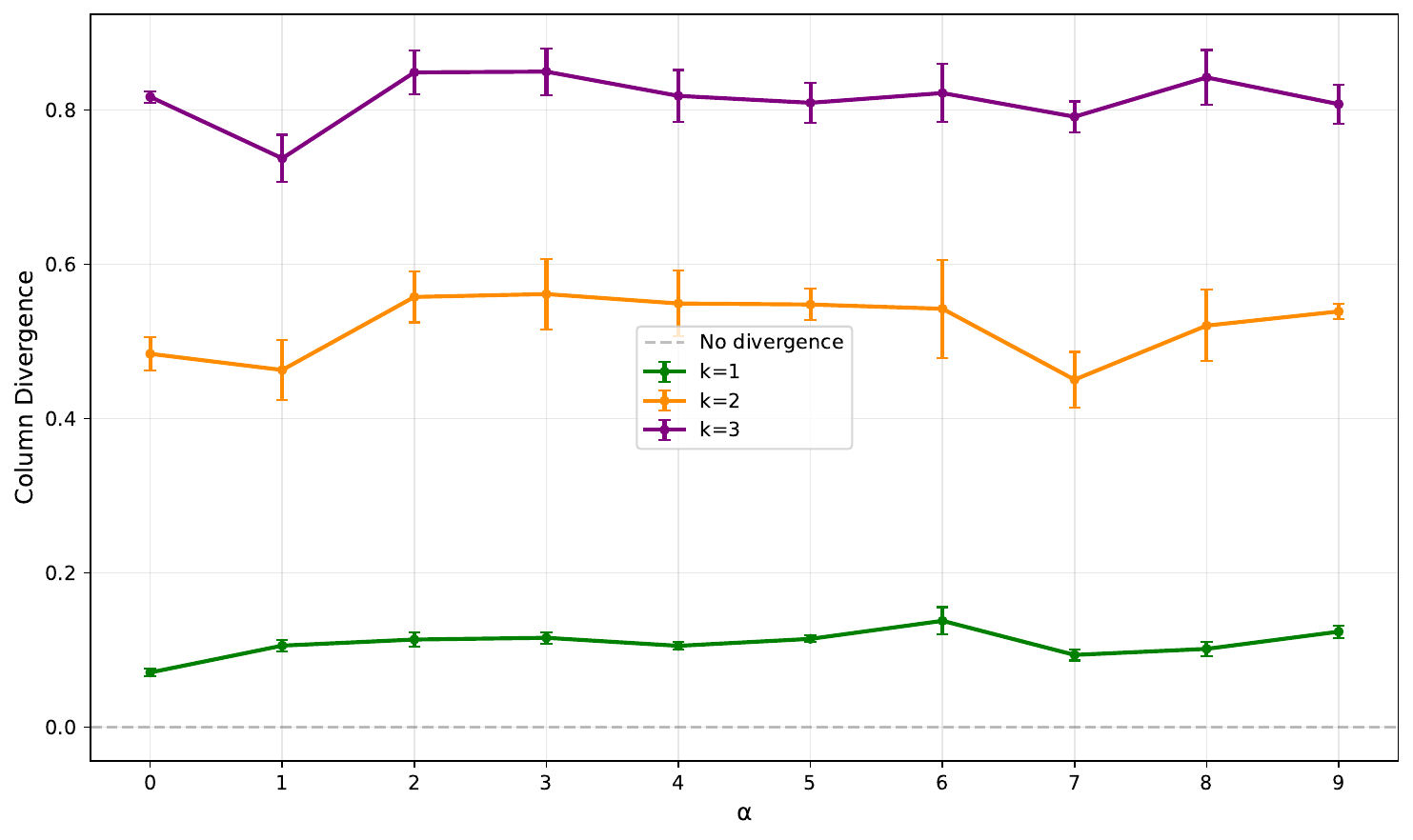}
    \caption{Column Divergences}
    \label{SO_column_divergences}
\end{subfigure}
\caption{
Zero row and minor concentration metrics in the $SO^{(\alpha)}$-\texttt{xor}$(32)$ models for $\alpha \in \{0, 1, \ldots, 9 \}$.
}
\label{SO_xor_interpolation_plots}
\end{figure}
The first is the $SO^{(\alpha)}$-\texttt{xor}$(\mathbf{x})$ family:
\begin{align}
\label{soxorfunction}
SO^{(\alpha)}\texttt{-xor}(\mathbf{x} ) & = \prod_i \sin(\pi z_i), \quad z_i = O^{(\alpha)}_{i i'} x_{i'} ,
\end{align}
where $O^{(\alpha)}$ is a randomly-chosen $n \times n$ orthogonal matrix.
Here, $\alpha$ simply labels the orthogonal transformation---we can think of it as the random seed for generating the matrix.
If $O^{(\alpha)}$ were the identity, $SO^{(\alpha)}$-\texttt{xor}$(\mathbf{x})$ would be the usual \texttt{xor} function.
We find that the $SO^{(\alpha)}$-\texttt{xor}$(32)$ model achieves training and test $R^2$ comparable to that of the \texttt{xor}$(32)$ model.
This function family is interesting to consider because one might worry that KA geometry would develop with respect to the basis induced by the transformation and therefore be undetectable, if present at all, by our metrics that make use of the preferred input and hidden space bases.
We find this not to be the case.

Fig.~\ref{SO_xor_interpolation_plots} shows how the zero row and minor concentration metrics vary over $10$ different rotations $O^{(\alpha)}$, $\alpha \in \{0, 1, \ldots, 9 \}$. 
There is no relation between the different $O^{(\alpha)}$ except that they are sampled uniformly from $SO(n)$.
The zero row and dependent zero row percentages are within error of the values reported in Table \ref{zerorowtable} for the \texttt{xor}$(32)$ model.
The normalized participation ratios and columns divergences are likewise similar---see Figs.~\ref{normalizedparticipationratiocomparison} and \ref{columnentropyocomparison}.
It is only the random rotation ratios in Fig.~\ref{SO_random_rotation_ratios} that differ by about an order of magnitude from the values in Fig.~\ref{normalizedmmrratiocomparison}.
Despite being smaller, this ratio remains greater than $1.0$, thus implying that minor alignment persists.
Taken together these results suggest that KA geometry continues to hold in the $SO^{(\alpha)}$-\texttt{xor} family.
In retrospect, if gradient descent is truly keying in on the KA-construction, this isn't surprising given the inner function's independence of the target.

\begin{figure}[h]
\centering
\begin{subfigure}[b]{0.45\textwidth}
\centering
\includegraphics[width=\textwidth]{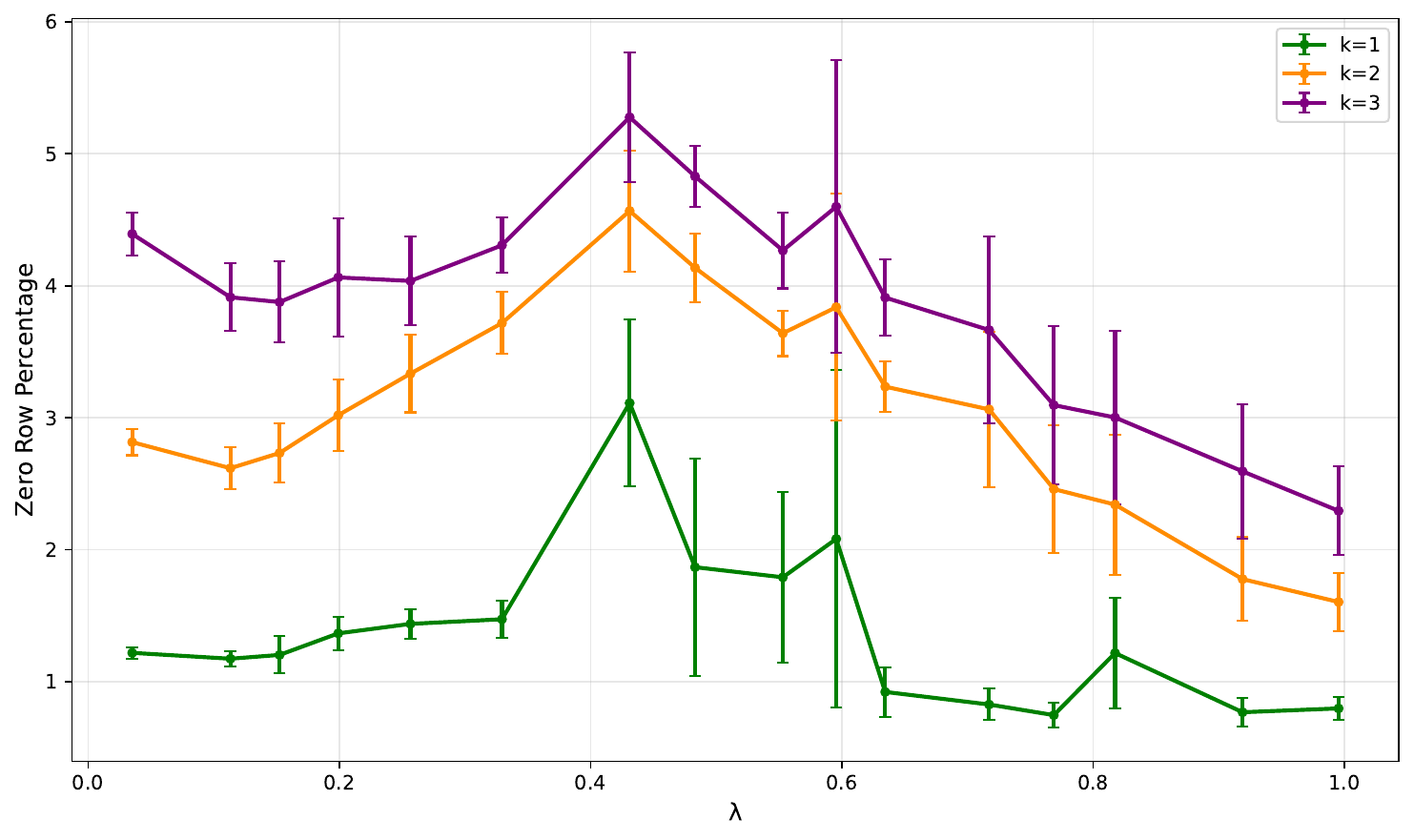}
\caption{Zero Rows}
\label{SO_zero_rows}
\end{subfigure}
 \hfill
\begin{subfigure}[b]{0.45\textwidth}
\centering
\includegraphics[width=\textwidth]{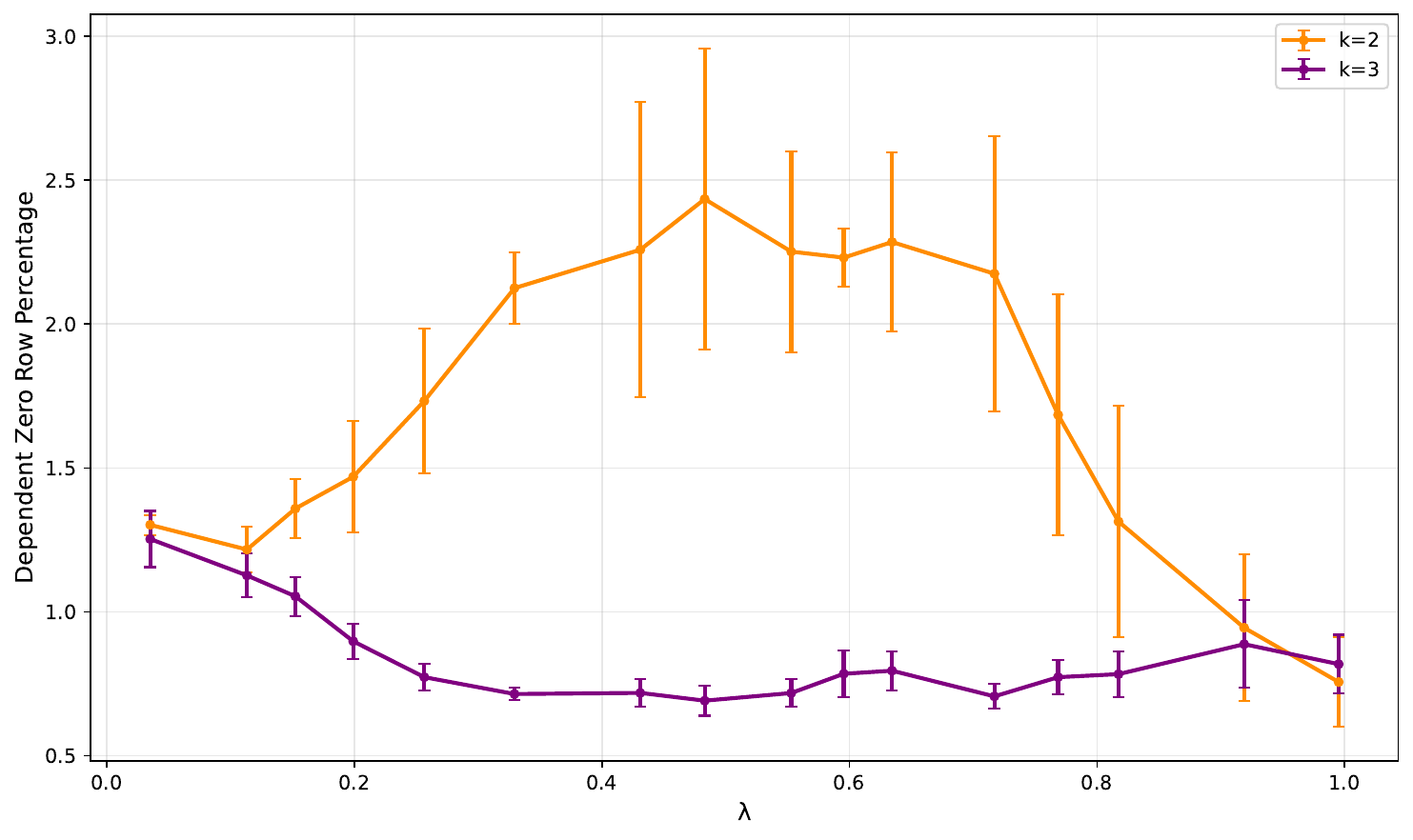}
\caption{Dependent Zero Rows}
\label{SO_dependent_zeros}
\end{subfigure}

\vspace{1em} 
\begin{subfigure}[b]{0.32\textwidth}
    \centering
    \includegraphics[width=\textwidth]{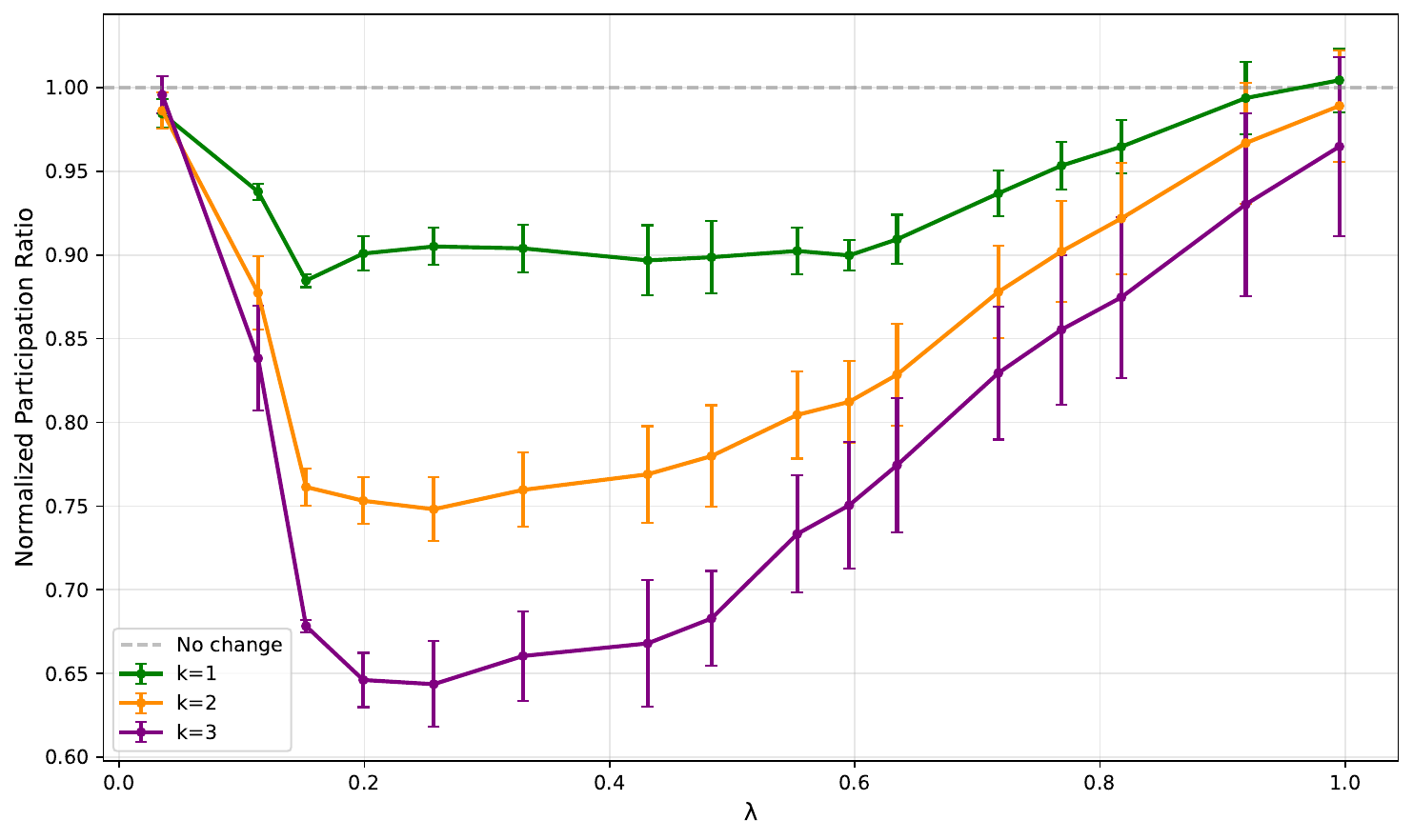}
    \caption{Participation Ratios}
    \label{SO_participation_ratios}
\end{subfigure}
\hfill
\begin{subfigure}[b]{0.32\textwidth}
    \centering
    \includegraphics[width=\textwidth]{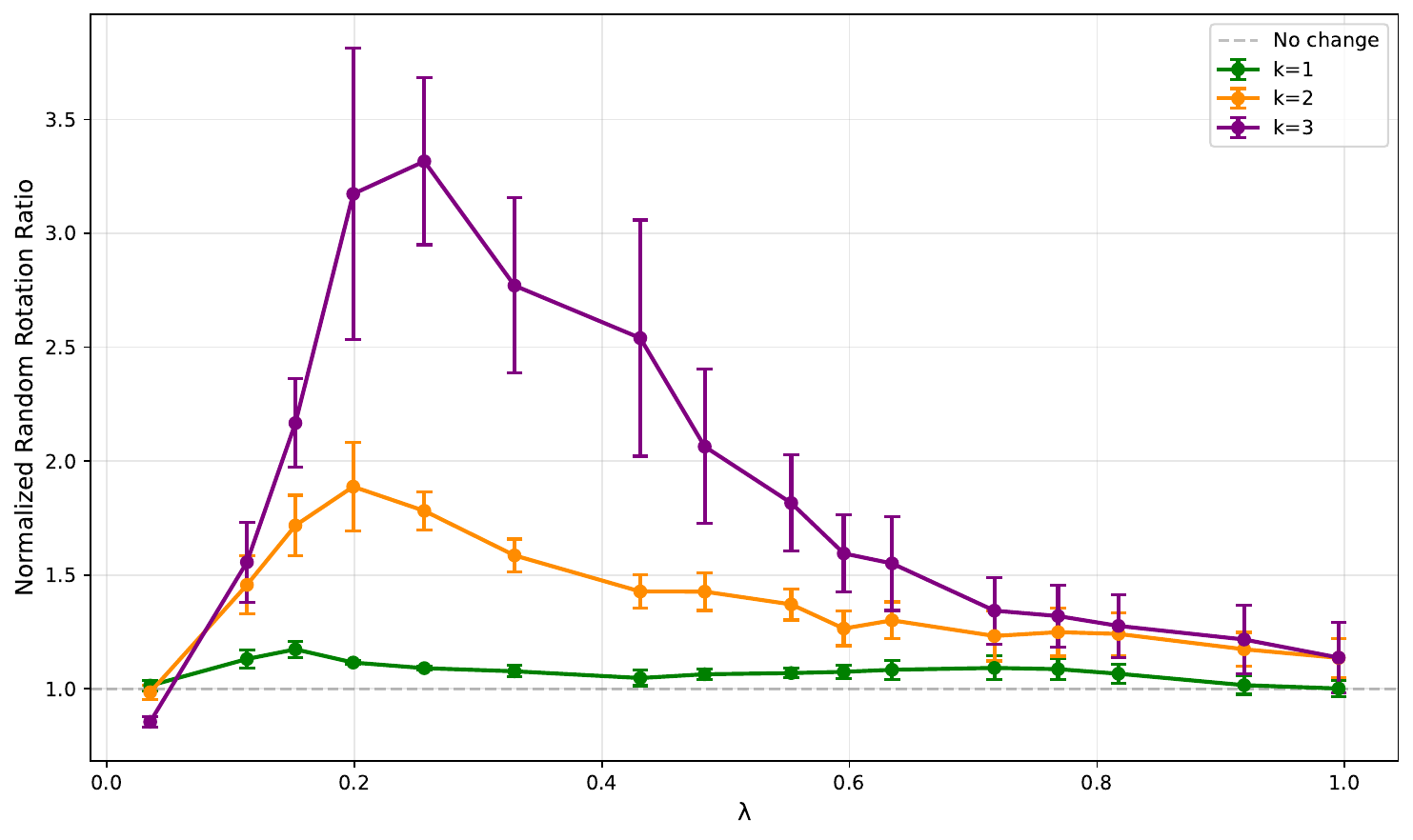}
    \caption{Random Rotation Ratios}
    \label{SO_random_rotation_ratios}
\end{subfigure}
\hfill
\begin{subfigure}[b]{0.32\textwidth}
    \centering
    \includegraphics[width=\textwidth]{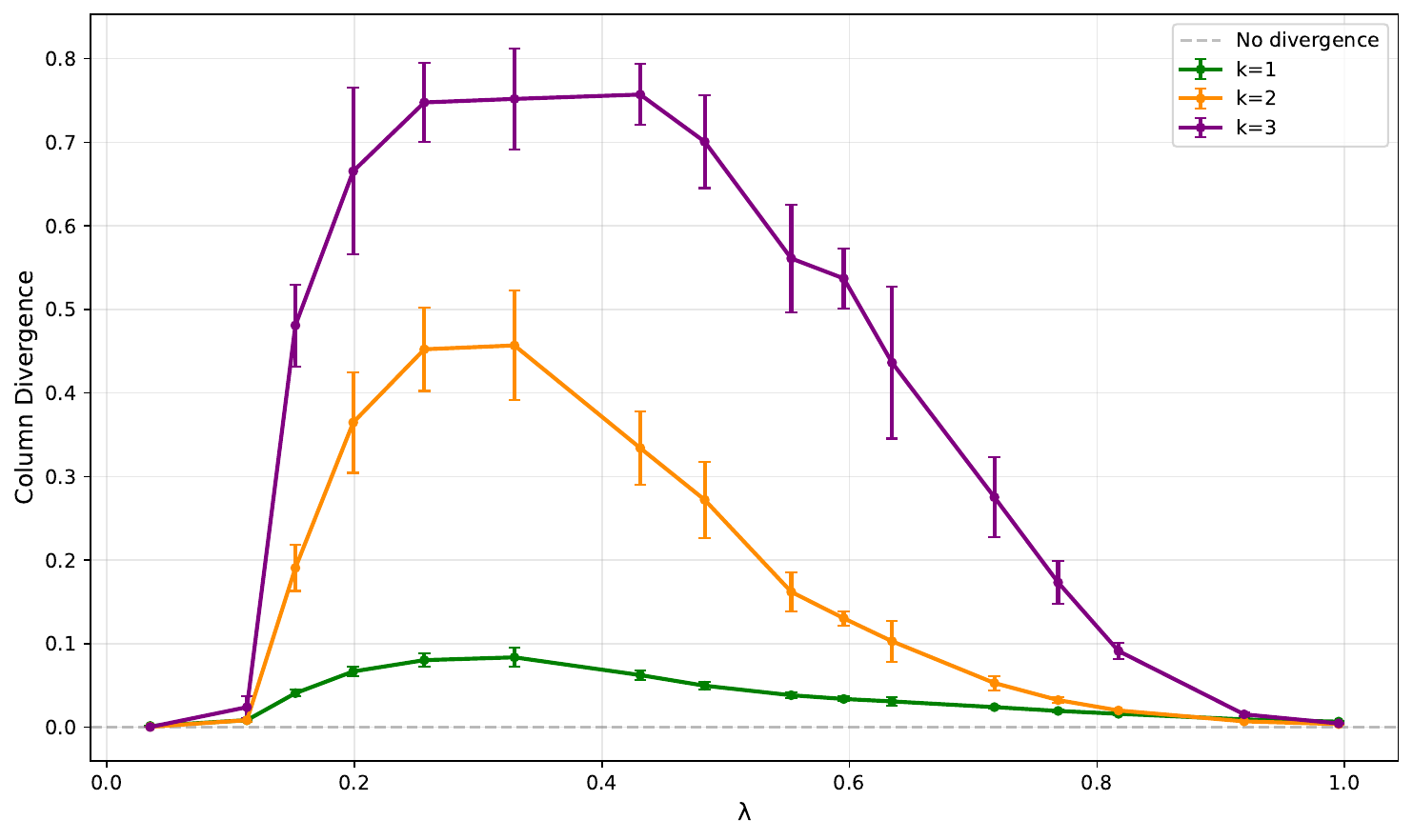}
    \caption{Column Divergences}
    \label{SO_column_divergences}
\end{subfigure}
\caption{
Zero row and minor concentration metrics in the \texttt{G}$^{(\lambda)}(32)$ models for $\lambda \in [0.10, 1.0]$.
}
\label{gauss_interpolation_plots}
\end{figure}
The second family of functions we study is 
\begin{align}
\texttt{G}^{(\lambda)}(\mathbf{x}) = \prod_i \exp(- {x_i^2 \over 2 \lambda^2}), \quad \lambda \in [.10, 1.0].
\end{align}
This function is positive on the domain $x_i \in [-1,1]$.
\texttt{G}$^{(\lambda)}(\mathbf{x})$ becomes increasingly peaked at the origin as $\lambda \rightarrow .10$, while for $\lambda \rightarrow 1.0$ the function becomes more uniform.
We find the function to be learnable (for our 1-hidden layer model with $m=32$ hidden dimensions) for all but the smallest values of $\lambda \approx .10$.
The zero row and minor concentration metrics are shown in Fig.~\ref{gauss_interpolation_plots}.
The plots are similar to the $\lambda$-\texttt{xor} interpolation plots in Fig.~\ref{zerorowmetricsinterpolation} and \ref{participation_permutation_entropy_interpolation}, except that now the function complexity decreases with increasing $\lambda$.
There seems to be a ``Goldilocks regime" for $0.15 \lessapprox \lambda \lessapprox 0.80$.
This is interesting because, in contrast to $\lambda$-\texttt{xor} family, the nodal structure of \texttt{G}$^{(\lambda)}(\mathbf{x})$ does not change with $\lambda$.

\bibliography{bigbib}
% \bibliography{bib_for_red_tok}

%\bibliography{draft8Notes}

\end{document}